\documentclass{article}

\makeatletter\def\input@path{{./latex_vendor/}{./}}\makeatother

\usepackage{ifthen}
\providecommand{\venue}{iclr}
\providecommand{\papermode}{final}

\newif\ificlr
\newif\ifpaperfinal
\ifthenelse{\equal{\venue}{iclr}}{\iclrtrue}{\iclrfalse}
\ifthenelse{\equal{\papermode}{final}}{\paperfinaltrue}{\paperfinalfalse}

\ifpaperfinal
  \newcommand{\projecturl}{https://trackeverything.github.io/}
\else
  \newcommand{\projecturl}{https://iclr-8302.netlify.app/}
\fi

\ificlr
  \usepackage{iclr2027_conference,times}
  \ifpaperfinal\iclrfinalcopy\fi
  \makeatletter
  \def\@maketitle{\vbox{\hsize\textwidth
    \centering
    {\LARGE\sc \@title\par}
    \ificlrfinal
      \def\And{\end{tabular}\hfil\linebreak[0]\hfil
              \begin{tabular}[t]{c}\bf\rule{\z@}{24pt}\ignorespaces}%
      \def\AND{\end{tabular}\hfil\linebreak[4]\hfil
              \begin{tabular}[t]{c}\bf\rule{\z@}{24pt}\ignorespaces}%
      \begin{tabular}[t]{c}\bf\rule{\z@}{24pt}\@author\end{tabular}\par
    \else
      \def\And{\end{tabular}\hfil\linebreak[0]\hfil
              \begin{tabular}[t]{c}\bf\rule{\z@}{24pt}\ignorespaces}%
      \def\AND{\end{tabular}\hfil\linebreak[4]\hfil
              \begin{tabular}[t]{c}\bf\rule{\z@}{24pt}\ignorespaces}%
      \begin{tabular}[t]{c}\bf\rule{\z@}{24pt}Anonymous authors\\Paper under double-blind review\end{tabular}\par
    \fi
    \vskip 0.3in minus 0.1in}}
  \makeatother
\else
  \ifpaperfinal
    \usepackage[preprint]{neurips_2026}
  \else
    \usepackage{neurips_2026}
  \fi
\fi

\usepackage[utf8]{inputenc} 
\usepackage[T1]{fontenc}    
\usepackage{hyperref}       
\usepackage{url}            
\usepackage{graphicx}       
\usepackage{adjustbox}      
\usepackage{booktabs}       
\usepackage{longtable}      
\usepackage[font=footnotesize]{subcaption} 
\usepackage{amsmath}        
\usepackage{amsfonts}       
\usepackage{nicefrac}       
\usepackage{microtype}      
\usepackage{xcolor}         
\usepackage{float}          

\usepackage{xspace}
\newcommand{\model}{TrackEverything\xspace}

\usepackage{multirow}

\usepackage{xcolor} 
\usepackage{colortbl}     
\usepackage{pifont}       
\usepackage[T1]{fontenc}

\definecolor{LightGray}{gray}{0.95}

\usepackage{cleveref}       
\newcommand{\projectsite}{\href{\projecturl}{\texttt{\projecturl}}}

\title{\model: \\Long Horizon Dense Tracking via De-Duplicating 3D Scene Representations}

\author{%
  Ayush Jain\textsuperscript{1,2} \quad
  Sreeharsha Paruchuri\textsuperscript{1}\ifpaperfinal\thanks{Equal contribution.}\fi \quad
  Ishita Gupta\textsuperscript{1}\ifpaperfinal\footnotemark[1]\fi \quad
  Fan Zhang\textsuperscript{2} \\
  \bf
  Tanner Schmidt\textsuperscript{2} \quad
  Jakob Engel\textsuperscript{2} \quad
  Katerina Fragkiadaki\textsuperscript{1} \quad
  Adam W.~Harley\textsuperscript{2} \\
  \normalfont
  \textsuperscript{1}Carnegie Mellon University \quad
  \textsuperscript{2}Meta \\
  \projectsite
}

\begin{document}

\maketitle
\ificlr
  \ifpaperfinal
    \lhead{}%
    \renewcommand{\headrulewidth}{0pt}%
  \else
    \lhead{Under review as a conference paper at ICLR 2027}%
  \fi
\fi
\begin{figure}[H]
\vspace{-2.5em}
\centering
\includegraphics[width=\textwidth]{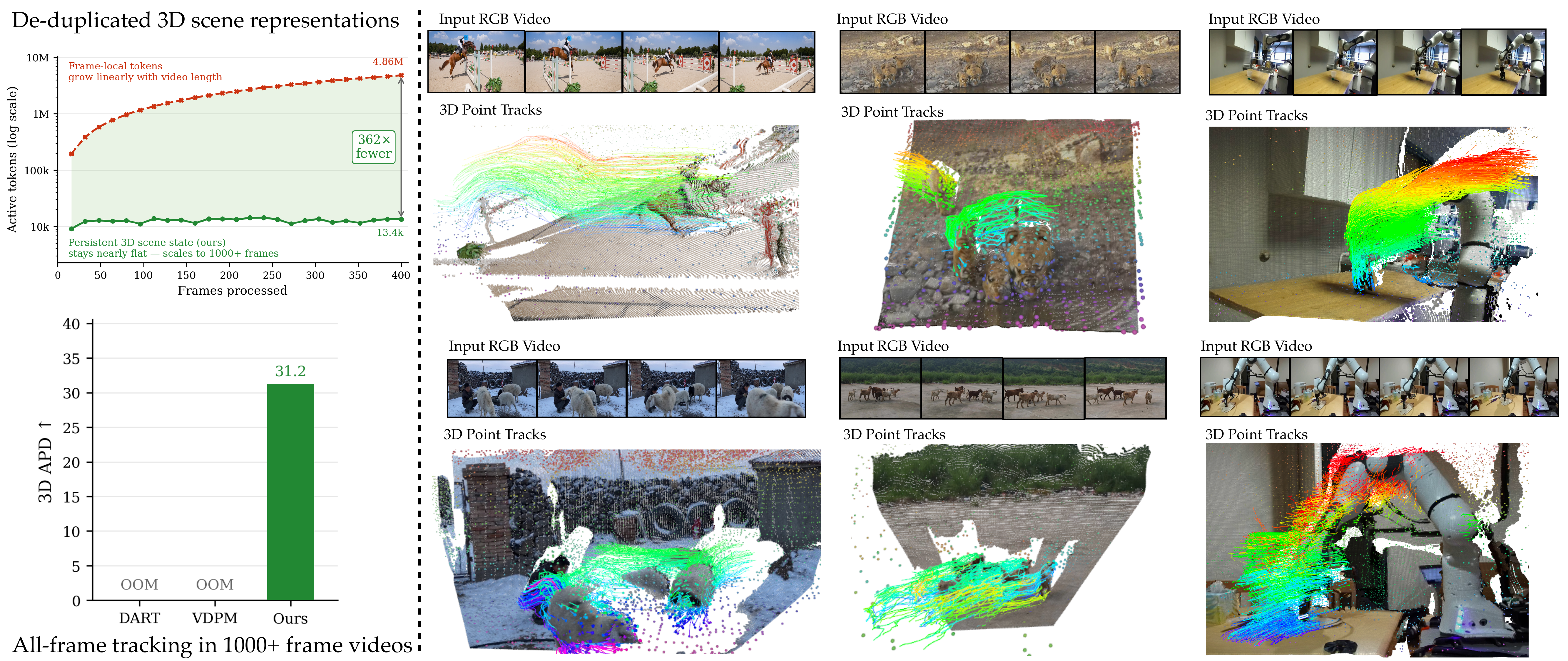}
\caption{\textbf{\model} tracks \emph{all} points across \emph{all} frames of long videos (1000+ frames) in 3D world-coordinate space.
\textbf{Left:} a persistent, voxel-de-duplicated 3D scene representation grows with unique scene content rather than video length, remaining practical while prior all-frame dense 3D trackers run out of memory.
\textbf{Right:} for each input video (top) we show the output dense 3D tracks (bottom). Trails are shown for dynamic points; static content is rendered as a 3D point cloud.}
\label{fig:teaser}
\vspace{-0.1in}
\end{figure}

    \begin{abstract}
    \vspace{-1em}
        Existing point tracking models face a fundamental tradeoff: they can either track a sparse set of query points over long horizons, or track all points  across only short clips. We introduce \textbf{\model}, a 3D point tracker that breaks this trade-off by representing videos as persistent 3D scene tracks in world coordinates. Grounded in the insight that videos are 2D projections of an underlying 3D world, \model decouples model complexity from video duration, allowing it to scale with unique physical scene geometry instead. 
        Our approach introduces three key innovations. \textbf{First}, we employ a \emph{voxelization-based de-duplication} mechanism at sliding-window boundaries to merge co-located tracks, preventing repeated observations of the same surface from redundantly accumulating. \textbf{Second}, we decompose tracking into an \emph{endpoint refiner} that predicts each point's destination and static-versus-dynamic classification, followed by a lightweight \emph{trajectory refiner} that decodes dense trajectories exclusively for dynamic points. \textbf{Third}, we propose \emph{3D WAFT}, replacing memory-prohibitive 4D correlation volumes with efficient feature sampling in the scene cloud. 
        To the best of our knowledge, \model is the first 3D tracker capable of tracking all visible points across videos exceeding 1000 frames within 40\,GB of GPU memory. On TAPVid-3D, \model outperforms all open-source all-frame dense 3D trackers by more than 20\% APD on short clips, while remaining competitive with state-of-the-art sparse trackers on long sequences,  despite tracking far more points. 
        \end{abstract}
\vspace{-0.08in}
\section{Introduction}
\vspace{-0.08in}

Video is the primary medium through which modern AI models perceive the dynamic physical world, underpinning vision-language models~\citep{gemini}, vision-language-action policies~\citep{pi}, and point trackers~\citep{alltracker} alike. Yet processing video in frame-by-frame 2D representations incurs a computational cost and memory footprint that scales linearly with video duration: each new frame introduces a full grid of visual tokens that must be processed, regardless of whether it reveals novel scene content or merely re-observes surfaces already seen. This per-frame redundancy severely limits scalability to long videos.

In point tracking, for example, this redundancy forces an unnatural divide. \emph{Sparse trackers}~\citep{tapip3d,spatialtrackerv2,cotracker3} can track over extended time horizons, but only for a sparse set of user-specified query points, leaving the rest of the scene unmodeled. \emph{Dense trackers}, in contrast, operate in frame-local pixel space where redundant re-encoding causes computation to explode with video length. Consequently, existing dense methods are forced into severe compromises: they either track only points visible in the first frame~\citep{alltracker,delta,startrack,any4d}, ignoring newly disoccluded surfaces, or track all points but remain restricted to short clips of 48--64 frames~\citep{d4rt,vdpm}. As a result, no prior method can track all visible scene points across full-length videos.

These standard video processing regimes under-exploit the fact that videos are merely streams of 2D camera projections of an underlying 3D world. 
In a 3D-based scene representation, there is an opportunity to scale with unique scene content rather than video duration. Exploiting this gap between \emph{frame-based scaling} and \emph{3D scene-centric scaling} is the central motivation of our work.

We introduce \textbf{\model}, a 3D point tracker representing videos as \textit{non-redundant} and \textit{long-horizon} 3D scene tracks in world coordinates (\Cref{fig:teaser}). Instead of redundantly tracking every 2D pixel across time, \model maintains a 3D representation that scales with unique physical geometry rather than frame count. Our architecture achieves this through three key innovations:
\textbf{First}, we introduce a \emph{voxelization-based de-duplication} mechanism at temporal window boundaries. When points from different frames converge onto the same physical surface, they occupy the same 3D coordinates and are merged into a single canonical track, preventing redundant copies from accumulating over time.
\textbf{Second}, we decompose tracking into an \emph{endpoint refiner} and a \emph{trajectory refiner}. The endpoint refiner estimates each point's destination at the end of the window and classifies it as static or dynamic; the trajectory refiner then decodes dense within-window trajectories exclusively for dynamic points. Because static points remain stationary in world coordinates, this factorization concentrates dense decoding on the small moving subset of the scene.
\textbf{Third}, we propose \emph{3D WAFT}, an extension of warp-aligned feature transforms~\citep{waft} to 3D point clouds. By sampling feature maps at projected source and target locations, 3D WAFT replaces memory-intensive 4D correlation volumes with efficient template matching at a fraction of the compute cost.

Concretely, \model processes a video with a sliding window strategy. Within each window, 2D visual features are unprojected into a world-coordinate feature cloud using camera poses and depth from sensors or off-the-shelf estimators (e.g., VGGT~\citep{vggt_omega}). An \emph{endpoint refiner} transformer iteratively updates each point's estimated 3D position at the end of the window using 3D WAFT feature sampling, while simultaneously predicting its static-vs.-dynamic classification. For points classified as dynamic, a \emph{trajectory refiner} decodes dense within-window trajectories conditioned on their motion and local scene context, while static points trivially retain their positions. Finally, at each window boundary, the 3D feature cloud is voxelized at the updated positions to merge co-located points, yielding a compact, de-duplicated set of tracks for the next window, to be tracked alongside newly revealed content.


We evaluate \model on standard 3D point tracking benchmarks including TAPVid-3D~\citep{tapvid3d}. On short clips, where prior all-frame dense baselines remain computationally viable, \model outperforms all open-source all-frame dense 3D tracking methods by more than 20\% APD. Crucially, while existing all-frame dense trackers exhaust GPU memory beyond roughly 96 frames, \model is the first 3D tracker that scales to tracking all points across videos exceeding 1000 frames—operating within 40\,GB of GPU memory. Despite tracking orders of magnitude more points, \model remains competitive with state-of-the-art sparse trackers and first-frame dense methods on long sequences, while providing dense tracks for the entire scene.

Beyond point tracking, we believe our persistent, de-duplicating 3D dynamic scene representations provide a long-sought substrate for downstream foundation models. While explicit 3D representations have shown clear benefits for vision-language models~\citep{qwen3d,univlg}, robotic manipulation policies~\citep{3dda}, and video generation~\citep{persist3d}, existing approaches remain largely confined to static or quasi-static scenes. By making all-point tracking computationally tractable over extended horizons, \model opens an exciting path toward extending these models to dynamic, real-world environments.

Our contributions are summarized as follows:
\begingroup
\setlength{\parskip}{1pt}%
\setlength{\topsep}{4pt}%
\setlength{\partopsep}{0pt}%
\setlength{\itemsep}{2pt}%
\setlength{\parsep}{0pt}%
\begin{itemize}
\item We introduce \model, which to the best of our knowledge is the first 3D point tracker capable of tracking all visible points across long horizons (1000+ frames).
\item We propose a voxelization-based de-duplication mechanism, making our representation complexity scale with unique physical scene geometry rather than video duration.
\item We introduce two efficiency gains: (1) ``endpoints before trajectories'' as a problem decomposition, to focus dense decoding on dynamic points, and (2) 3D WAFT to replace expensive 4D correlations with efficient feature sampling, together making all-frame tracking practical.
\item On TAPVid-3D, \model outperforms all open-source all-frame dense 3D trackers by more than 20\% APD on short clips, and matches state-of-the-art sparse trackers on long sequences, and is the only method that scales all-point tracking to 1000+ frames.
\end{itemize}
\endgroup

We invite readers to view full-length dense 3D tracking visualizations on our project website: \projectsite.
\vspace{-0.1in}
\section{Related Work}
\vspace{-0.08in}

\noindent \textbf{Sparse Point Tracking.}\quad
PIPs~\citep{pips} and TAP-Vid~\citep{tapvid} established the modern paradigm for sparse point tracking: tracking a small set of user defined query points across videos. Subsequent methods~\citep{cotracker3} follow the same recipe: sliding-window inference for long videos, and recurrent iterative refinement guided by 4D correlation volumes between query features and dense video features~\citep{raft}. However, constructing and querying dense correlation volumes scales with the number of query points, thus limiting the number of points that can be tracked. Recent work extends sparse tracking into 3D: SpatialTracker~\citep{spatialtracker} represents points via tri-plane representations, TAPIP-3D~\citep{tapip3d} performs neighborhood attention within 3D feature clouds, and SpatialTracker-v2~\citep{spatialtrackerv2} jointly refines geometry and motion using both 2D and 3D correlations. While these methods use 3D representations, their internal feature clouds retain a full grid of points for every video frame, causing memory to scale linearly with video duration. \model adopts the recurrent sliding-window methodology, but diverges in three respects: (1)~our 3D representation is not tied to pixel space, but grows only when new scene content appears, via voxelization-based de-duplication; (2)~we replace expensive 4D correlations with a cheaper 3D extension of WAFT~\citep{waft}; and (3)~we track \emph{all} points across all frames of a long video, rather than being restrictedto a sparse query set.

\noindent \textbf{Dense Point Tracking.}\quad
Dense point tracking aims to track points sampled densely in a grid instead of user-specified sparse query points. In 2D, DELTA~\citep{delta}, AllTracker~\citep{alltracker}, and CoWTracker~\citep{cowtracker} track dense point grids in long videos, but are strictly limited to points visible in the first frame. Omnimotion~\citep{omnimotion} can track all points across all frames, but requires 8+ hours of per-video test-time optimization. 
In 3D, ST4RTrack~\citep{startrack}, DPM~\citep{dpm}, and Any4D~\citep{any4d} use DUSt3R/VGGT-style backbones for dense tracking, but remain limited to tracking points visible in first frame and only on short clips. VDPM~\citep{vdpm}, TraceAnything~\citep{traceanything}, and D4RT~\citep{d4rt} aim to track all points across all frames, but remain confined to short videos as well due to the growing cost of pixel-space representations. Since the same surface is represented independently in every frame, tracking all points in a $512{\times}512$, 200-frame video requires on the order of $10^{10}$ predictions. \model represents videos within a persistent, world-coordinate 3D scene space. By continuously de-duplicating co-located points across sliding windows, our representation collapses temporal redundancy and expands only when genuinely new geometry is revealed, making all-frame dense tracking across 1000+ frame videos computationally viable.

\begin{figure}[t]
\centering
    \includegraphics[width=\textwidth]{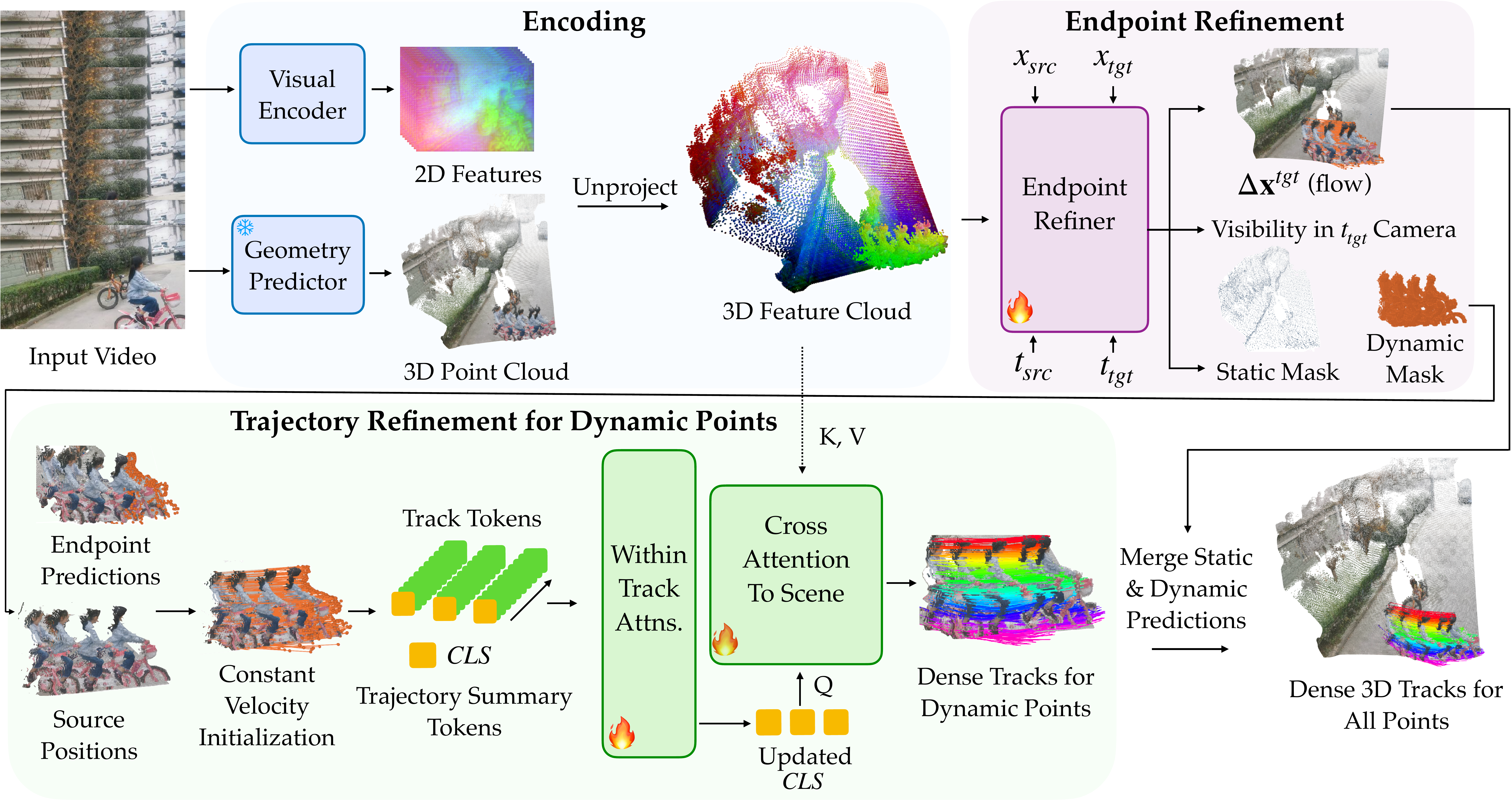}
    \caption{\textbf{\model architecture.}
    \model processes videos in  sliding windows of length $L$.
    \textit{Encoding:} Video frames are encoded and unprojected into a world-coordinate 3D feature cloud using depths/pointmaps from sensors or  geometry predictors.
    \textit{Endpoint Refinement:} Conditioned on source/target position embeddings and 3D WAFT feature sampling, a recurrent transformer iteratively predicts each point's 3D offset $\Delta\mathbf{x}$ at the target timestep, visibility, and a static/dynamic classification.
    \textit{Trajectory Refinement:} Trajectories for dynamic points are initialized via constant-velocity interpolation and refined via within-track temporal self-attention and cross-attention from track summary tokens (\textsc{[cls]}) to the scene cloud.
    \textit{Output and De-duplication:} At window boundaries, points arriving at identical 3D voxels are merged by mean-pooling before initializing the subsequent window, bounding representation size across long videos.
    }
    \label{fig:method}
\end{figure}
\section{Method}

Given an RGB video $\mathbf{I} \in \mathbb{R}^{T \times H \times W \times 3}$ and per-frame 3D pointmaps $\mathbf{P} \in \mathbb{R}^{T \times H \times W \times 3}$ in world coordinates (obtained from sensors or feedforward reconstruction models such as VGGT-$\Omega$~\citep{vggt_omega}), our goal is to track all visible scene surfaces over time.
Specifically, \model predicts dense 3D trajectories $\mathbf{X} \in \mathbb{R}^{N \times T \times 3}$, visibility logits $\mathbf{V} \in \mathbb{R}^{N \times T}$, and static/dynamic logits $\mathbf{S} \in \mathbb{R}^N$, for all $N$ unique scene points  (\Cref{fig:method}), with support also for user-specified subpixel query points.
Standard frame-by-frame 2D tracking incurs memory and compute costs that scale linearly with video duration $T$, as each new frame redundantly re-introduces previously observed surfaces.
\model resolves this bottleneck by tracking directly on persistent 3D physical entities in world coordinates across temporal windows of length $L$.
Within each window, frames are encoded and unprojected into a world-coordinate \emph{feature cloud} (\Cref{sec:encoding}).
An iterative refinement process then decouples tracking into an \emph{endpoint refiner} (\Cref{sec:transformer}) that estimates end-of-window  locations and classifies points as static or dynamic, and a \emph{trajectory refiner} (\Cref{sec:trajectory}) that decodes dense within-window paths only for dynamic points. Crucially, \model achieves dense tracking with low inference complexity by coupling sliding-window inference with cross-window voxelization (\Cref{sec:sliding}): rather than carrying all accumulated trajectories forward, we merge tracks that arrive at the same 3D voxel, since two physical entities cannot occupy the same position at the same time.

\vspace{-0.08in}
\subsection{Encoding}\label{sec:encoding}
\vspace{-0.1in}
We process the video in non-overlapping windows of $L{=}16$ frames.
Within each window, every frame is encoded by a frozen DINOv3-small~\citep{dinov3} backbone followed by a trainable ViT Adapter head~\citep{vitadapter}, yielding a feature map of shape $(L, H/4, W/4, D)$ with $D{=}384$.
Each frame's feature map is reshaped into a set of $\tfrac{H}{4}{\times}\tfrac{W}{4}$ tokens, paired with world-coordinate 3D positions given by the corresponding entry in the pointmap (downsampled to the same resolution). Following \cite{dust3r}, the scene geometry is normalized to a canonical scale globally in the reference frame of the first camera ($\mathrm{cam}_0$).
To reduce spatial redundancy within each frame, tokens are quantized into voxels of size $v$ via $\lfloor \mathbf{x} / v \rfloor$, and tokens sharing a voxel index are merged via mean-pooling.
This voxelization is performed per-frame rather than across frames to avoid erroneous merges when distinct surfaces occupy the same 3D region at different times. 
The per-frame voxelized tokens are concatenated across all $L$ frames to form a space-time feature cloud of 
shape $(N, D)$. We refer to these features individually as $\mathbf{f}_i$. 
Note that until we finish processing the window (\Cref{sec:sliding}), we retain multiple representatives of any surface observed in multiple timesteps.

\vspace{-0.08in}
\subsection{Iterative Refinement}\label{sec:iterative}
\vspace{-0.1in}

Following the standard point tracking approach~\citep{pips,tapir}, we begin with zero-velocity initializations, coming from either observed 3D position (for newly disoccluded points) or the previous window's output (for tracked points), 
and we seek to iteratively refine these initializations into trajectories that track the scene. 

Directly decoding full $L$-frame trajectories for every point in the dense feature cloud is computationally prohibitive and largely wasteful, as many scene points in real-world environments are stationary in world coordinates.
Furthermore, \emph{propagating tracking state across sliding windows does not require full trajectories}---to hand points off to the next window and merge co-located surfaces, the model only needs to know each point's destination at the window boundary.

We therefore decouple tracking into two stages:
(1) An \textbf{endpoint refiner} (\Cref{sec:transformer}) that runs over all active scene points to predict their 3D positions at the window boundary and classify them as static or dynamic. 
(2) A \textbf{trajectory refiner} (\Cref{sec:trajectory}) that decodes dense within-window paths exclusively for dynamic points. 

\vspace{-0.05in}
\subsubsection{Endpoint Refiner}\label{sec:transformer}
\vspace{-0.1in}
The goal of the endpoint refiner is to iteratively update each point's estimated 3D location at a designated target timestep $t_{\mathrm{tgt}}$, while predicting its visibility and classifying it as static or dynamic.

\textit{Timestep definitions.}
Each point is associated with a \emph{source timestep} $t_{\mathrm{src}}$: the frame in which it was first observed in the current window ($t_{\mathrm{src}} \in \{0, \dots, L{-}1\}$), or $t_{\mathrm{src}} = -1$ for points carried over from prior windows.
The \emph{target timestep} $t_{\mathrm{tgt}}$ designates the destination frame for which positions are estimated.
At inference time and for all intermediate windows during training, we set $t_{\mathrm{tgt}} = L-1$ (the window boundary), as boundary destinations are precisely what is needed to advance the sliding window.
On the final window during training, $t_{\mathrm{tgt}}$ is sampled uniformly at random from $\{0, \dots, L{-}1\}$, teaching the model to predict flow to arbitrary timesteps.

\textit{Position embeddings:}
To inform the transformer of both where/when points originate and where/when they are being tracked, we enrich each point's feature $\mathbf{f}_i$ with four sinusoidal embeddings~\citep{nerf}:
a spatial embedding for the initial 3D coordinate $\mathbf{x}_{\mathrm{src}, i}$, a spatial embedding for the current estimated destination $\mathbf{x}_{\mathrm{tgt}, i}^{(k-1)}$, a temporal embedding for the source observation frame $t_{\mathrm{src}, i}$, and a temporal embedding for the target frame $t_{\mathrm{tgt}}$. Each sinusoidal embedding is projected to $D$ dimensions with a dedicated linear layer ($E_{a}^{b}$) before the embeddings are merged via a sum, yielding $\mathbf{f}_i \leftarrow \mathbf{f}_i + E^{\mathrm{xyz}}_{\mathrm{src}}(\mathbf{x}_{\mathrm{src},i}) + E^{\mathrm{xyz}}_{\mathrm{tgt}}(\mathbf{x}_{\mathrm{tgt},i}^{(k-1)}) + E^{t}_{\mathrm{src}}(t_{\mathrm{src},i}) + E^{t}_{\mathrm{tgt}}(t_{\mathrm{tgt}})$.


\textit{Feature construction:} 
Before each transformer pass, we apply a 3D variant of the Warp-Aligned Feature Transform (WAFT)~\citep{waft}.
For every point, we project its current target-position estimate onto the 2D feature map of frame $t_{\mathrm{tgt}}$ and bilinearly sample the image feature there, producing a \emph{target feature} $\mathbf{g}_{\mathrm{tgt}}$.
We similarly look up the \emph{source feature} $\mathbf{g}_{\mathrm{src}}$ at its birth location, so that the model cannot ``forget'' the original appearance.
These are concatenated with the main feature $\mathbf{f}_i$ to form the actual transformer input:
$[\mathbf{f}_i;\; \mathbf{g}_{\mathrm{src},i};\; \mathbf{g}_{\mathrm{tgt},i}]$.
Providing both source and target features gives the model an implicit template-matching signal: when the two features are similar, the current estimate is likely accurate; when they are dissimilar, refinement is needed. As shown by WAFT~\citep{waft}, self-attention over features constructed this way is sufficient for correspondence-finding. 
We note that the source feature $\mathbf{g}_\mathrm{src}$ only needs to be sampled once. 

\textit{Attention and decoding:}
The concatenated features are processed by $N_{\mathrm{attn}}$ layers of multi-head self-attention over the full feature cloud.
A small MLP head then decodes three outputs from each $\mathbf{f}_i$: a 3D residual $\Delta\mathbf{x}$, a visibility logit $v$, and a static/dynamic logit $s$.
The target position is updated as $\mathbf{x}^{(k)}_{\mathrm{tgt}} = \mathbf{x}^{(k-1)}_{\mathrm{tgt}} + \Delta\mathbf{x}^{(k)}$. The updated  $\mathbf{f}_i$ are carried forward to the next module and the next window, but we do not backpropagate through time. 


\vspace{-0.05in}
\subsubsection{Trajectory Refiner}\label{sec:trajectory}
\vspace{-0.1in}
After the endpoint refiner
 updates each point's target-time position $\mathbf{x}^{(k)}_{\mathrm{tgt}}$, 
 and static/dynamic classification logit $s$, we refine the $L$-frame (window-length) trajectory of each dynamic point. 

\textit{Trajectory initialization:}
In the first iteration, each dynamic point's trajectory is initialized by \textit{constant}-velocity interpolation between its source and target positions. Otherwise, we use the previously estimated trajectory but with the $t_\mathrm{tgt}$ position updated by the endpoint refiner. 

\textit{Feature construction:} For each timestep along the trajectory, we concatenate three vectors: (i) the point's feature from the transformer output, (ii) its source feature $\mathbf{g}_{\mathrm{src}}$ (sampled at the point's birth location), and (iii) a feature bilinearly sampled from the 2D feature map at the position predicted by the constant-velocity estimate for that timestep.
We also append a \textsc{[cls]} token (initialized from the cloud feature) to serve as a track-level summary, yielding track features of shape $(N_{\mathrm{dyn}}, L{+}1, D)$.

\textit{Attention and decoding:}
The decoder applies multiple layers of two alternating operations: (1) within-track temporal self-attention, which lets each point's tokens exchange information along its own trajectory, and (2) cross-attention from the \textsc{[cls]} token to the full feature cloud, which injects scene context back into each track.
Tracks do not attend to one another, following PIPs~\citep{pips} and D4RT~\citep{d4rt}; this independence means we can decode a random subset of tracks during training yet decode all tracks at test time without distribution shift.
After the final attention layer, 
an MLP decodes a per-point per-timestep 3D flow residual and visibility logit, for each of the $N_{\mathrm{dyn}} \cdot L$ tokens (discarding the \textsc{[cls]} token of each track).

\paragraph{Iterative loop.}
The iterative refinement steps (endpoint update and trajectory update) are repeated for $K$ iterations within each window.
Each iteration re-samples the WAFT features at the newly updated target positions, giving the transformer progressively better evidence from the bilinear samples. Additionally, the $\mathrm{xyz}$ position embeddings to the feature cloud are updated with the most recent 3D estimates. This recurrent design is what allows the architecture to also chain predictions across sliding windows: the mechanism is identical, differing only in that the initialization comes from the previous window rather than from the previous iteration. We describe this next. 


\vspace{-0.08in}
\subsection{Sliding Window Inference and De-Duplication}\label{sec:sliding}
\vspace{-0.1in}
At the end of each window, we voxelize the feature cloud based on predicted target positions. Unlike the per-frame voxelization used during encoding, this operation merges points across multiple origin frames. This cross-frame merging is valid because all points are now tracked to the same target timestep; we thus leverage the rule that two entities cannot occupy the same voxel at the same time.  

\vspace{-0.08in}
\subsection{Implementation details}\label{sec:training}
\vspace{-0.1in}
\paragraph{Supervision.}
We supervise the predicted target positions/trajectories with an L2 loss against 3D ground-truth. 
Visibility and static/dynamic classification are each supervised with cross-entropy loss.
Because our training datasets provide ground-truth tracks for only a subset of points, we compute losses only over the labelled points.
To save compute, we decode full trajectories for a subset of dynamic points at training time.
\vspace{-0.1in}
\paragraph{Training recipe.}
\model has 41M trainable parameters. The DINOv3-small backbone (20M) is frozen; the ViT Adapter head and all other modules are trained from scratch, except for the transformer attention layers which are initialized from the last layers of DINOv2-small~\citep{dinov2}.
We train on 8 L40S-46GB GPUs with a per-GPU batch size of 1 and a learning rate of $1{\times}10^{-4}$.
Training proceeds in two stages: 100k iterations on 32-frame videos, followed by 300k iterations on a mix of 32- and 64-frame videos.
Point clouds are normalized to unit norm following DUSt3R~\citep{dust3r}. We apply color jittering, random image/depth blur, and erase augmentations following TAPIP-3D~\citep{tapip3d}. Our training data consists of Kubric~\citep{kubric}, PointOdyssey~\citep{pod}, and Dynamic Replica~\citep{dr}, each of which provide 3D track annotations for a sparse set of points.

\vspace{-0.1in}
\paragraph{Sparse query points (optional).}\label{sec:query}
In addition to dense tracking, \model can accept a set of query points specified as $(t, x, y)$ tuples. This is useful for tracking points at subpixel locations. Their features are obtained by bilinearly sampling the 2D feature map at the given frame and pixel coordinates. These query features are concatenated with the feature cloud and processed identically to grid-sampled points throughout the transformer and trajectory decoder, with the sole difference that they are excluded from voxelization so that they are never merged with nearby points.

\section{Experiments}\label{sec:experiments}
\vspace{-0.1in}
We evaluate \model on standard 3D point tracking benchmarks and compare against both sparse and dense tracking baselines. Beyond benchmarking our own model, we aim to provide a comprehensive evaluation across method paradigms to elucidate the trade-offs in this space.
\vspace{-0.1in}
\subsection{3D Point Tracking Overview}

\noindent \textbf{Benchmark.}
We use the TAPVid-3D benchmark~\citep{tapvid3d}, which comprises three datasets spanning diverse settings: Aria Digital Twin (ADT)~\citep{adt} with egocentric indoor videos (${\sim}300$ frames), DriveTrack~\citep{drivetrack} with egocentric driving videos ($25$--$300$ frames), and Panoptic Studio (PStudio)~\citep{pstudio} with multi-camera lab captures (${\sim}150$ frames). Because dense 3D ground truth is infeasible to annotate for real-world dynamic scenes, TAPVid-3D annotates and evaluates on a sparse set of query points. Evaluating on these sampled points provides an unbiased indicator of tracking fidelity across the full scene, while enabling direct comparison with sparse tracking baselines that track only the queried points. Following prior work~\citep{any4d}, we evaluate our method and baselines on the official minival split of 50 video clips from each TAPVid-3D dataset. Where dense ground truth is available in simulation, we evaluate dense tracking on held-out splits of PointOdyssey~\citep{pod} and Dynamic Replica~\citep{dr} in the supplementary (\Cref{tab:dr_pod_dense}). We also provide extensive dense tracking visualizations on diverse real-world scenes on our project page.

\noindent \textbf{Metrics.}
We report the official \textbf{APD} metric~\citep{tapvid3d}, which measures the fraction of predicted points within a distance threshold of the ground truth, averaged over a set of thresholds. Prior work uses two threshold conventions. \textbf{APD-P} (point-wise) uses depth-relative thresholds obtained by unprojecting pixel thresholds $\{1,2,4,8,16\}$\,px into 3D via the camera intrinsics; this is the TAPVid-3D metric used by sparse trackers. \textbf{APD-M} (metric) uses fixed distance thresholds $\{0.1, 0.3, 0.5, 1.0\}$\,m, as adopted by several dense trackers. APD-P is typically stricter than APD-M. We report both to enable fair comparison across methods. We additionally report Average Jaccard (AJ) and Occlusion Accuracy (OA) for all methods in the supplementary material.

\noindent \textbf{Baselines.}
We compare against three classes of methods. (a)~\emph{Sparse trackers} track a user-specified set of query points: SpatialTracker-v2~\citep{spatialtrackerv2}, TAPIP-3D~\citep{tapip3d}, and CoTracker3~\citep{cotracker3} (a 2D tracker lifted to 3D). (b)~\emph{First-frame dense trackers} track all points visible in the first frame but cannot track points that appear later: DeltaV2~\citep{delta} and Any4D~\citep{any4d}. DeltaV2 also has a sparse branch that can track user-specified query points originating in any frame, but that branch does not perform dense tracking. (c)~\emph{All-frame dense trackers} track all points across all frames: VDPM~\citep{vdpm} and D4RT~\citep{d4rt}. We include D4RT's reported numbers for reference, though we note that D4RT has no publicly available code or weights, uses ${\sim}20{\times}$ our parameter count with private training data, and is evaluated only on 48-frame clips.

\noindent \textbf{Geometry and evaluation protocol.}
Several baselines accept or require external 3D geometry. We provide VGGT-$\Omega$~\citep{vggt_omega} pointmaps to SpatialTracker-v2, TAPIP-3D, CoTracker3, Any4D, DeltaV2, and our model; for SpatialTracker-v2 and Any4D we also report numbers with their built-in geometry. VDPM and D4RT do not support external geometry. Prior dense trackers have only been evaluated on short clips (48 or 64 frames); we therefore report all methods on 48-frame clips and additionally report sparse trackers, DeltaV2, and \model on the full-length sequences.

\noindent \textbf{Training data and model size.}
\model has 61M parameters (including a frozen 20M DINOv3-small backbone) and is trained on Kubric~\citep{kubric}, PointOdyssey~\citep{pod}, and Dynamic Replica~\citep{dr}. In contrast, several baselines train on significantly larger corpora: for instance, SpatialTracker-v2 trains on 17 datasets. D4RT is a 1B+ parameter model trained on orders of magnitude more public data alongside large undisclosed private datasets. A full breakdown of training datasets for all baselines is detailed in the supplementary.

\vspace{-0.08in}
\subsection{Main Results}
\Cref{tab:tapvid3d_full} presents the full comparison. Among the all-frame dense trackers, \model outperforms VDPM by over 20\% APD-P on average. D4RT reports higher numbers on 48-frame DriveTrack and PStudio clips (and trails behind on ADT), but uses ${\sim}20{\times}$ our parameters, trains on significantly more data including private datasets, has no public code or weights, and does not track beyond 48 frames. Neither VDPM nor D4RT scales to full-length sequences (marked N/A); \model is the only all-frame dense tracker that does. Compared to first-frame dense trackers, \model is competitive with DeltaV2 using same VGGT-$\Omega$ backend: ahead on average and on ADT and slightly behind on DriveTrack and PStudio. DeltaV2 can only track dense points originating in the first frame (and sparse user-specified points from other frames), whereas \model densely tracks all points across all frames---a strictly more demanding setting. On full-length videos, \model is competitive with sparse 3D trackers despite tracking orders of magnitude more points.
\providecommand{\dgray}[1]{\textcolor{gray!55}{#1}}
\providecommand{\na}{--}
\providecommand{\second}[1]{\underline{#1}}
\providecommand{\nrun}{N/A}
\providecommand{\nrunrow}{\nrun & \nrun & \nrun & \nrun & \nrun & \nrun & \nrun & \nrun}
\providecommand{\nrunrowg}{\dgray{\nrun} & \dgray{\nrun} & \dgray{\nrun} & \dgray{\nrun} & \dgray{\nrun} & \dgray{\nrun} & \dgray{\nrun} & \dgray{\nrun}}
\providecommand{\oom}{\textsc{oom}}
\providecommand{\TEname}{\shortstack[l]{\textbf{TrackEverything}\\[-0.12em]\textbf{(Ours)}}}
\providecommand{\TEnamesingle}{\textbf{TrackEverything (Ours)}}
\providecommand{\geoO}{$\Omega$}
\providecommand{\geoP}{Pi3}
\providecommand{\geoN}{--}
\definecolor{oursblue}{RGB}{232,240,254}
\providecommand{\oursrow}{\rowcolor{oursblue}}

\begin{table}[H]
    \vspace{-0.06in}
    \centering
    \caption{\textbf{3D point tracking on TAPVid-3D}.
    $\Omega$: VGGT-$\Omega$ geometry; --: native geometry.
    \textbf{Bold} = best, \underline{underline} = second-best in that block. 
    D4RT is closed-source and is excluded from ranking.
    }
    \label{tab:tapvid3d_full}
    \footnotesize
    \setlength{\tabcolsep}{3.4pt}
    \renewcommand{\arraystretch}{1.05}
    \begin{adjustbox}{max width=\textwidth,center}
    \begin{tabular}{@{}lc*{8}{c}@{\hskip 0.7em}*{8}{c}@{}}
        \toprule
        & & \multicolumn{8}{c}{\textbf{first 48 frames}}
        & \multicolumn{8}{c}{\textbf{full-length video}} \\
        \cmidrule{3-10} \cmidrule{11-18}
        & & \multicolumn{2}{c}{ADT} & \multicolumn{2}{c}{DriveTrack} & \multicolumn{2}{c}{PStudio} & \multicolumn{2}{c}{Avg}
        & \multicolumn{2}{c}{ADT} & \multicolumn{2}{c}{DriveTrack} & \multicolumn{2}{c}{PStudio} & \multicolumn{2}{c}{Avg} \\
        \cmidrule{3-4} \cmidrule{5-6} \cmidrule{7-8} \cmidrule{9-10}
        \cmidrule{11-12} \cmidrule{13-14} \cmidrule{15-16} \cmidrule{17-18}
        \textbf{Method} & Geo
        & APD-P & APD-M & APD-P & APD-M & APD-P & APD-M & APD-P & APD-M
        & APD-P & APD-M & APD-P & APD-M & APD-P & APD-M & APD-P & APD-M \\
        \midrule
        \multicolumn{18}{@{}l}{\textit{Sparse tracking}} \\
        CoTracker3 & \geoO
            & 43.1 & 87.8 & \textbf{29.7} & \textbf{42.8} & 29.6 & 87.9 & \second{34.1} & \textbf{72.8}
            & 34.5 & 84.4 & \second{27.4} & \second{43.9} & 27.1 & 87.9 & 29.7 & \textbf{72.1} \\
        SpatialTracker-v2 & \geoN
            & 37.7 & 87.0 & 28.6 & 41.3 & 25.1 & 84.1 & 30.5 & 70.8
            & \oom & \oom & \oom & \oom & 23.9 & 84.0 & \na & \na \\
        SpatialTracker-v2 & \geoO
            & 43.8 & 87.4 & \second{29.4} & \second{42.4} & 24.9 & 84.2 & 32.7 & 71.3
            & 36.6 & 84.6 & \textbf{28.1} & \textbf{45.1} & 23.7 & 84.6 & 29.5 & 71.4 \\
        TAPIP-3D & \geoO
            & \second{45.6} & \textbf{89.0} & 26.7 & 39.4 & \second{30.0} & \second{88.2} & \second{34.1} & 72.2
            & \second{37.2} & \second{85.0} & 24.5 & 40.3 & \textbf{27.6} & \second{88.2} & \second{29.8} & 71.2 \\
        \oursrow
        \TEnamesingle & \geoO
            & \textbf{45.7} & \second{88.7} & 28.2 & 40.3 & \textbf{30.1} & \textbf{88.7} & \textbf{34.7} & \second{72.6}
            & \textbf{40.0} & \textbf{85.1} & 26.5 & 42.4 & \second{27.2} & \textbf{88.5} & \textbf{31.2} & \second{72.0} \\
        \addlinespace[0.25em]
        \multicolumn{18}{@{}l}{\textit{First-frame dense tracking}} \\
        ST4RTrack & \geoN
            & \na & \na & \na & 1.0 & \na & 53.1 & \na & \na
            & \nrunrow \\
        Any4D & \geoN
            & 6.0 & \second{64.4} & 7.1 & 11.8 & 3.0 & 63.8 & 5.4 & 46.7
            & \nrunrow \\
        Any4D & \geoO
            & 6.8 & 63.2 & 7.7 & 10.9 & 4.5 & \second{65.8} & 6.3 & 46.6
            & \nrunrow \\
        DeltaV2 & \geoO
            & \second{44.3} & \textbf{88.7} & \textbf{28.4} & \textbf{41.5} & \textbf{30.4} & \textbf{88.7} & \second{34.4} & \textbf{73.0}
            & \second{34.7} & \second{79.9} & \textbf{27.3} & \textbf{44.2} & \textbf{27.7} & \second{87.9} & \second{29.9} & \second{70.7} \\
        \oursrow
        \TEnamesingle & \geoO
            & \textbf{45.7} & \textbf{88.7} & \second{28.2} & \second{40.3} & \second{30.1} & \textbf{88.7} & \textbf{34.7} & \second{72.6}
            & \textbf{40.0} & \textbf{85.1} & \second{26.5} & \second{42.4} & \second{27.2} & \textbf{88.5} & \textbf{31.2} & \textbf{72.0} \\
        \addlinespace[0.25em]
        \multicolumn{18}{@{}l}{\textit{All-frame dense tracking}} \\
        \dgray{D4RT} & \dgray{\geoN}
            & \dgray{40.8} & \dgray{\na} & \dgray{41.0} & \dgray{\na} & \dgray{49.6} & \dgray{\na} & \dgray{43.8} & \dgray{\na}
            & \nrunrowg \\
        VDPM & \geoN
            & \second{4.8} & \second{65.0} & \second{14.7} & \second{23.3} & \second{13.1} & \second{82.1} & \second{10.9} & \second{56.8}
            & \nrunrow \\
        \oursrow
        \TEnamesingle & \geoO
            & \textbf{45.7} & \textbf{88.7} & \textbf{28.2} & \textbf{40.3} & \textbf{30.1} & \textbf{88.7} & \textbf{34.7} & \textbf{72.6}
            & \textbf{40.0} & \textbf{85.1} & \textbf{26.5} & \textbf{42.4} & \textbf{27.2} & \textbf{88.5} & \textbf{31.2} & \textbf{72.0} \\
        \bottomrule
    \end{tabular}
    \end{adjustbox}
    \vspace{-0.1in}
\end{table}

\vspace{-0.08in}
\subsection{Complexity Analysis of Dense and Sparse Point Tracking}
\begin{figure}[H]
  \centering
  \includegraphics[width=\textwidth]{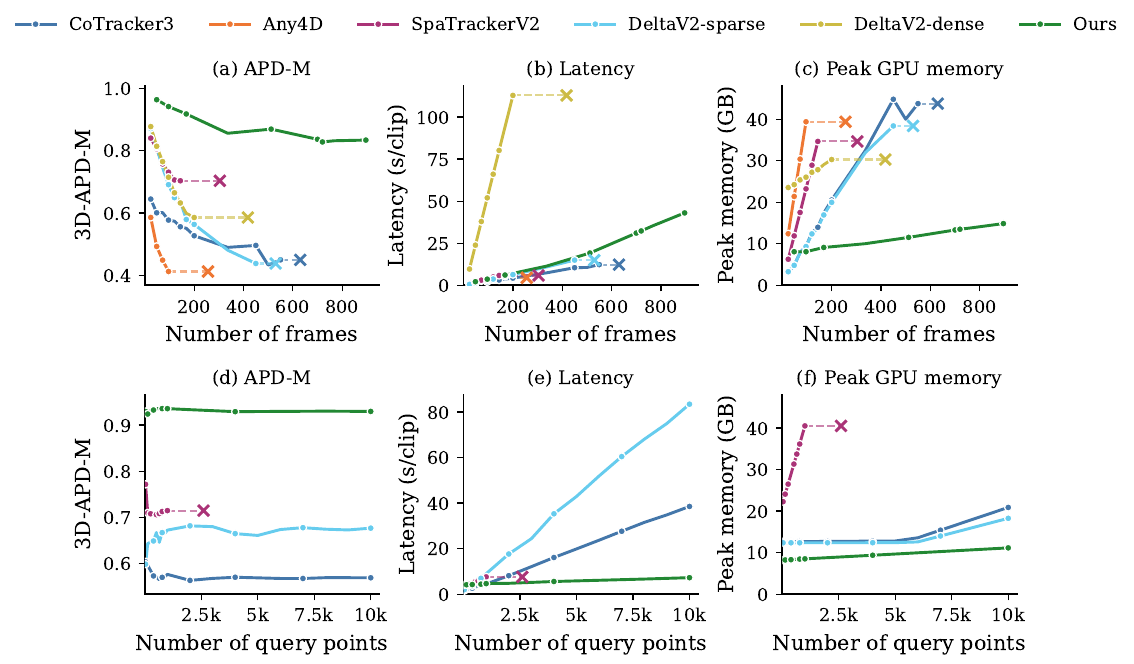}
  \caption{\textbf{Complexity analysis on PointOdyssey.} Accuracy (APD-M), end-to-end latency, and peak GPU memory vs.  video length at 384 query points (top row, a--c), and query point count at 120 frames (bottom row, d--f). `\texttt{x}' marks out-of-memory. Dense baselines exceed memory in 100-200 frames; sparse baselines track only a few thousand points, and OOM at ~600 frames. TrackEverything uniquely tracks all points across all frames with modest latency and under 30GB GPU memory.}
  \label{fig:all_methods_benchmarking}
  \vspace{-0.1in}
\end{figure}
To understand how existing methods scale with video length and query point count, we measure accuracy (APD-M), inference latency, and peak GPU memory as we independently vary (a)~input frame count and (b)~query point count. All experiments run on a single L40S-46G GPU and include end-to-end inference time (\Cref{fig:all_methods_benchmarking}). Notably, \model densely tracks \textit{all} points in the video in addition to any supplied query points, whereas the sparse baselines track only the queried subset. We  include similar APD-P analysis in the supplementary.

\noindent \textbf{Setup.}
We compare against SpatialTracker-v2 (sparse 3D), CoTracker3-3D (sparse 2D lifted to 3D), and dense tracking methods DeltaV2 and Any4D. We evaluate on the PointOdyssey~\citep{pod} validation set, which is one of the only datasets with long videos ($1000{+}$ frames) and fairly dense annotations ($10\text{k}{+}$ points). Our method, Any4D, and SpatialTracker-v2 all include PointOdyssey in their training data; CoTracker3 and DeltaV2 do not. The focus in these experiments is mainly on latency and memory trends, and not on absolute accuracy. We report APD-M here and APD-P in the supplementary. When a method exceeds GPU memory we mark it as OOM.

\noindent \textbf{Scaling with video length.}
We fix the number of query points to $384$ for methods that accept this parameter. As shown in~\Cref{fig:all_methods_benchmarking}a, accuracy degrades for all methods as the video grows longer, but the decline is notably steeper for Any4D, which processes all frames jointly. In terms of latency (\Cref{fig:all_methods_benchmarking}b), CoTracker3 is the fastest method, though it tracks only the $384$ queried points whereas \model tracks every point in the video. DeltaV2-dense is the slowest, exceeding 100\,s per clip at 200 frames for tracking first-frame points alone. For peak GPU memory (\Cref{fig:all_methods_benchmarking}c), \model exhibits the lowest footprint and the slowest growth rate: it remains under 30\,GB even at 900 frames. Any4D exceeds memory at 96 frames, SpatialTracker-v2 and DeltaV2-dense at approximately 200 frames, DeltaV2-sparse at approximately 500 frames, and CoTracker3 at approximately 600 frames. Three design choices contribute to our efficiency: (i)~a compact 3D representation that de-duplicates spatially redundant points, (ii)~WAFT-based attention in place of expensive 4D correlation volumes, and (iii)~a sliding-window inference scheme.

\noindent \textbf{Scaling with number of query points.}
We fix the video length to 120 frames and vary the number of query points. We omit Any4D and DeltaV2-dense, as they do not accept query-point input. As shown in~\Cref{fig:all_methods_benchmarking}d, all methods largely maintain accuracy as the query count increases. However, their computational profiles diverge sharply. CoTracker3 and DeltaV2-sparse start with the lowest latency but surpass \model at approximately 750 points (DeltaV2-sparse) and 2{,}000 points (CoTracker3), as shown in~\Cref{fig:all_methods_benchmarking}e. SpatialTracker-v2's latency rises steeply and exceeds memory at 1{,}000 points. \model maintains near-constant latency because additional query points do not incur expensive 4D correlation operations. In terms of memory (\Cref{fig:all_methods_benchmarking}f), SpatialTracker-v2 is the most demanding method, while CoTracker3 and \model both remain below 30\,GB.

\noindent \textbf{Takeaways.}
Existing dense trackers (Any4D, VDPM, DeltaV2-dense) do not scale to long videos in accuracy, latency, or memory. Sparse 3D trackers like SpatialTracker-v2 exhibit rapid growth in both latency and memory as either axis increases. CoTracker3 (2D sparse lifted to 3D) scales well in both dimensions but exceeds memory at approximately 600 frames and becomes slow beyond several thousand query points. \model is the only method that can densely track all points across all frames of long videos while also accepting up to 10{,}000 additional query points, with high accuracy, modest latency, and a GPU footprint that remains under 30\,GB. Overall, \model is the first method that makes dense, long-video 3D point tracking practical on a single GPU.

\vspace{-0.08in}
\subsection{Additional Experiments and Ablation Studies}

\begin{table*}[!t]
    \centering
    \footnotesize
    \setlength{\tabcolsep}{3.5pt}
    \captionsetup[subtable]{justification=centering}
    \caption{\textbf{Additional experiments and ablations.} $^\dagger$reported by D4RT.
    TE:~\model.}
    \label{tab:ablation_studies}
    \begin{subtable}[t]{0.32\textwidth}
        \centering
        \subcaption{Input Geometry source.}
        \label{subtab:sensor_geometry}
        \begin{adjustbox}{max width=\linewidth,center}
        \begin{tabular}{@{}lcc@{}}
            \toprule
            & PStudio & ADT \\
            \midrule
            D4RT$^\dagger$ & 49.6 & 40.8 \\
            TE+Pi3 & 23.5 & 39.3 \\
            TE+VGGT-$\Omega$ & 30.1 & 45.7 \\
            TE+sensor & \textbf{71.7} & \textbf{46.5} \\
            \bottomrule
        \end{tabular}%
        \end{adjustbox}
    \end{subtable}%
    \hfill
    \begin{subtable}[t]{0.35\textwidth}
        \centering
        \subcaption{Static/dynamic decomposition.}
        \label{subtab:static_dynamic}
        \begin{adjustbox}{max width=\linewidth,center}
        \begin{tabular}{@{}lccc@{}}
            \toprule
            & APD-P & s/fr$\downarrow$ & Mem.$\downarrow$ \\
            \midrule
            TE & \textbf{31.2} & \textbf{0.05} & \textbf{9.1\,GB} \\
            $-$ static/dyn cls & \textbf{31.2} & 0.24 & 22.4\,GB \\
            \bottomrule
        \end{tabular}%
        \end{adjustbox}
    \end{subtable}%
    \hfill
    \begin{subtable}[t]{0.24\textwidth}
        \centering
        \subcaption{Model design.}
        \label{subtab:ablation_design}
        \begin{adjustbox}{max width=\linewidth,center}
        \begin{tabular}{@{}lc@{}}
            \toprule
            & APD-P \\
            \midrule
            Full model & \textbf{31.2} \\
            $-$ WAFT & 29.3 \\
            $-$ iter.\ refin. & 26.4 \\
            \bottomrule
        \end{tabular}%
        \end{adjustbox}
    \end{subtable}
    \vspace{0.6em}

    \begin{subtable}[t]{0.58\textwidth}
        \centering
        \subcaption{Voxel size.}
        \label{subtab:ablation_voxel}
        \begin{adjustbox}{max width=\linewidth,center}
        \begin{tabular}{@{}lccccccc@{}}
            \toprule
            & no vox & 0.005 & 0.01 & 0.02 & 0.05 & 0.1 & 0.2 \\
            \midrule
            APD-P & -- & \textbf{31.4} & 31.3 & 30.9 & 28.5 & 25.8 & 23.5 \\
            s/fr$\downarrow$ & -- & 0.23 & 0.18 & 0.11 & 0.04 & 0.04 & \textbf{0.03} \\
            Peak mem.$\downarrow$ (GB) & OOM & 24.0 & 21.0 & 15.0 & 11.2 & 10.8 & \textbf{10.7} \\
            \bottomrule
        \end{tabular}%
        \end{adjustbox}
    \end{subtable}%
    \hfill
    \begin{subtable}[t]{0.36\textwidth}
        \centering
        \subcaption{Window length.}
        \label{subtab:ablation_window}
        \begin{adjustbox}{max width=\linewidth,center}
        \begin{tabular}{@{}lccc@{}}
            \toprule
            $L$ & APD-P & s/fr$\downarrow$ & Mem.$\downarrow$ \\
            \midrule
            8 & \textbf{31.3} & \textbf{0.04} & \textbf{8.1\,GB} \\
            16 & 31.2 & 0.05 & 9.1\,GB \\
            24 & 29.8 & 0.06 & 11.3\,GB \\
            \bottomrule
        \end{tabular}%
        \end{adjustbox}
    \end{subtable}
    \vspace{-0.08in}
\end{table*}

\noindent \textbf{Geometry Source.}
\Cref{tab:ablation_studies}(\subref{subtab:sensor_geometry}) evaluates \model across different input pointmap sources without any retraining.
Replacing Pi3 with VGGT-$\Omega$ improves APD-P by $+6.6$ on PStudio ($30.1$) and $+6.4$ on ADT ($45.7$). Sensor geometry further boosts accuracy to $71.7$ on PStudio and $46.5$ on ADT, surpassing D4RT ($49.6$ and $40.8$) despite using ${\sim}20\times$ fewer parameters.
Crucially, while monolithic trackers like D4RT cannot ingest external geometry, \model seamlessly inherits improvements in geometry backbones and sensor hardware without retraining.

\noindent \textbf{Static/dynamic decomposition.}
\Cref{tab:ablation_studies}(\subref{subtab:static_dynamic}) evaluates the static/dynamic decomposition. The classifier reliably segments moving objects ($93.1$ and $92.5$ F1 on Dynamic Replica and PointOdyssey). Forcing all points through the dynamic trajectory decoder leaves tracking accuracy identical ($31.2$ APD-P) but increases latency by $4.8\times$ and peak GPU memory by $2.5\times$, confirming that concentrating compute on dynamic surfaces provides substantial efficiency gains without accuracy loss.

\noindent \textbf{Model design choices.}
\Cref{tab:ablation_studies}(\subref{subtab:ablation_design}) ablates key architectural components. Removing iterative trajectory refinement reduces APD-P from $31.2$ to $26.4$ ($-4.8$). Removing 3D WAFT feature sampling drops APD-P to $29.3$ ($-1.9$). We include qualitatives for these ablations in the supplementary.

\noindent \textbf{Voxel size.}
\Cref{tab:ablation_studies}(\subref{subtab:ablation_voxel}) sweeps the de-duplication voxel size $v$ on PointOdyssey.
Without voxelization, the model quickly runs out of memory (OOM) as it retains as many tokens as 2D representations, highlighting the necessity of de-duplication for tracking all points across all frames. 
Tracking accuracy remains stable from a fine voxel size of $0.005$ up to $0.02$ ($\sim$31 APD-P), dropping thereafter.
Coarser voxels increase inference speed and reduce peak memory  (from $0.23$\,s/frame and $24.0$\,GB at $0.005$ to $0.03$\,s and $10.7$\,GB at $0.2$).
We adopt $v{=}0.02$ as the default,  an effective trade-off that retains peak accuracy while halving latency and memory.

\noindent \textbf{Window length.}
\Cref{tab:ablation_studies}(\subref{subtab:ablation_window}) ablates the window length $L \in \{8, 16, 24\}$ on TapVid-3D. $L{=}8$ and $L{=}16$ achieve similar APD-P ($31.3$ vs.\ $31.2$), while $L{=}24$ drops to $29.8$. Although $L{=}8$ has lower test latency, training with $L{=}16$ is $26\%$ faster per step. 
We therefore set $L{=}16$.

\ifpaperfinal
\noindent \textbf{Qualitative Results}
\newcommand{\abcols}{3}
\newlength{\absep}\setlength{\absep}{1.5pt}
\newlength{\ablabw}\setlength{\ablabw}{12pt}
\newlength{\abgap}\setlength{\abgap}{3pt}
\newlength{\abw}
\setlength{\abw}{\dimexpr(\textwidth-\ablabw-\abgap-2\absep)/\abcols\relax}
\newcommand{\ablabel}[1]{%
  \makebox[\ablabw][c]{%
    \raisebox{\dimexpr0.3772\abw-0.5\height\relax}{\rotatebox{90}{\scriptsize #1}}}}

\ifpaperfinal
\begin{figure*}[t]
\else
\begin{figure}[H]
\fi
  \centering
  \setlength{\tabcolsep}{0pt}
  \begin{tabular}{@{}c@{\hspace{\abgap}}c@{\hspace{\absep}}c@{\hspace{\absep}}c@{}}
    \ablabel{Ours} %
      & \includegraphics[width=\abw]{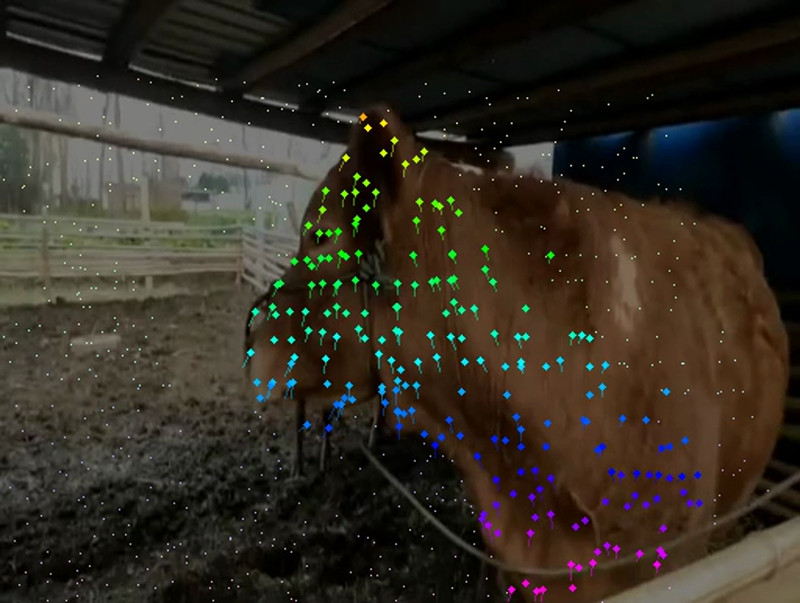}
      & \includegraphics[width=\abw]{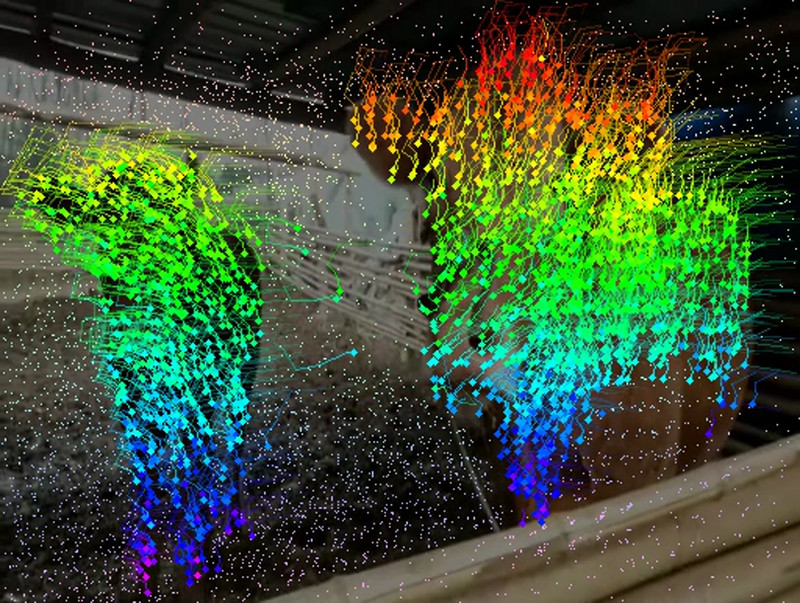}
      & \includegraphics[width=\abw]{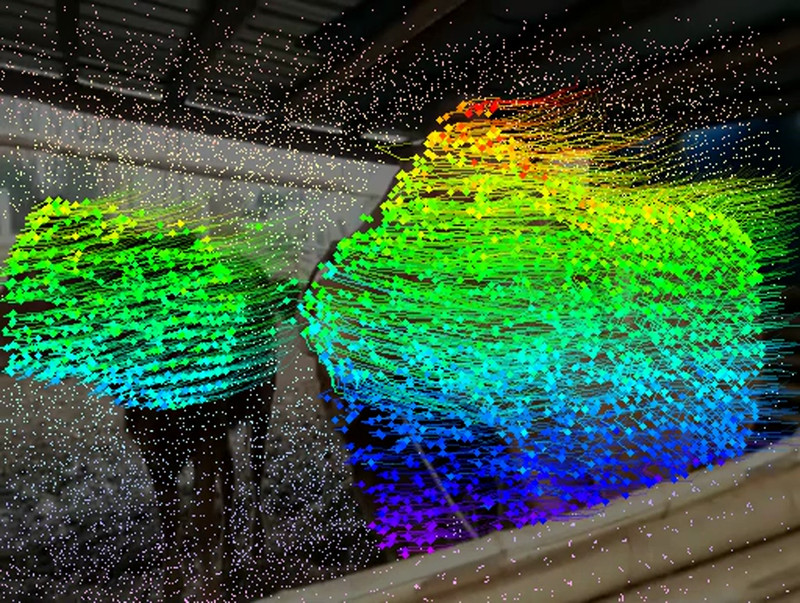} \\[\absep]
    \ablabel{SpaTracker\,V2} %
      & \includegraphics[width=\abw]{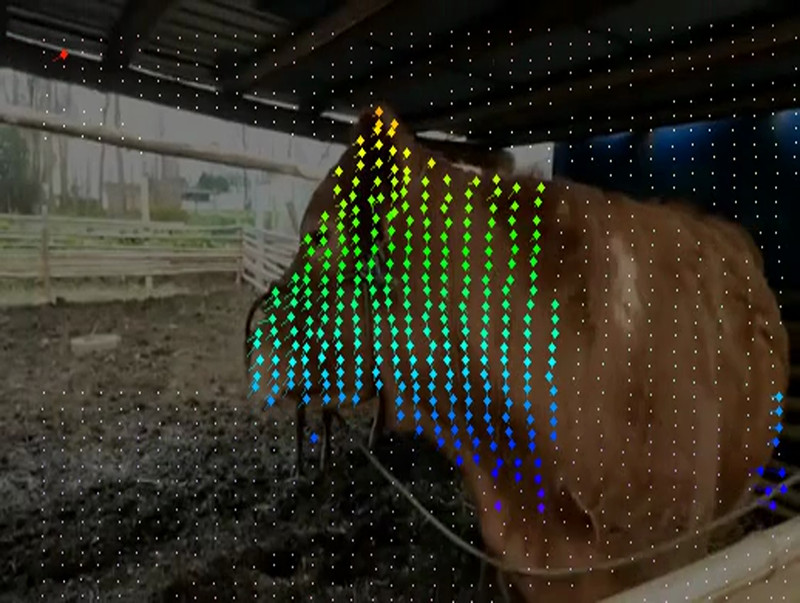}
      & \includegraphics[width=\abw]{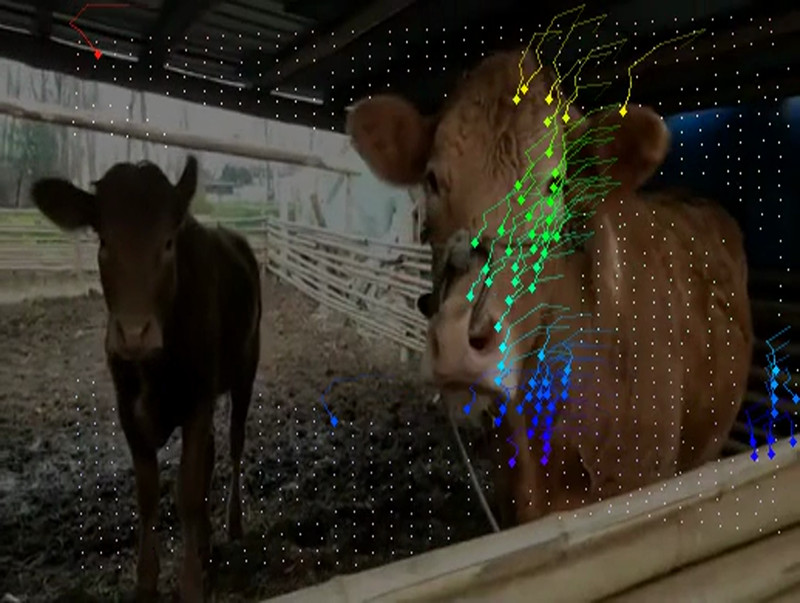}
      & \includegraphics[width=\abw]{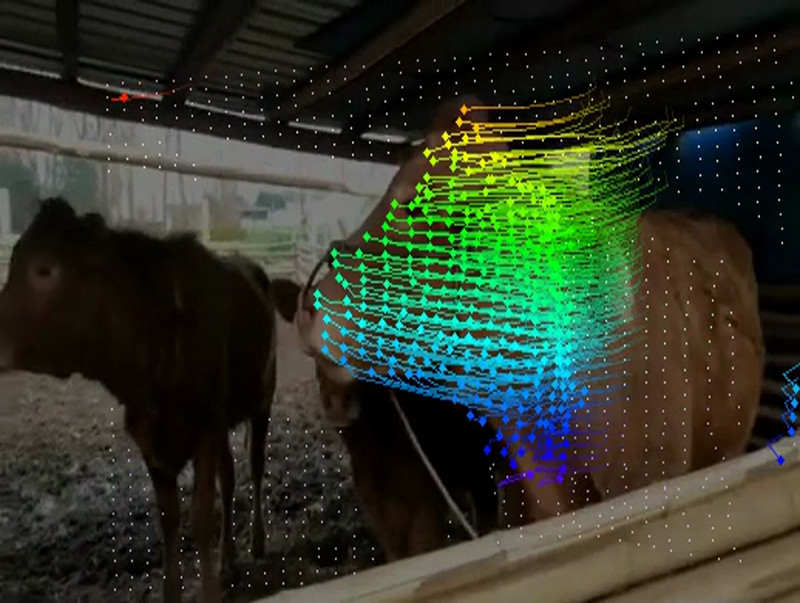} \\[\absep]
    \ablabel{CoTracker3} %
      & \includegraphics[width=\abw]{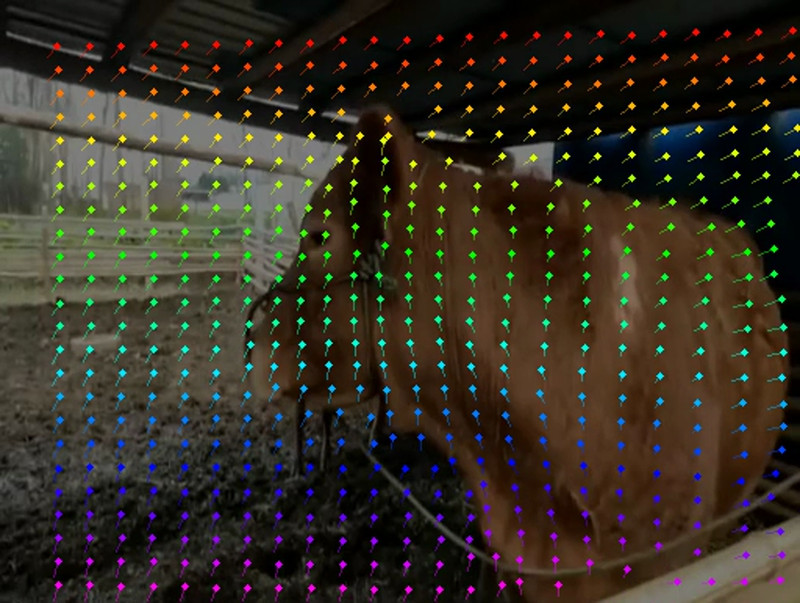}
      & \includegraphics[width=\abw]{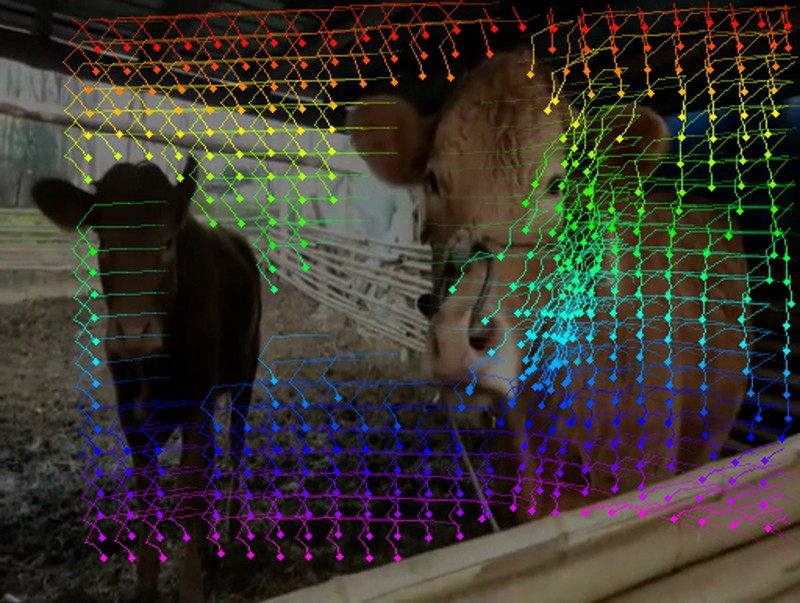}
      & \includegraphics[width=\abw]{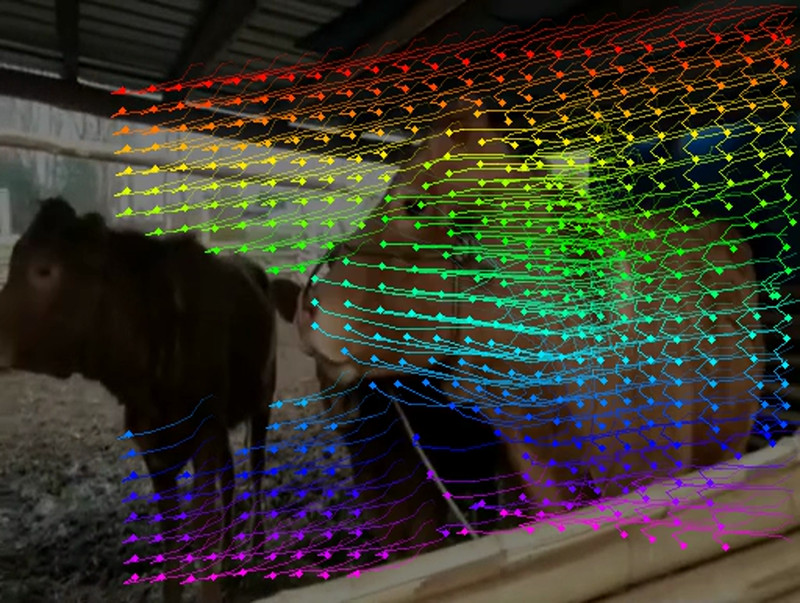} \\[\absep]
    \ablabel{Delta\,V2} %
      & \includegraphics[width=\abw]{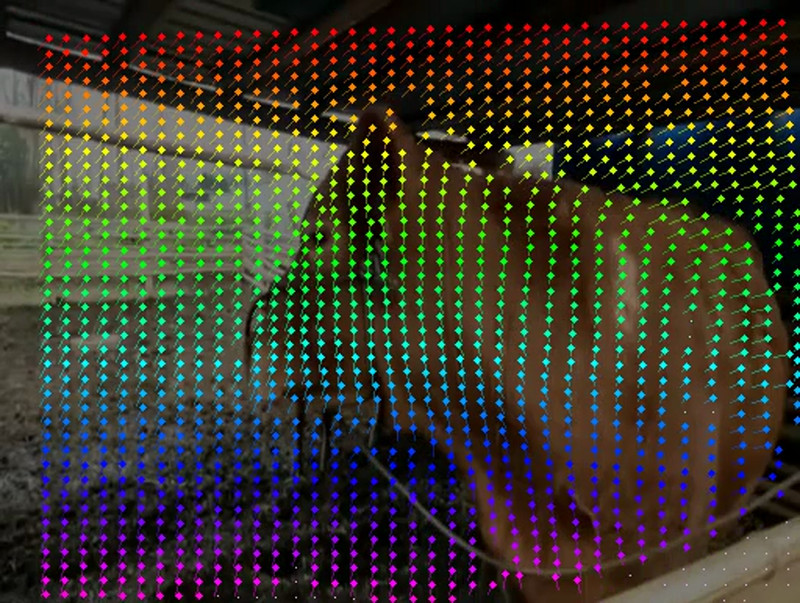}
      & \includegraphics[width=\abw]{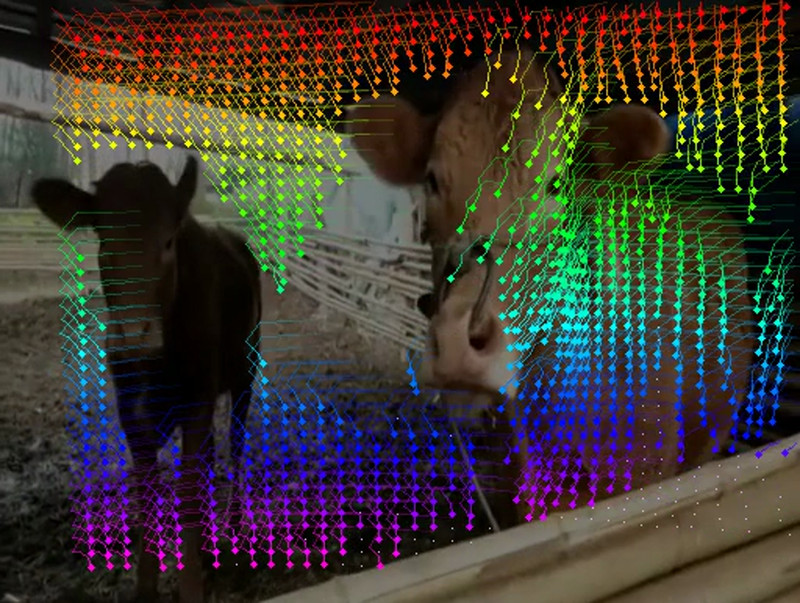}
      & \includegraphics[width=\abw]{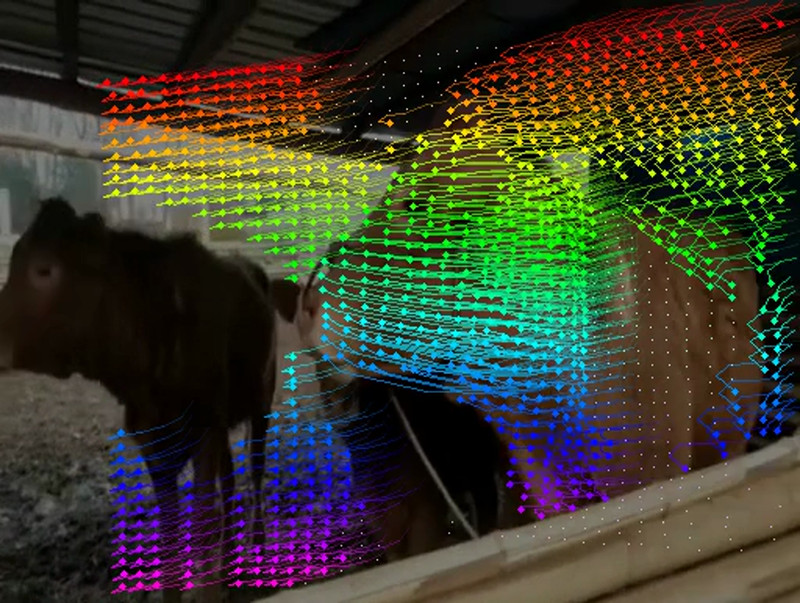} \\
  \end{tabular}
  \caption{\textbf{Qualitative comparison against baselines.}
  \model tracks every visible point across time, including newly emerging content. Notice that only one cow is visible in the first frame, while the second cow emerges later. Sparse trackers need human-specified queries and are usually demoed with a first-frame grid; densifying queries over later frames quickly becomes prohibitive (\Cref{fig:all_methods_benchmarking}). First-frame dense trackers have the same blind spot by design. In all visualizations, we show trails for dynamic points only; the method tracks all points and accounts for camera motion, but we omit static-point trails in the 2D renderings for clarity.}
  \label{fig:qualitative_results}
\ifpaperfinal
\end{figure*}
\else
\end{figure}
\fi

\Cref{fig:qualitative_results} compares \model against SpatialTracker-v2, CoTracker3, and DeltaV2.
\model produces dense tracks over all visible points, including regions that appear after the first frame, while baselines either track only queried/first-frame points or leave large untracked areas.
Additional long-video results, static/dynamic visualizations, and qualitative ablations of WAFT and iterative refinement are available on our project page and in the supplementary.
\else
Qualitative comparisons against SpatialTracker-v2, CoTracker3, and DeltaV2 and other qualitative results are in the supplementary (\Cref{fig:qualitative_results}).
\fi

\vspace{-0.1in}
\section{Limitations}
\vspace{-0.06in} 

While \model enables all-frame dense 3D tracking over long videos, limitations remain. First, while voxelized de-duplication bounds memory growth with scene content, scaling to 1000+ frames where prior dense trackers fail beyond $\sim$96 frames, active memory still continues to expand during perpetual open-world exploration; integrating spatial cache eviction for hour-long trajectories is a natural next step. Second, like prior 3D tracking models~\citep{tapip3d,spatialtrackerv2}, \model relies on external geometry. While this modularity allows sensor integration and benefits from ongoing advances in feedforward 3D reconstruction without retraining, performance remains coupled to underlying pointmap fidelity. 
Third, the cross-window voxelization de-duplication mechanism merges tracks purely based on spatial proximity at temporal window boundaries. Because this merging is executed via irreversible mean-pooling, tracking errors can cause distinct physical surfaces to permanently collapse into a single canonical track that cannot be disentangled in subsequent frames. While users can already protect important points from being merged by specifying them through our non-voxelized query-point branch (\Cref{sec:query}), future work to improve the de-duplication mechanism is an important direction. Finally, scaling model capacity and incorporating broader pretraining data is likely to yield further gains on complex real-world dynamics.

\vspace{-0.1in}
\section{Conclusion}
\vspace{-0.06in} 

We presented \model, a 3D point tracker that tracks all visible surfaces across long videos by representing them as persistent scene tracks in world coordinates. Grounded in the principle that complexity should scale with unique physical geometry rather than frame count, \model resolves the computational bottleneck of dense tracking via three key innovations: voxelized de-duplication across sliding windows, an endpoint-then-trajectory decomposition focusing dense decoding on moving surfaces, and 3D WAFT for efficient feature sampling without costly 4D correlations. Empirically, \model is the first tracker capable of tracking all visible points across 1000+ frame sequences within 40\,GB of GPU memory, outperforming open-source all-frame dense baselines on TAPVid-3D by over 20\% APD while matching state-of-the-art sparse trackers on long sequences. By making long-horizon dense 3D tracking practical, \model provides an essential foundation for extending persistent 3D world representations to dynamic environments. 
\ifpaperfinal
\ificlr
\subsubsection*{Acknowledgments}
\else
\section*{Acknowledgments and Disclosure of Funding}
\fi

This material is based upon work supported in part by an NSF Career award, ONR award N00014-23-1-2415, AFOSR Grant FA9550-23-1-0257. Any opinions, findings, and conclusions or recommendations expressed in this material are those of the author(s) and do not necessarily reflect the views of the National Science Foundation.
Ayush Jain was a summer intern at Meta for the initial duration of the project and is supported in part also by the Meta AI Mentorship Fellowship. The authors thank Yash Jangir, Jay Karhade, Nikhil Keetha, Ananya Bal, Gabriel Sarch, Naitik Khandelwal and Brian Yang for their helpful discussions and support.
\fi

\vspace{-0.08in}
\section*{Generative AI Disclosure}
\vspace{-0.06in}
In this work, we used generative AI tools to assist in implementing our method. This was limited to coding assistance from AI coding agents while implementing components of the system. We have not used generative AI tools to design or provide feedback on research methodology or experiments, to support qualitative or thematic data analysis, or to interpret results. Generating synthetic datasets, developing theoretical models or conceptual frameworks, formulating mathematical claims, providing critical ingredients for proving mathematical claims, assisting in the writing of proofs, proposing or refining hypotheses, assisting with translation, and cleaning or reformatting datasets are not applicable to this work. Additionally, we used generative AI tools, to refine the writing, and to restructure tables in the paper. We have reviewed all AI-assisted work. The authors reviewed AI-assisted code before using it. We take responsibility for the final content of this work, including text, claims, and artifacts produced with the aid of generative AI.

\ificlr
  \bibliographystyle{iclr2027_conference}
\else
  \bibliographystyle{plainnat}
\fi
\bibliography{main}



\newpage
\clearpage
\appendix
\raggedbottom
\section{Appendix}

\subsection{Dense Tracking on Dynamic Replica and PointOdyssey}
Because TAPVid-3D provides only sparse query annotations, we additionally evaluate tracking on 48 frame clips of Dynamic Replica and PointOdyssey under estimated (EST) and ground-truth (GT) geometry (\Cref{tab:dr_pod_dense}).
\begin{table}[H]
    \centering
    \caption{\textbf{Tracking Performance on Dynamic Replica and PointOdyssey.}
    DeltaV2 and \model use geometry from Pi3 in EST setups and simulator GT geometry in GT setups.
    All evaluations run on 48 frames.}
    \label{tab:dr_pod_dense}
    \begin{adjustbox}{max width=\textwidth,center}
    \scriptsize
    \setlength{\tabcolsep}{1.2pt}
    \begin{tabular}{@{}l*{5}{c}*{5}{c}*{5}{c}*{5}{c}@{}}
        \toprule
        & \multicolumn{5}{c}{Dynamic Replica (EST)} & \multicolumn{5}{c}{Dynamic Replica (GT)} & \multicolumn{5}{c}{PointOdyssey (EST)} & \multicolumn{5}{c}{PointOdyssey (GT)} \\
        \cmidrule(lr){2-6} \cmidrule(lr){7-11} \cmidrule(lr){12-16} \cmidrule(lr){17-21}
        \textbf{Method} & APD-P & APD-M & AJ & OA & EPE$\downarrow$ & APD-P & APD-M & AJ & OA & EPE$\downarrow$ & APD-P & APD-M & AJ & OA & EPE$\downarrow$ & APD-P & APD-M & AJ & OA & EPE$\downarrow$ \\
        \midrule
        VDPM & 13.1 & 73.5 & 8.1 & 93.1 & 0.23 & -- & -- & -- & -- & -- & 15.8 & \textbf{71.8} & 9.3 & 84.1 & \textbf{0.29} & -- & -- & -- & -- & -- \\
        Any4D & 19.5 & 79.6 & 12.7 & \textbf{93.7} & 0.18 & -- & -- & -- & -- & -- & 10.6 & 62.2 & 5.8 & 81.3 & 0.45 & -- & -- & -- & -- & -- \\
        DeltaV2 & 36.2 & \textbf{86.9} & 25.8 & 88.1 & \textbf{0.14} & 67.3 & 98.4 & 55.0 & 90.0 & 0.03 & 23.8 & 69.9 & 14.9 & 83.0 & 0.88 & 33.2 & 81.0 & 22.2 & 73.1 & 0.23 \\
        \textbf{\model (Ours)} & \textbf{36.6} & \textbf{86.9} & \textbf{26.3} & 91.8 & \textbf{0.14} & \textbf{85.2} & \textbf{99.7} & \textbf{76.5} & \textbf{95.8} & \textbf{0.02} & \textbf{24.9} & 70.2 & \textbf{16.3} & \textbf{84.2} & 0.46 & \textbf{73.2} & \textbf{96.9} & \textbf{59.0} & \textbf{89.2} & \textbf{0.05} \\
        \bottomrule
    \end{tabular}%
    \end{adjustbox}
\end{table}

\model matches or outperforms all baselines on both datasets. Under estimated Pi3 geometry, \model is competitive with DeltaV2 (36.6 vs.\ 36.2 APD-P on Dynamic Replica; 24.9 vs.\ 23.8 on PointOdyssey) while still tracking all points in all frames rather than first-frame points only; with GT geometry the gap widens substantially (85.2 vs.\ 67.3 APD-P on Dynamic Replica; 73.2 vs.\ 33.2 on PointOdyssey).

On full-length 1000+ frame videos of PointOdyssey, and evaluation on all labelled points (10k+), \model achieves APD-P of 31.8, APD-M of 74.5, static/dynamic accurary of 91.2\% (78.5\% dynamic F1, 92.5\% static F1). Since, no existing baselines can run in this setting, we do not report any quantitative comparisons here. 

We also show dense tracking results on diverse real-world scenes on our project page.

\subsection{Inference Speed Breakdown}
\begin{table}[H]
    \centering
    \footnotesize
    \setlength{\tabcolsep}{4pt}
    \captionsetup[subtable]{justification=centering}
    \caption{\textbf{Early merge on full-length TAPVid-3D.}
    APD-P uses VGGT-$\Omega$, as in \Cref{tab:tapvid3d_full}.
    FPS averages are the mean of ADT, DriveTrack, and PStudio.}
    \label{tab:voxep_fps}
    \begin{subtable}{\linewidth}
        \centering
        \subcaption{Accuracy and average speed, with and without early merge.}
        \label{subtab:voxep_apd}
        \begin{tabular}{@{}lcccccc@{}}
            \toprule
            & \multicolumn{4}{c}{APD-P} & \multicolumn{2}{c}{Avg FPS$\uparrow$} \\
            \cmidrule(lr){2-5} \cmidrule(lr){6-7}
            & ADT & DriveTrack & PStudio & Avg
            & w/ VGGT-$\Omega$ & tracker \\
            \midrule
            w/o early merge & 40.0 & \textbf{26.5} & 27.2 & 31.2 & 5.6 & 12.6 \\
            w/ early merge & \textbf{40.1} & 26.4 & \textbf{27.3} & \textbf{31.3} & \textbf{7.2} & \textbf{23.1} \\
            \bottomrule
        \end{tabular}
    \end{subtable}

    \vspace{0.6em}

    \begin{subtable}{0.72\linewidth}
        \centering
        \subcaption{FPS with early merge.}
        \label{subtab:voxep_fps_only}
        \begin{tabular}{@{}lcccc@{}}
            \toprule
            & ADT & DriveTrack & PStudio & Avg \\
            \midrule
            w/ VGGT-$\Omega$ & 4.5 & 8.7 & 8.4 & 7.2 \\
            tracker only & \textbf{8.4} & \textbf{19.5} & \textbf{41.5} & \textbf{23.1} \\
            \bottomrule
        \end{tabular}
    \end{subtable}
\end{table}

\paragraph{Early Merge at Endpoints After First Iteration.}
While point de-duplication is primarily performed at sliding-window boundaries, our architecture supports an optional early-merge step (\texttt{voxelize\_on\_endpoints}) immediately following iteration~0 of the endpoint refiner.
Because static/dynamic classification converges within a single iteration, points classified as static that project to the same 3D spatial voxel at their predicted destinations are de-duplicated early.
This aggressively prunes redundant static tokens before iterations $k \ge 1$ and completely bypasses trajectory refinement for these points.
Crucially, dynamic points and evaluated query cells are preserved 1:1, ensuring moving surface tracks are not conflated prior to full within-window trajectory decoding.

As shown in \Cref{tab:voxep_fps}, this static-only early merge acts as a zero-cost acceleration: on full-length TAPVid-3D, tracking accuracy remains unchanged ($31.3$ vs.\ $31.2$ APD-P) while providing substantial speedups (\Cref{subtab:voxep_apd}).
Average tracker-only throughput nearly doubles from $12.6$ to $23.1$\,FPS ($1.8\times$), and end-to-end throughput (including feedforward VGGT-$\Omega$ geometry estimation) rises from $5.6$ to $7.2$\,FPS ($1.3\times$).
In settings where geometry is provided directly by onboard sensors or precomputed pointmaps, only the tracker runs, operating at $23.1$\,FPS on average (\Cref{subtab:voxep_fps_only}).
Per-dataset throughput directly reflects physical scene extent: focused lab scenes like PStudio run at speeds of $41.5$\,FPS (and $19.5$\,FPS on DriveTrack), whereas expansive multi-room environments in ADT run at $8.4$\,FPS due to higher unique surface accumulation.


\subsection{Training Data and Model Sizes}
\model is trained on Kubric~\citep{kubric}, PointOdyssey~\citep{pod}, and Dynamic Replica~\citep{dr}, and has 61M parameters (with a frozen 20M DINOv3-small backbone). SpatialTracker-v2 is trained on these three plus 14 additional datasets. Any4D is trained on these three, Waymo-DriveTrack, and 4 more datasets. VDPM initializes from a VGGT backbone and trains on Kubric, PointOdyssey, Waymo, and 2 static datasets. TAPIP-3D and DeltaV2 are trained only on Kubric. CoTracker3 is trained on Kubric and on additional 15k real-world videos. D4RT is trained on all of the above, 8 additional public datasets, and undisclosed internal datasets, with a 1B-parameter backbone and 144M-parameter decoder. Please check the respective papers for more details.

\subsection{Additional Method and Implementation Details}

\begin{figure}[H]
  \centering
  \includegraphics[width=\textwidth]{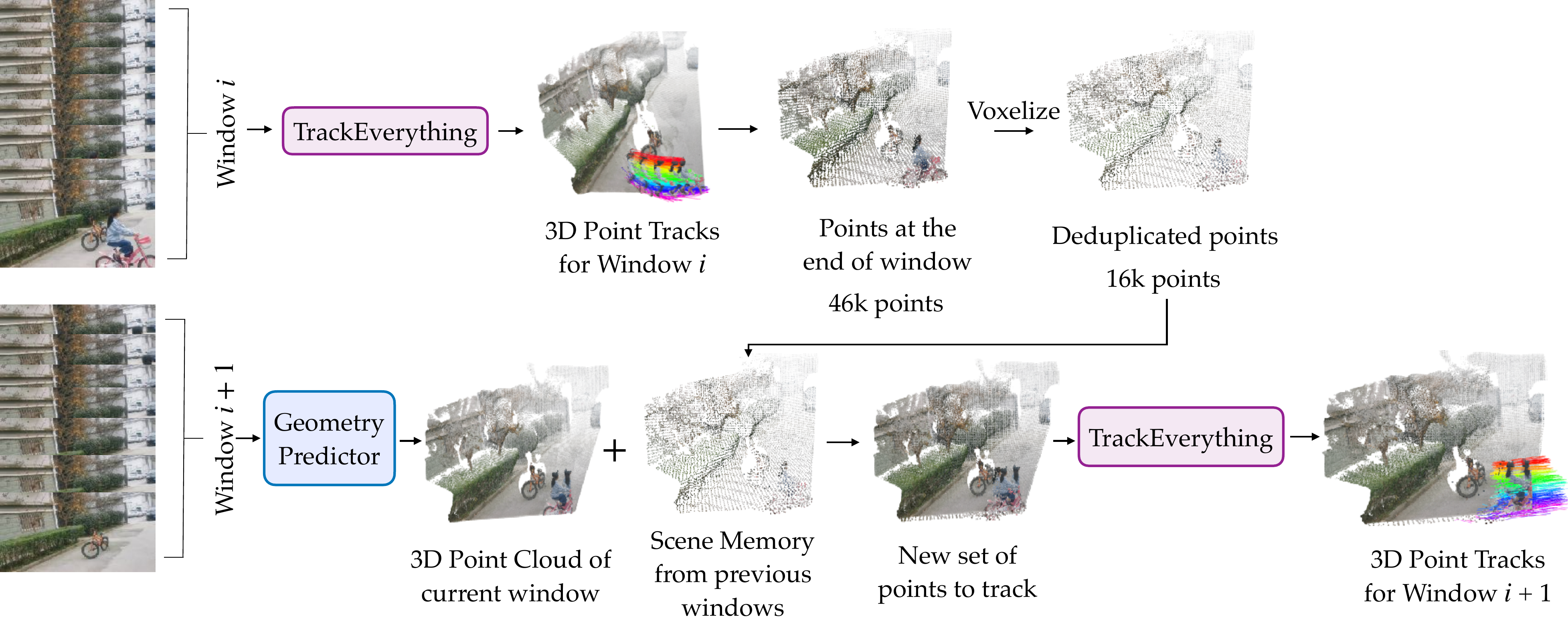}
  \caption{\textbf{Sliding-window tracking and cross-window de-duplication pipeline.}
  \textit{Top row (Window $i$):} A temporal window of frames is processed by \model to predict dense 3D tracks across all points.
  At the final timestep of the window, the predicted 3D point locations (here 46k points) are voxelized by quantizing coordinates into spatial voxels and mean-pooling co-located tokens. This cross-frame de-duplication prunes redundant surface observations (reducing point count from 46k to 16k).
  \textit{Bottom row (Window $i+1$):} When the subsequent frame window arrives, its frames are lifted into a 3D point cloud using a feedforward geometry predictor (or sensor depths).
  These newly observed points are concatenated with the de-duplicated points carried over from previous windows (serving as persistent 3D scene memory) to construct the active set of points to track.
  \model then produces dense 3D trajectories for all points across all timesteps of Window $i+1$.
  By chaining sliding-window processing with boundary de-duplication, \model tracks extended video sequences while bounding active token counts to unique physical scene geometry rather than video duration.
  }
  \label{fig:deduplication_sliding_window}
\end{figure}

\begin{figure}[H]
  \centering
  \includegraphics[width=0.9\textwidth]{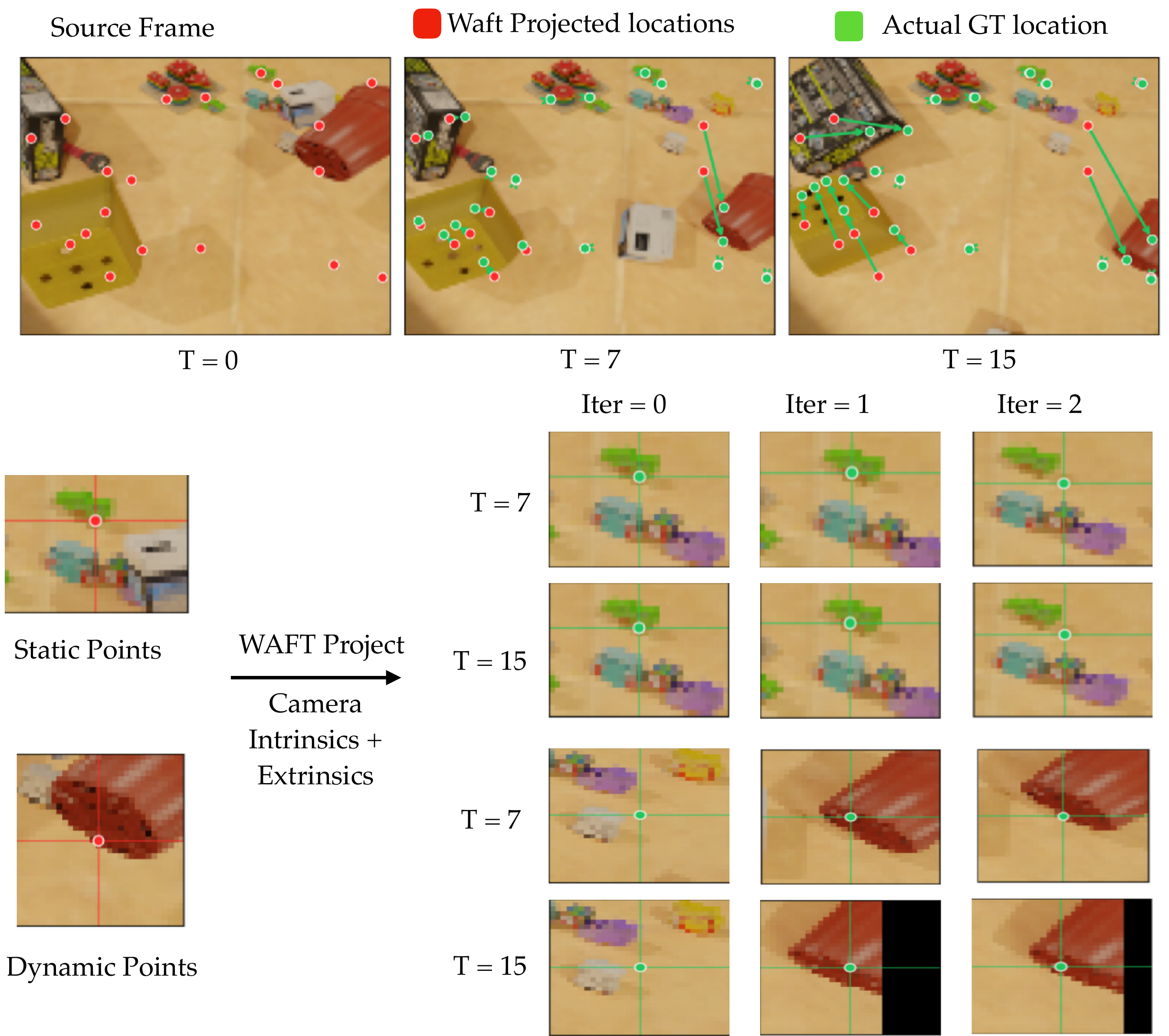}
  \caption{\textbf{3D WAFT projection and iterative refinement dynamics.}
  \textit{Top:} Projections across time under zero-velocity initialization. Source points initialized at frame $T=0$ (red dots) are projected to future timesteps ($T=7$ and $T=15$) using camera extrinsics and intrinsics. Red dots denote projected 2D locations under the zero-velocity prior ($\mathbf{x}_{\mathrm{tgt}}^{(0)} = \mathbf{x}_{\mathrm{src}}$ in world space); green dots mark true ground-truth (GT) locations, with green arrows at $T=15$ indicating displacement error. For static objects, camera projection accounts for ego-motion and aligns projected points near GT; for dynamic objects, world motion leads to significant spatial offset.
  \textit{Bottom:} Zoomed-in patches across refinement iterations ($\text{Iter} \in \{0, 1, 2\}$) at $T=7$ and $T=15$.
  \textit{Static point (top):} The zero-velocity initialization at $\text{Iter}=0$ already places the sampled point close to the true location. The bilinearly sampled target feature $\mathbf{g}_{\mathrm{tgt}}$ closely matches the source feature $\mathbf{g}_{\mathrm{src}}$, signaling to the model that no large displacement is required and maintaining stability across iterations 1 and 2.
  \textit{Dynamic point (bottom):} At $\text{Iter}=0$, the zero-velocity projection misses the moving object by multiple pixels (landing on the background), creating an appearance mismatch between $\mathbf{g}_{\mathrm{src}}$ and $\mathbf{g}_{\mathrm{tgt}}$ in the concatenated WAFT feature.
  Driven by this mismatch, the endpoint refiner iteratively updates 3D positions: by $\text{Iter}=1$ the point moves substantially closer, and by $\text{Iter}=2$ the sampled point locks onto the correct surface feature.
  }
  \label{fig:waft_refinement}
\end{figure}

\paragraph{Visual Illustrations.}
We visually illustrate two components of our model, sliding-window de-duplication and 3D WAFT, in \Cref{fig:deduplication_sliding_window,fig:waft_refinement}.

\paragraph{Track Lineage and Historical Trajectory Reconstruction.}
While cross-window voxelization bounds active GPU memory by collapsing co-located points, full historical trajectories $\mathbf{X} \in \mathbb{R}^{N \times T \times 3}$ for all original surface points across the video can be reconstructed.
During streaming inference, each voxelization step records an integer point-to-voxel mapping array $\mathbf{p} \in \mathbb{N}^{N_{\mathrm{pre}}}$ and a spatial offset residual $\mathbf{r}_i = \mathbf{x}_i - \mathbf{x}_{\lfloor \mathbf{x}_i / v \rfloor} \in \mathbb{R}^3$.
When multiple observations merge into a single active voxel token, birth metadata is preserved by prioritizing the earliest observation: the source birth timestep $t_{\mathrm{src}}$ and initial unprojected coordinate $\mathbf{x}_{\mathrm{src}}$ are retained via an \texttt{argmin} reduction over the merged cluster.
To export full dense trajectories for all original pixels $(t, h, w)$, the model performs an exact inverse unvoxelization by traversing the sequence of chained point-to-voxel mappings in reverse and restoring the stored spatial residuals:
\begin{equation}
    \mathbf{X}_{\mathrm{orig}}(t) = \mathbf{X}_{\mathrm{vox}}[\mathbf{p}](t) + \mathbf{r}.
\end{equation}
Because the spatial residual $\mathbf{r}$ broadcasts across all timesteps, tracking fidelity is preserved for individual surface elements.
Crucially, this design decouples active tracking state from image features: once a sliding window finishes processing, its large 2D image feature maps are immediately released from GPU memory, retaining only the compact 3D point tracks and the 3D to 2D mapping tables.

\paragraph{Static/Dynamic Ground-Truth Formulation.}
We derive binary static/dynamic ground truth labels $y_{\mathrm{dyn}} \in \{0, 1\}$ from the world-space displacement of each trajectory over the temporal window $[t_{\mathrm{start}}, t_{\mathrm{end}}]$.
For each track, we evaluate the spatial extent across its annotated frames $\mathcal{V} \subseteq [t_{\mathrm{start}}, t_{\mathrm{end}}]$ as the L2 norm of its axis-aligned bounding box:
\begin{equation}
    \Delta_{\mathrm{motion}} = \left\| \max_{t \in \mathcal{V}} \mathbf{x}(t) - \min_{t \in \mathcal{V}} \mathbf{x}(t) \right\|_2.
\end{equation}
A trajectory is labeled dynamic ($y_{\mathrm{dyn}} = 1$) if $\Delta_{\mathrm{motion}} > \tau_{\mathrm{dyn}}$ and static ($y_{\mathrm{dyn}} = 0$) otherwise, where $\tau_{\mathrm{dyn}} = 0.05$ in normalized scene units (or $0.10$ for unnormalized synthetic clips).

\paragraph{Static/Dynamic Classifier Robustness and Error Dynamics.}
Our static/dynamic decomposition relies on a binary classifier at the endpoint stage. The classifier is designed such that edge-case misclassifications degrade gracefully. Specifically, if a truly static surface point is erroneously predicted as dynamic, it passes into the trajectory refiner. Because the trajectory refiner is trained on arbitrary 3D motions, including stationary points, it can simply predict a zero-displacement trajectory, leaving tracking accuracy unaffected at the expense of marginal extra compute for that token. Conversely, if a dynamic point is erroneously classified as static within a window, its trajectory is held stationary for that window duration; upon entering subsequent sliding windows, its motion state is re-evaluated, allowing it to resume dynamic tracking once the displacement becomes prominent.

\paragraph{3D WAFT Projection and Boundary Conditions.}
The 3D Warp-Aligned Feature Transform (3D WAFT) samples appearance evidence by projecting 3D candidate positions onto 2D feature maps. Camera extrinsics and intrinsics are either obtained from onboard sensors or estimated via feedforward geometry backbones~\citep{vggt_omega,pi3}.
When points project outside the visible image frame during rapid camera ego-motion, coordinates are clamped within an extended image canvas $[-2W, 2W] \times [-2H, 2H]$ and bilinearly sampled using boundary-replicate padding.
Points that are occluded or remain outside the true field of view produce an appearance mismatch between the source feature $\mathbf{g}_{\mathrm{src}}$ and target sample $\mathbf{g}_{\mathrm{tgt}}$, which provides an explicit signal for the transformer's visibility head to predict low visibility logits.

\paragraph{Coordinate System and Voxelization Details}
All 3D scene points and camera poses are expressed in a unified world coordinate frame anchored to the initial camera ($\mathrm{cam}_0$, with pose $[\mathbf{I} \mid \mathbf{0}]$).
Following DUSt3R~\citep{dust3r}, the coordinate system is normalized by a global scale factor defined once at the sequence level, ensuring that the metric scale and spatial origin remain strictly consistent throughout the video.
As consecutive sliding windows stream in, newly observed surfaces are unprojected directly into this existing, shared $\mathrm{cam}_0$ world coordinate frame.

Within this persistent world space, voxelization assigns each 3D point $\mathbf{x} \in \mathbb{R}^3$ an integer lattice coordinate by quantizing with a fixed edge length $v$ (default $0.02$):
\begin{equation}
    \mathbf{k} = \lfloor \mathbf{x} / v \rfloor \in \mathbb{Z}^3.
\end{equation}
Tokens that share identical integer lattice coordinates are merged via mean-pooling.
Because voxelization is implemented via sparse hashing over the discrete integer coordinates $\mathbf{k}$, it dynamically allocates indices only for occupied cells rather than discretizing a bounded volumetric grid.
Consequently, our representation naturally supports arbitrary, expanding scenes without requiring spatial bounds, coordinate clipping, or truncation.

\paragraph{Optimization Mechanics and Gradient Flow.}
To train our model across extended sequences without running out of GPU memory, we avoid backpropagating through time across sliding windows while still allowing the 2D visual encoder (the ViT Adapter) to receive supervisory signal across the entire video.
Specifically, 2D feature maps $\mathbf{F} = \mathcal{E}(\mathbf{I})$ are extracted across the sequence and wrapped in a detached leaf tensor $\mathbf{F}_{\mathrm{param}}$ with gradients enabled.
During sliding-window execution, forward and backward passes are evaluated window-by-window with recurrent state detached at window boundaries.
Within each window, the combined loss:
\begin{equation}
    \mathcal{L} = \mathcal{L}_{\mathrm{coord}} + \lambda_{\mathrm{vis}}\mathcal{L}_{\mathrm{vis}} + \lambda_{\mathrm{dyn}}\mathcal{L}_{\mathrm{dyn}} + \lambda_{\mathrm{conf}}\mathcal{L}_{\mathrm{conf}}
\end{equation}
is computed, and a local backward pass immediately accumulates gradients into the parameter gradient buffer $\mathbf{F}_{\mathrm{param}}.\mathrm{grad}$ and the tracker modules.
This releases intermediate activations after each window.
Once all windows have completed, a single backward pass is executed through the 2D visual backbone via the scalar inner product $\langle \mathbf{F}_{\mathrm{param}}.\mathrm{grad}, \mathbf{F} \rangle$, propagating the accumulated visual gradients into the trainable ViT Adapter head in a single step before calling \texttt{optimizer.step()}.

\noindent \textbf{Fair comparisons.} 
Reproducing sparse baselines on TAPVid-3D has proven difficult due to the lack of standardized  evaluation code. Some prior dense 3D tracking papers~\citep{vdpm} do not compare against sparse trackers at all. We carefully re-evaluate all baselines under a unified protocol and find, for example, that SpatialTracker-v2~\citep{spatialtrackerv2} achieves over 20\% higher APD-M than what prior work~\citep{any4d} reports for it on TAPVid-3D.

\subsection{Fixes to the PointOdyssey Dataset}
During our training runs, we observed significant loss spikes, which we could trace back to the PointOdyssey dataset. We identified several critical annotation errors and developed a programmatic correction pipeline:
\begin{enumerate}
    \item \emph{Off-by-One Camera Extrinsics:} Across 16 sequences (about 25k frames), the recorded extrinsic $\mathbf{T}_t$ is the pose that rendered frame $t-1$, while the 3D tracks are indexed correctly. We reindex the camera to frame $t+1$ and recompute the 2D tracks from that pose. The shift is a few pixels when the camera is slow and exceeds $600\,\mathrm{px}$ during fast pans (up to $59^\circ/\mathrm{frame}$).
    \item \emph{Corrupted Visibility Flags:} Visibility was computed from the uncorrected pose, so the stored flags are wrong where the two poses disagree. We do not rewrite them. Point supervision is dropped on the 502 frames where the repair moves tracks by at least $8\,\mathrm{px}$ ($2\%$ of the affected sequences; at most $3.8\%$ of any one sequence). The images stay in the clip, and the camera reindexing above still supervises the remaining frames.
    \item \emph{Temporally Incoherent Sequences:} The sequence \texttt{scene\_recording\_20210910\_S05\_S06\_0\_ego1} exhibits sub-frame camera jitter that cannot be resolved by an integer frame shift; we exclude it from training and evaluation splits.
\end{enumerate}
We also briefly explored using object segmentation masks to remove wrong point track labels by using the fact that a point born within the segmentation mask of an object should always stay within the mask. However, we found that the segmentation mask had bugs too, which makes this approach infeasible. 

We will release the corrections to the PointOdyssey dataset with our codebase. With these fixes, we noticed reduced loss spikes. We note, however, that some noisy points still remain in the dataset, which we could not fix.

\subsection{Qualitatives}
\begin{figure}[H]
  \centering
  \includegraphics[width=\textwidth]{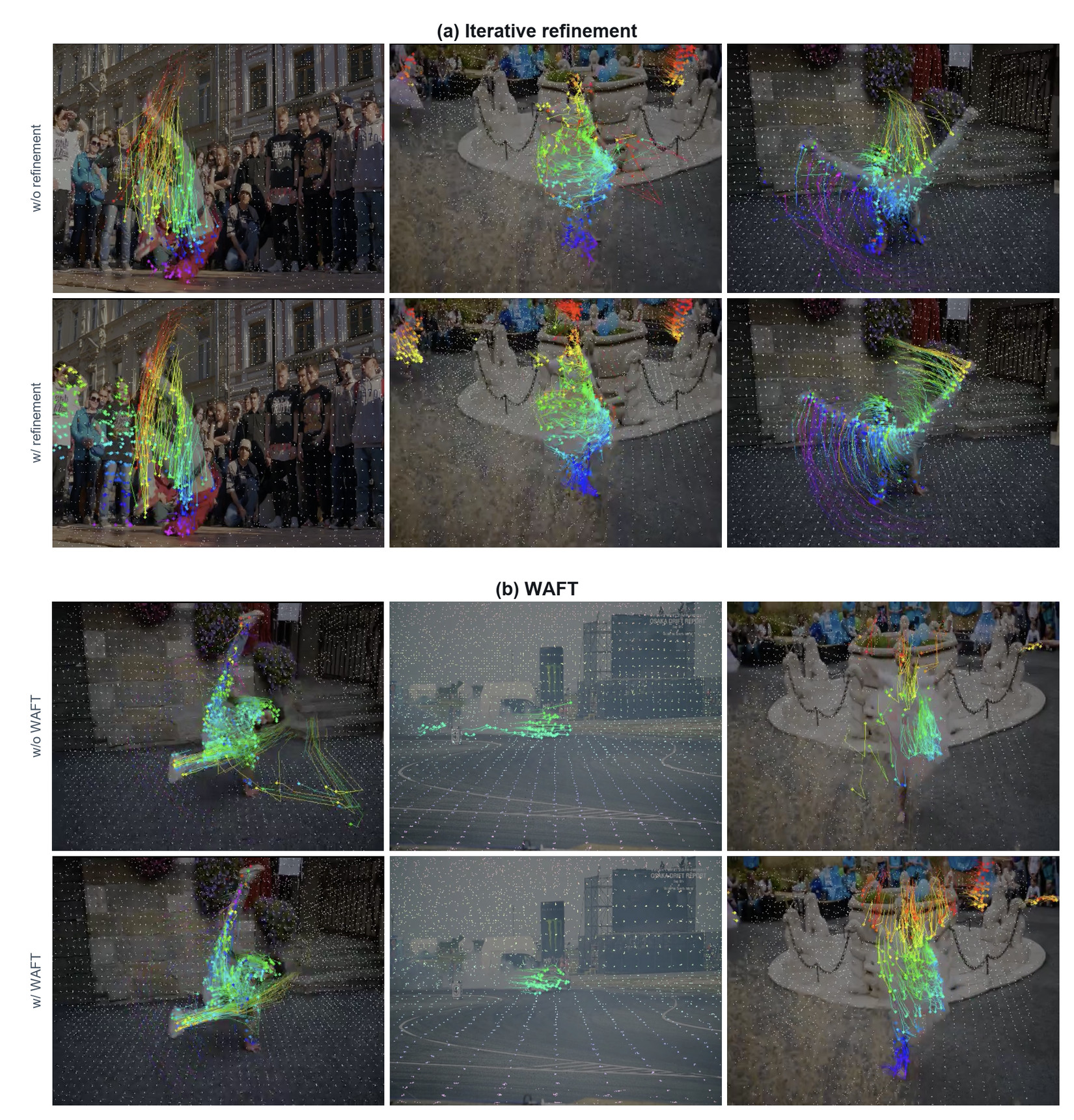}
  \caption{\textbf{Qualitative ablations of iterative refinement and WAFT.}
  (a)~Without iterative refinement, tracks are noisier and less coherent; with refinement, trajectories follow motion more cleanly.
  (b)~Without WAFT, tracks drift with long stray trails on fast motion; with WAFT, trajectories remain tightly localized on the moving subject.}
  \label{fig:ablation_combined}
\end{figure}

\ifpaperfinal\else

\Cref{fig:qualitative_results} compares \model against SpatialTracker-v2, CoTracker3, and DeltaV2.
\model produces dense tracks over all visible points, including regions that appear after the first frame, while baselines either track only queried/first-frame points or leave large untracked areas.

\fi
We show additional qualitative comparisons against baselines on Breakdance, Dance-jump, and a hand-manipulation sequence (\Cref{fig:qualitative_results_supp}), qualitative ablations of iterative refinement and WAFT (\Cref{fig:ablation_combined}), and static/dynamic classification visualizations (\Cref{fig:static_dynamic}).
Without iterative refinement or WAFT, tracks fragment or drift on fast motion, while enabling each recovers coherent long trails.
\Cref{fig:static_dynamic} shows per-point static/dynamic labels over time: dynamic points in red, static in blue.


\setlength{\abw}{\dimexpr(\textwidth-\ablabw-\abgap-2\absep)/\abcols\relax}

\begin{figure}[H]
  \centering
  \noindent
  \setlength{\tabcolsep}{0pt}%
  \begin{tabular}{@{}c@{\hspace{\abgap}}c@{\hspace{\absep}}c@{\hspace{\absep}}c@{}}
    & \multicolumn{3}{c}{\small\bfseries Breakdance} \\[4pt]
    \ablabel{Ours} %
      & \includegraphics[width=\abw]{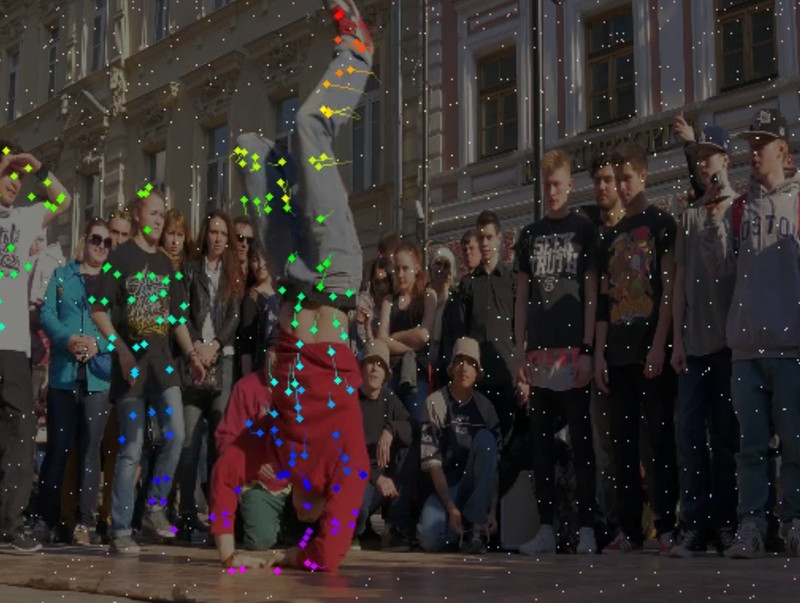}
      & \includegraphics[width=\abw]{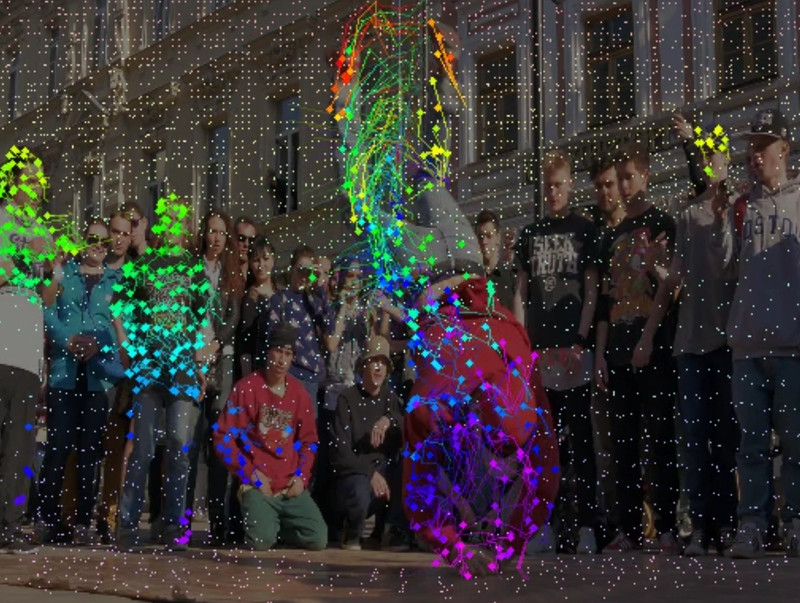}
      & \includegraphics[width=\abw]{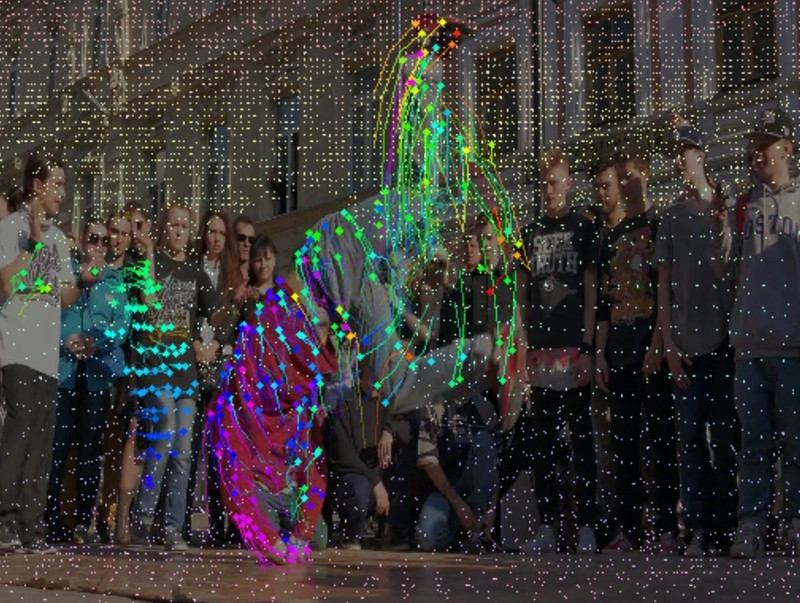} \\[\absep]
    \ablabel{SpaTracker\,V2} %
      & \includegraphics[width=\abw]{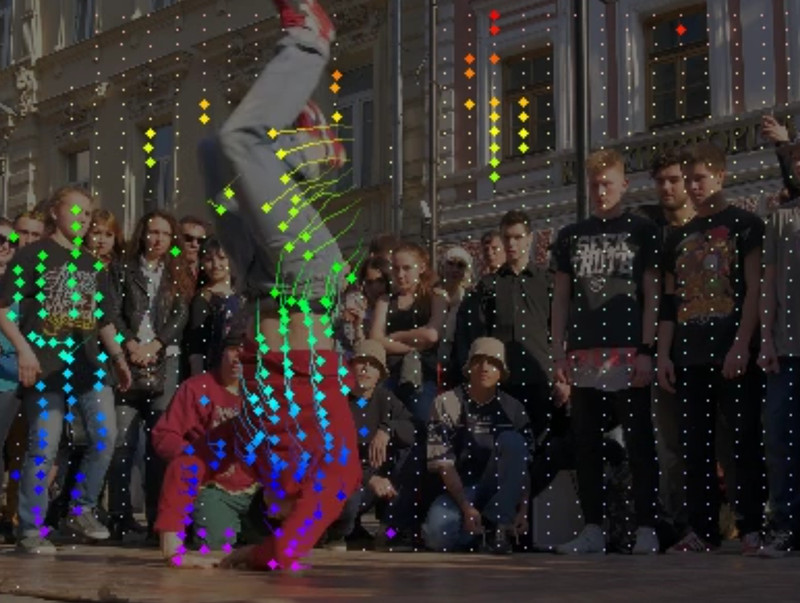}
      & \includegraphics[width=\abw]{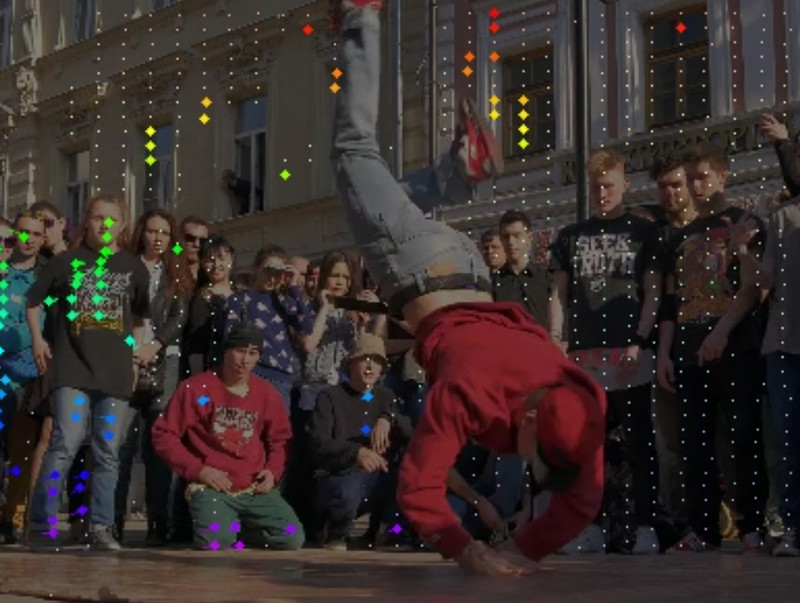}
      & \includegraphics[width=\abw]{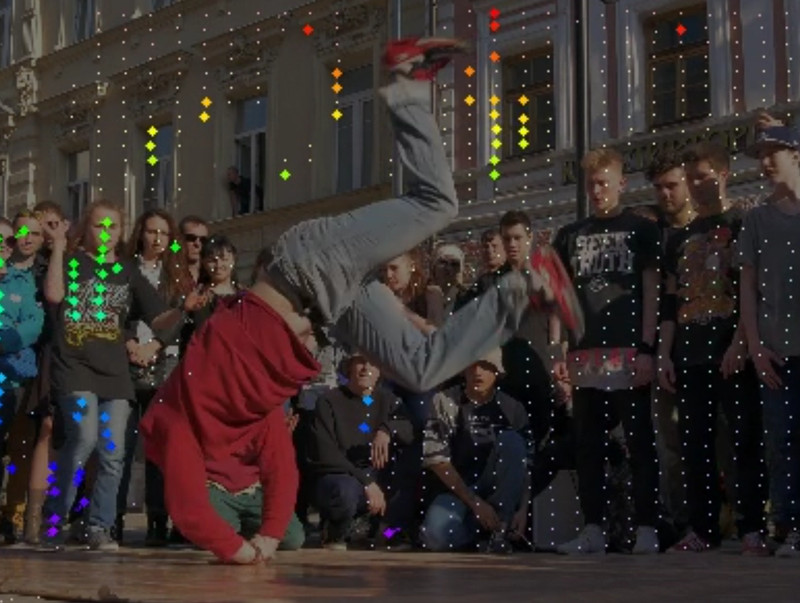} \\[\absep]
    \ablabel{CoTracker3} %
      & \includegraphics[width=\abw]{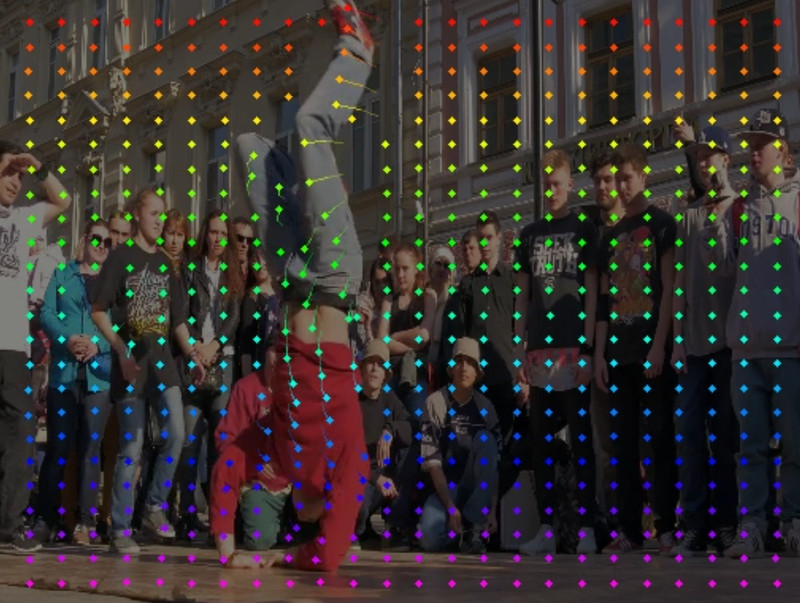}
      & \includegraphics[width=\abw]{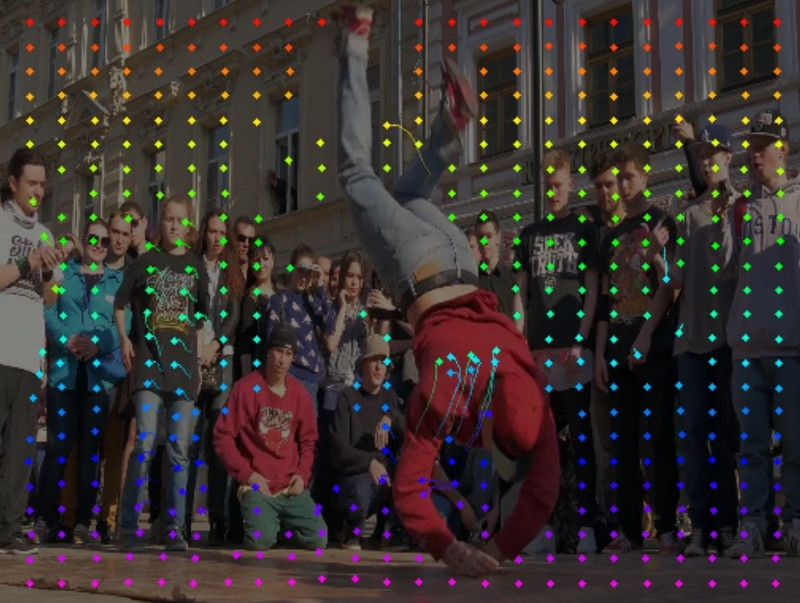}
      & \includegraphics[width=\abw]{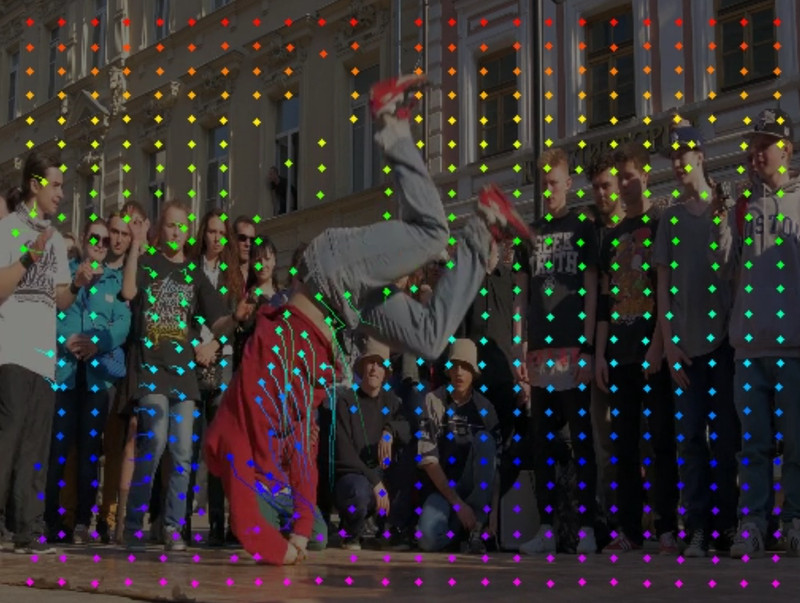} \\[\absep]
    \ablabel{Delta\,V2} %
      & \includegraphics[width=\abw]{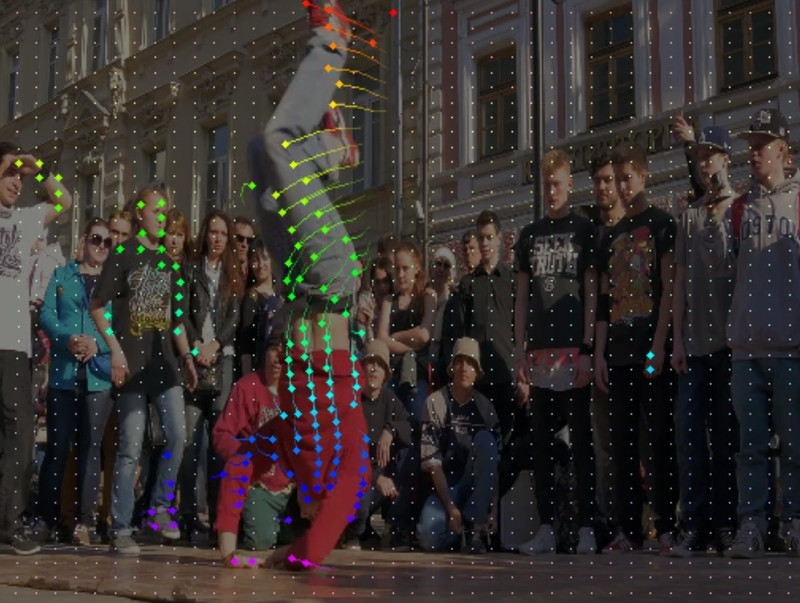}
      & \includegraphics[width=\abw]{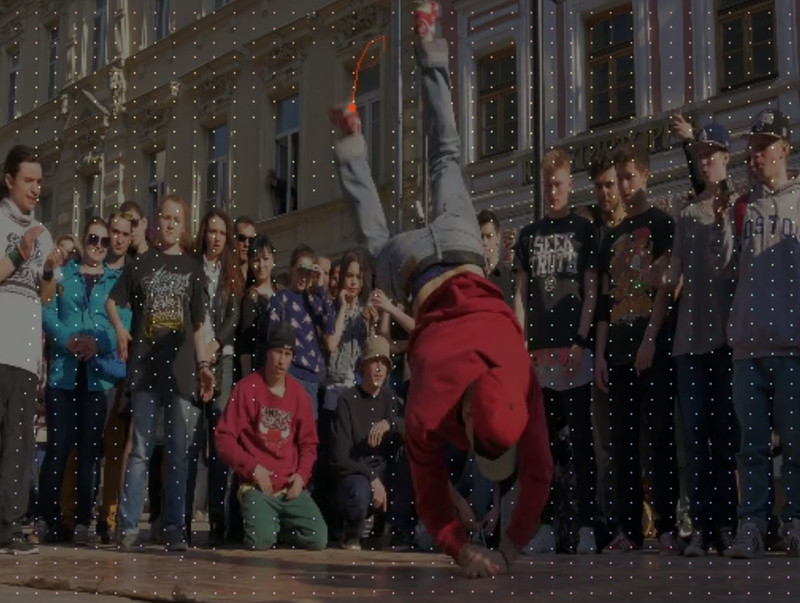}
      & \includegraphics[width=\abw]{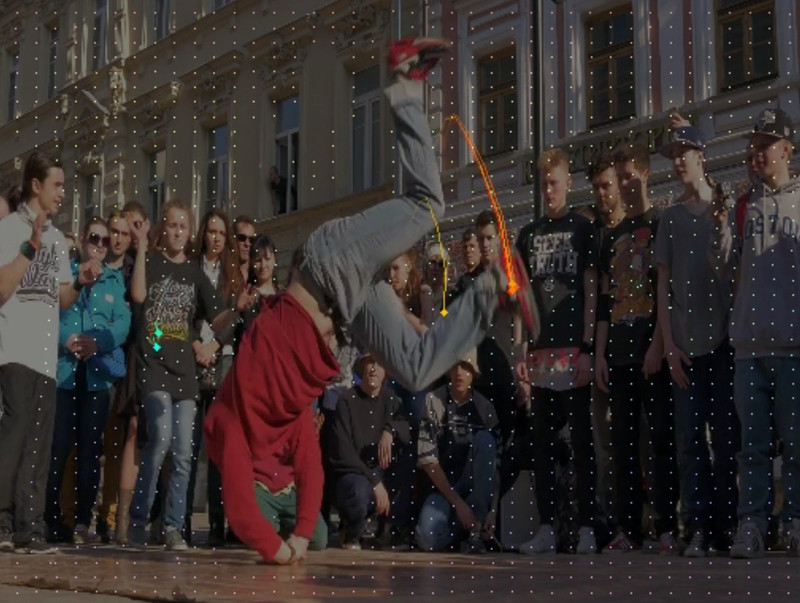} \\
  \end{tabular}\par
  \caption{\textbf{Qualitative comparison against baselines} on the Breakdance sequence
  (Dance-jump and hand manipulation continue on the following pages).
  Rows show \model, SpatialTracker-v2, CoTracker3, and DeltaV2; time advances left to right.
  \model tracks every visible point across time, including newly emerging content.
  Sparse trackers need human-specified queries and are usually demoed with a first-frame grid;
  densifying queries over later frames quickly becomes prohibitive (\Cref{fig:all_methods_benchmarking}).
  First-frame dense trackers have the same blind spot by design.
  In all visualizations, we show trails for dynamic points only; the method tracks all points and accounts for camera motion, but we omit static-point trails in the 2D renderings for clarity.}
  \label{fig:qualitative_results_supp}
\end{figure}

\clearpage
\begin{figure}[H]
  \ContinuedFloat
  \centering
  \noindent
  \setlength{\tabcolsep}{0pt}%
  \begin{tabular}{@{}c@{\hspace{\abgap}}c@{\hspace{\absep}}c@{\hspace{\absep}}c@{}}
    & \multicolumn{3}{c}{\small\bfseries Dance-jump} \\[4pt]
    \ablabel{Ours} %
      & \includegraphics[width=\abw]{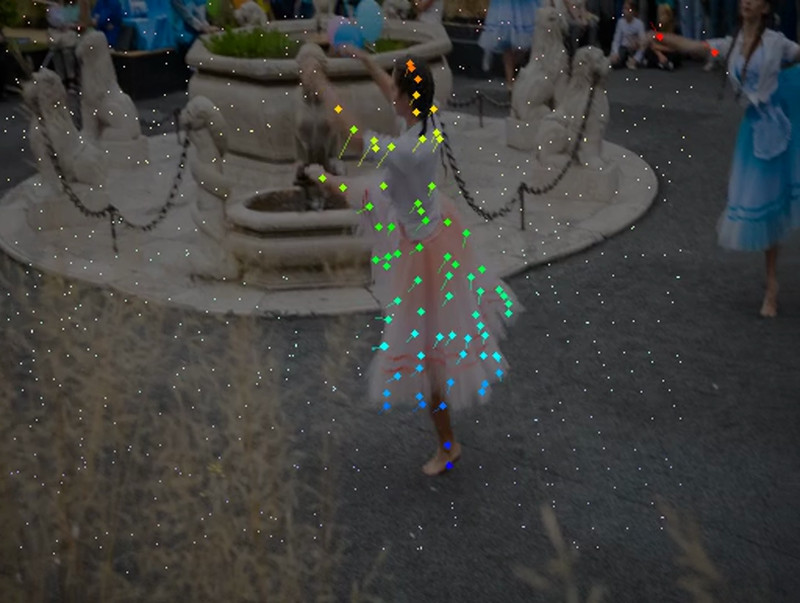}
      & \includegraphics[width=\abw]{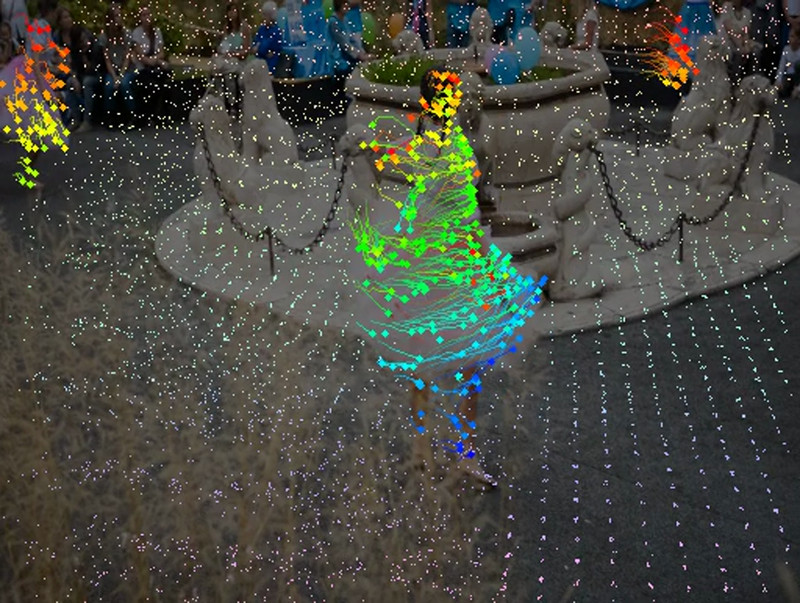}
      & \includegraphics[width=\abw]{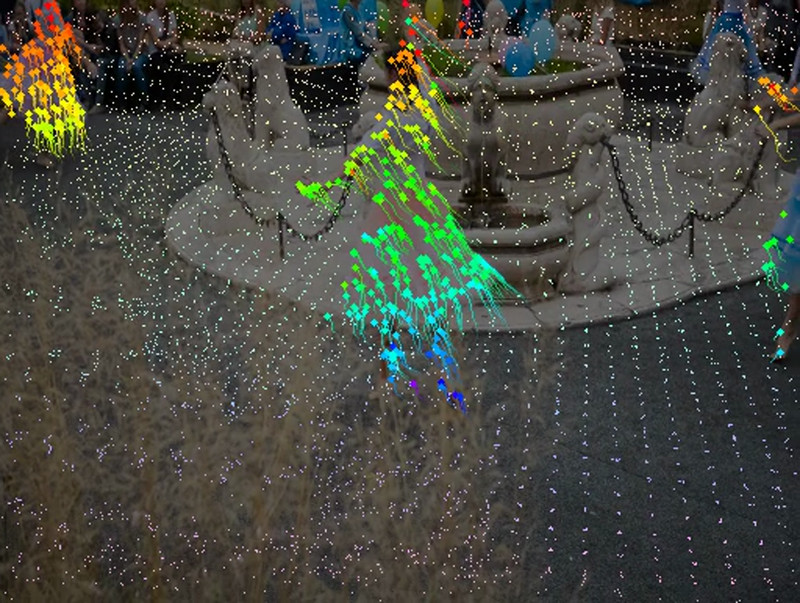} \\[\absep]
    \ablabel{SpaTracker\,V2} %
      & \includegraphics[width=\abw]{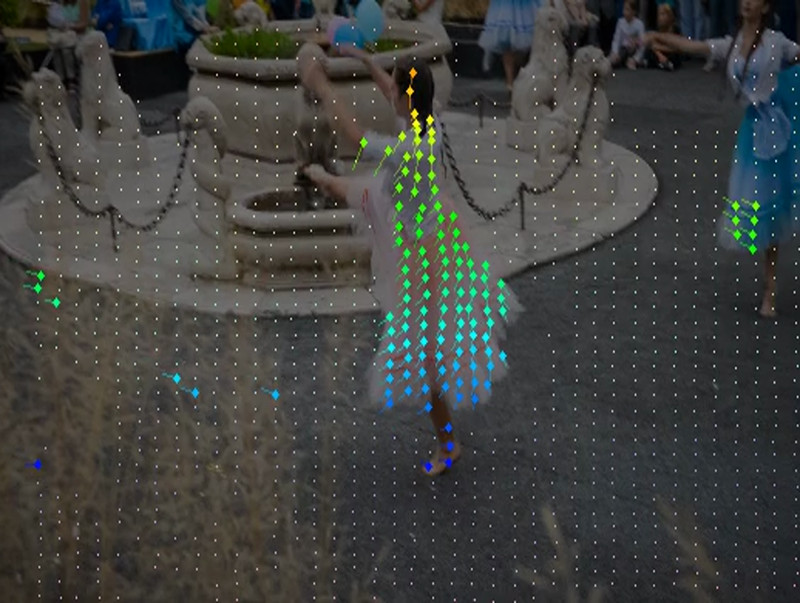}
      & \includegraphics[width=\abw]{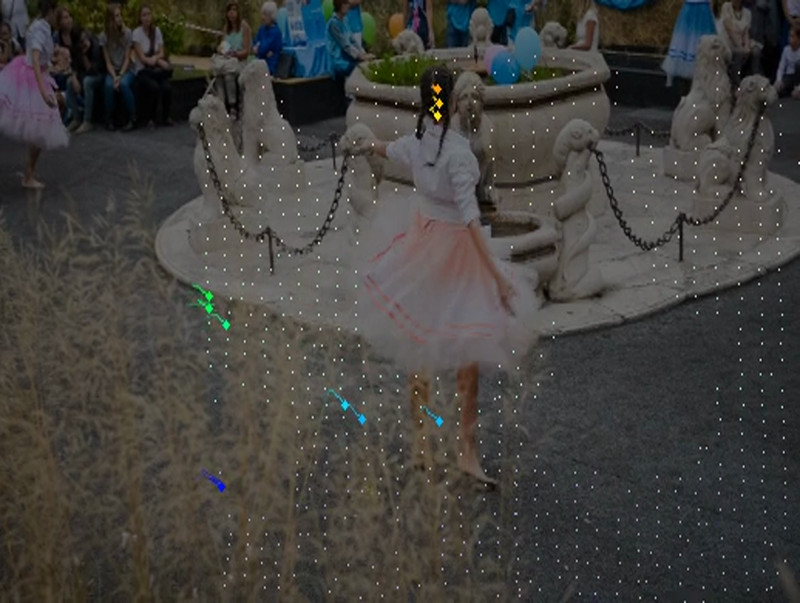}
      & \includegraphics[width=\abw]{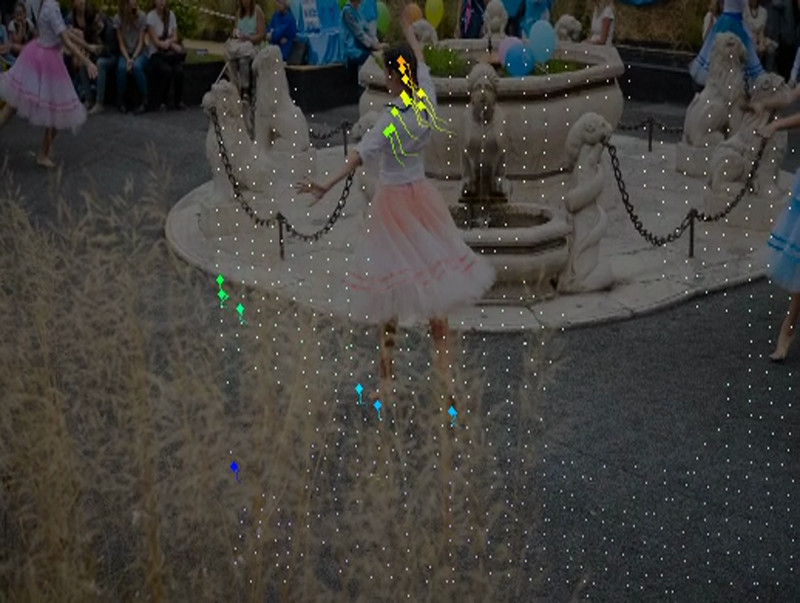} \\[\absep]
    \ablabel{CoTracker3} %
      & \includegraphics[width=\abw]{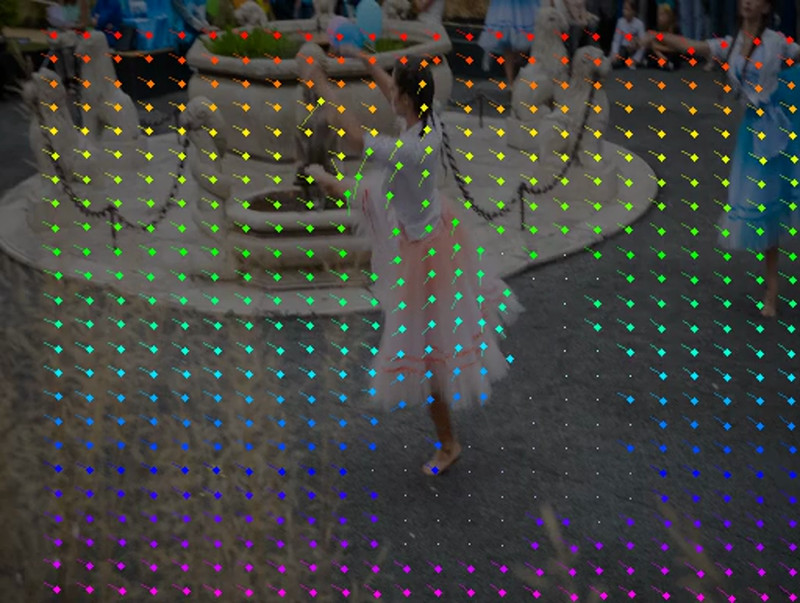}
      & \includegraphics[width=\abw]{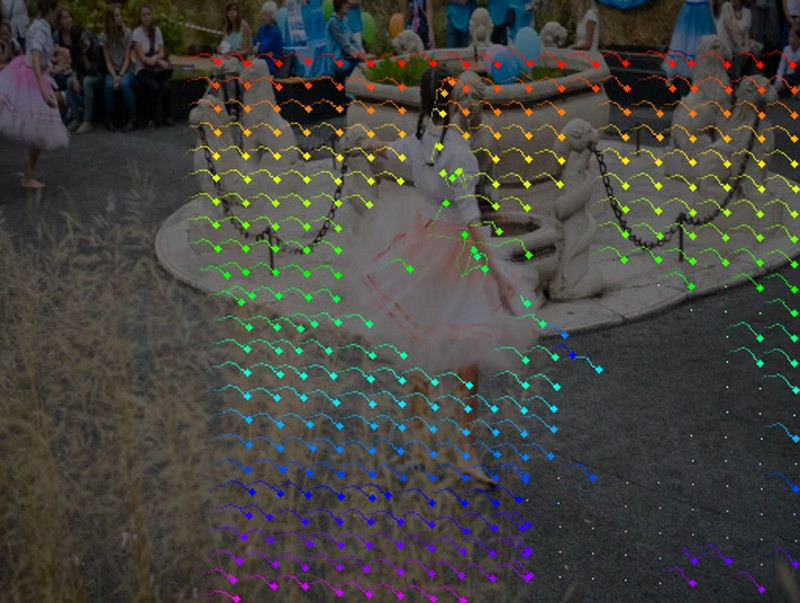}
      & \includegraphics[width=\abw]{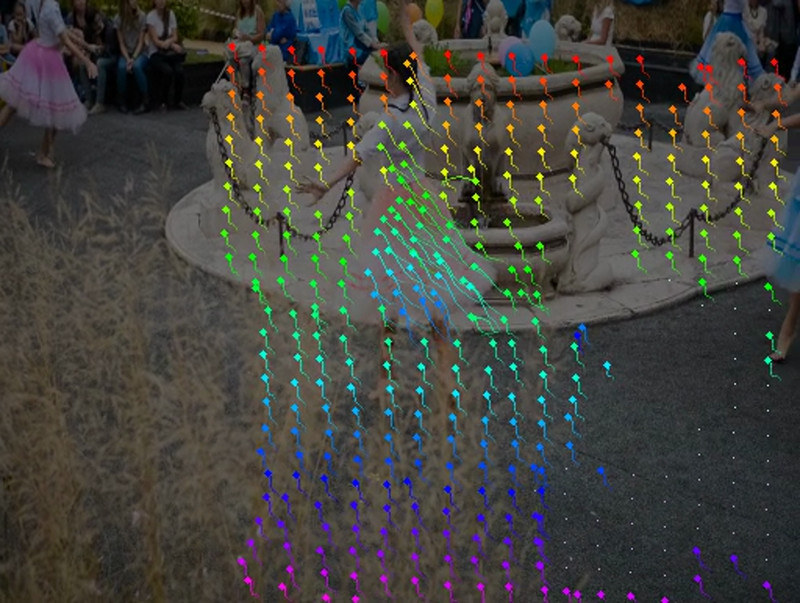} \\[\absep]
    \ablabel{Delta\,V2} %
      & \includegraphics[width=\abw]{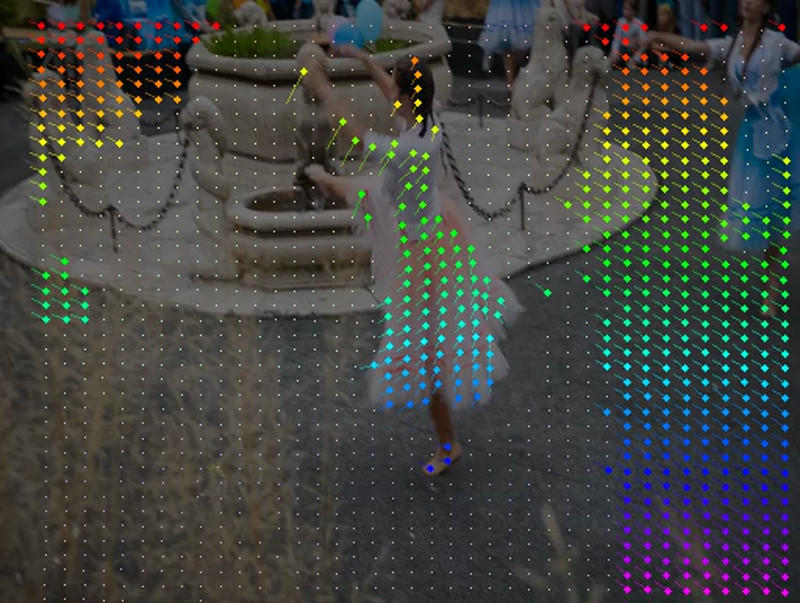}
      & \includegraphics[width=\abw]{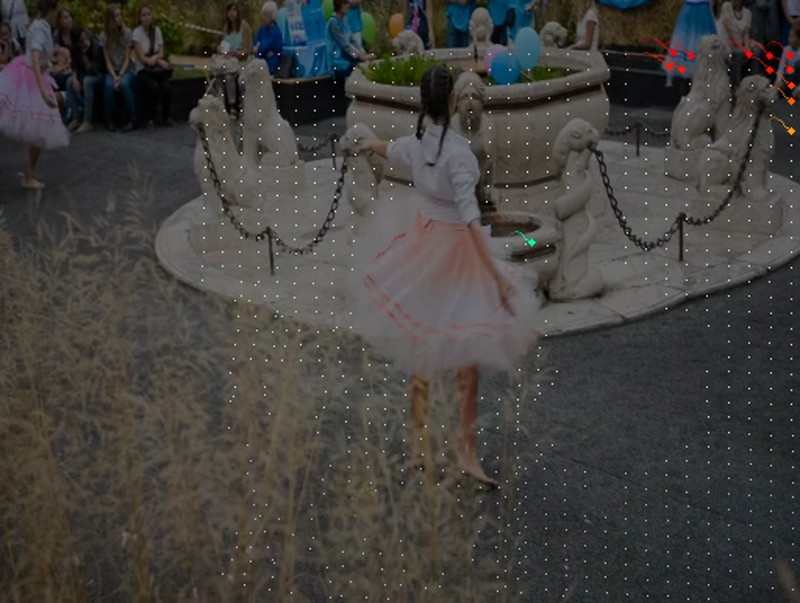}
      & \includegraphics[width=\abw]{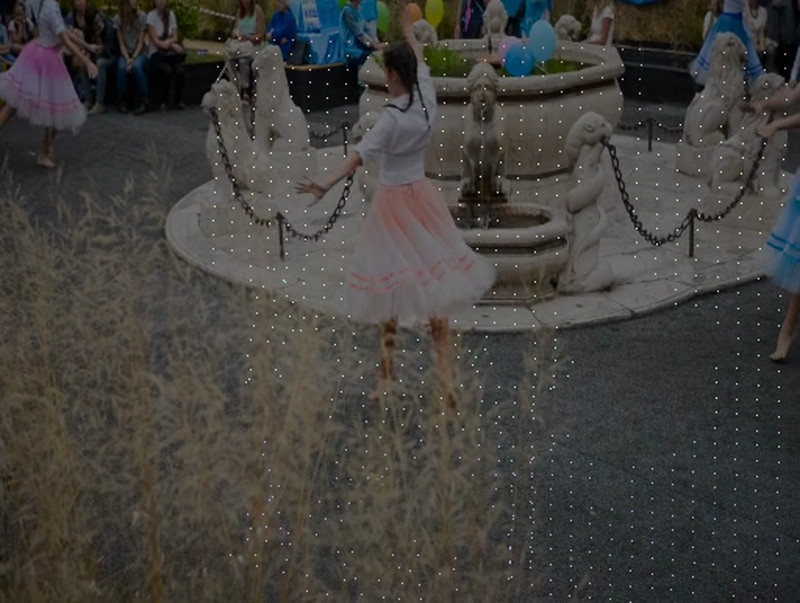} \\
  \end{tabular}\par
  \caption[]{\textbf{Qualitative comparison against baselines} (continued).
  Dance-jump sequence.}
\end{figure}

\clearpage
\begin{figure}[H]
  \ContinuedFloat
  \centering
  \noindent
  \setlength{\tabcolsep}{0pt}%
  \begin{tabular}{@{}c@{\hspace{\abgap}}c@{\hspace{\absep}}c@{\hspace{\absep}}c@{}}
    & \multicolumn{3}{c}{\small\bfseries Hand manipulation} \\[4pt]
    \ablabel{Ours} %
      & \includegraphics[width=\abw]{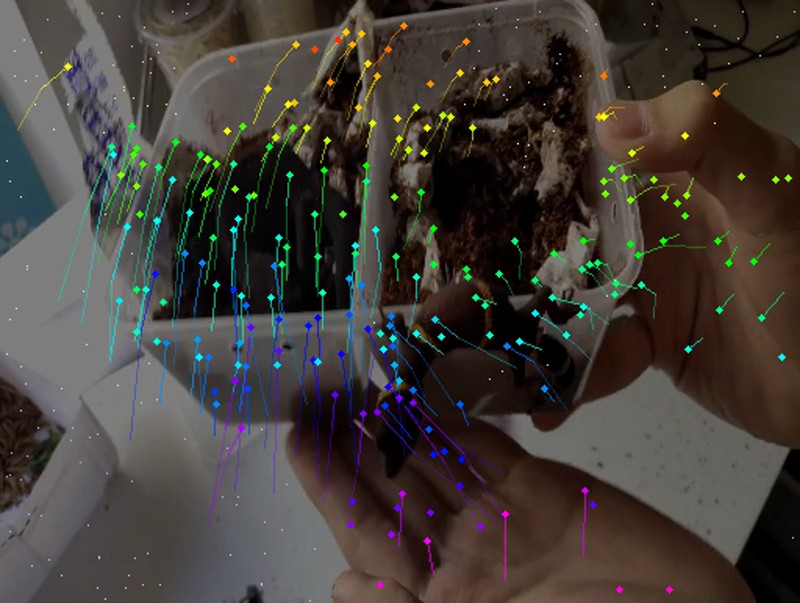}
      & \includegraphics[width=\abw]{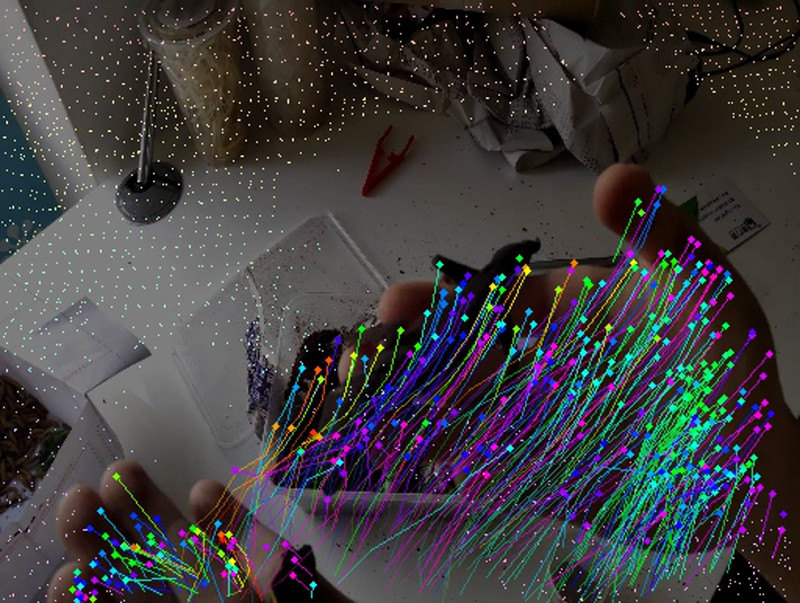}
      & \includegraphics[width=\abw]{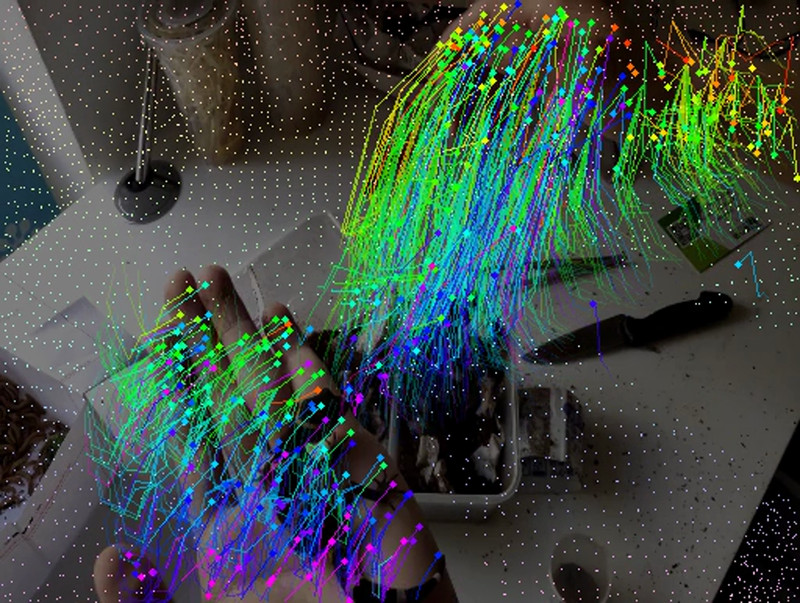} \\[\absep]
    \ablabel{SpaTracker\,V2} %
      & \includegraphics[width=\abw]{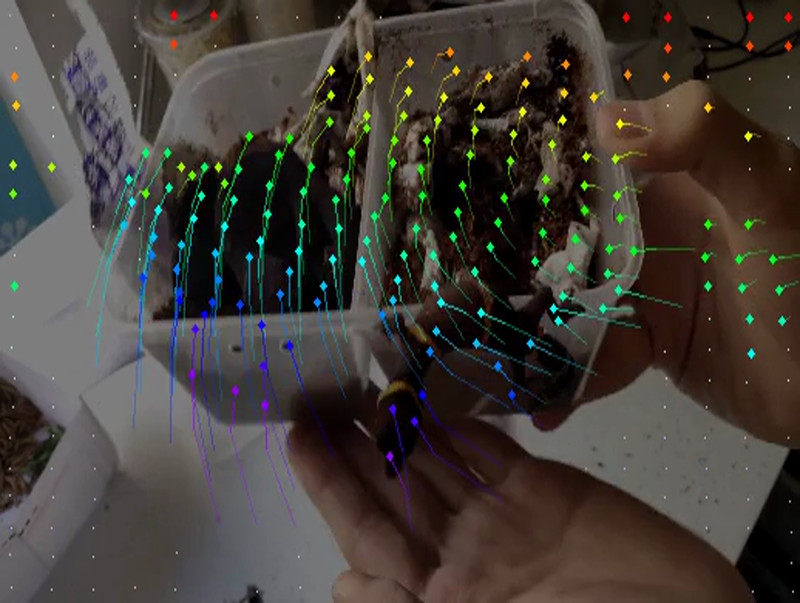}
      & \includegraphics[width=\abw]{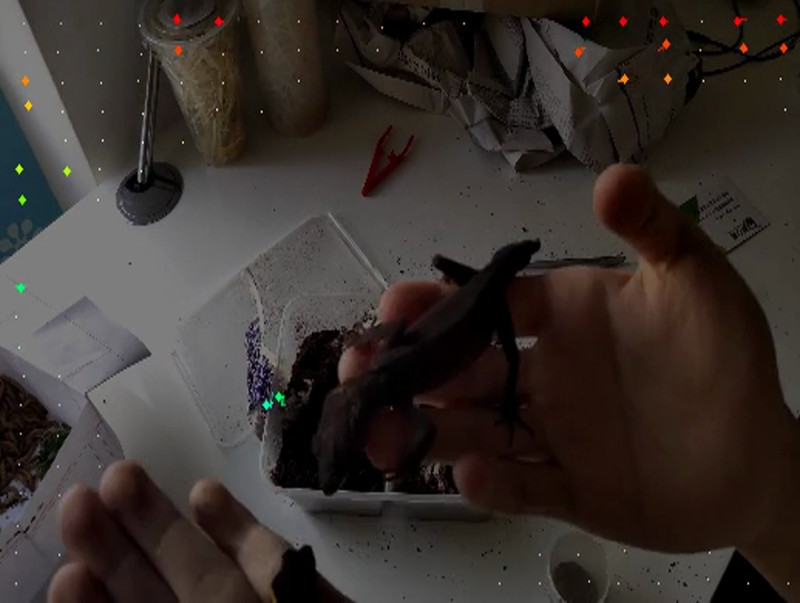}
      & \includegraphics[width=\abw]{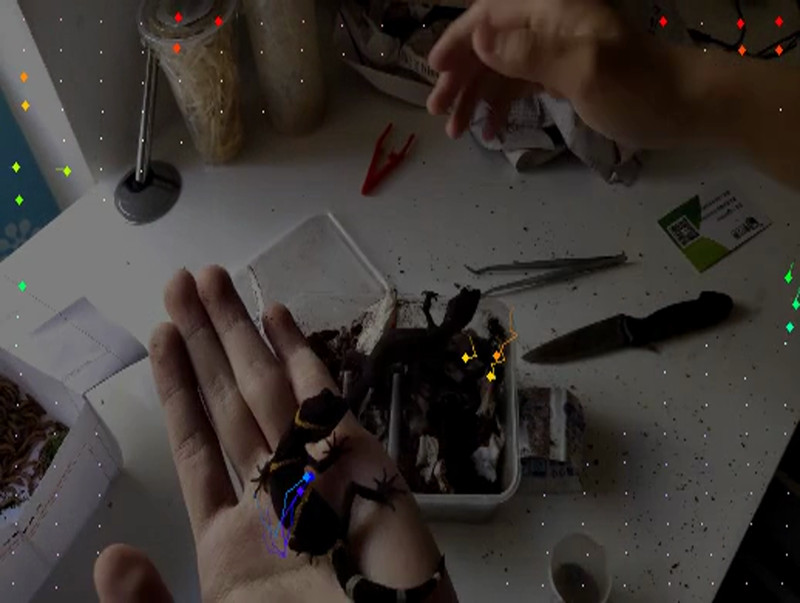} \\[\absep]
    \ablabel{CoTracker3} %
      & \includegraphics[width=\abw]{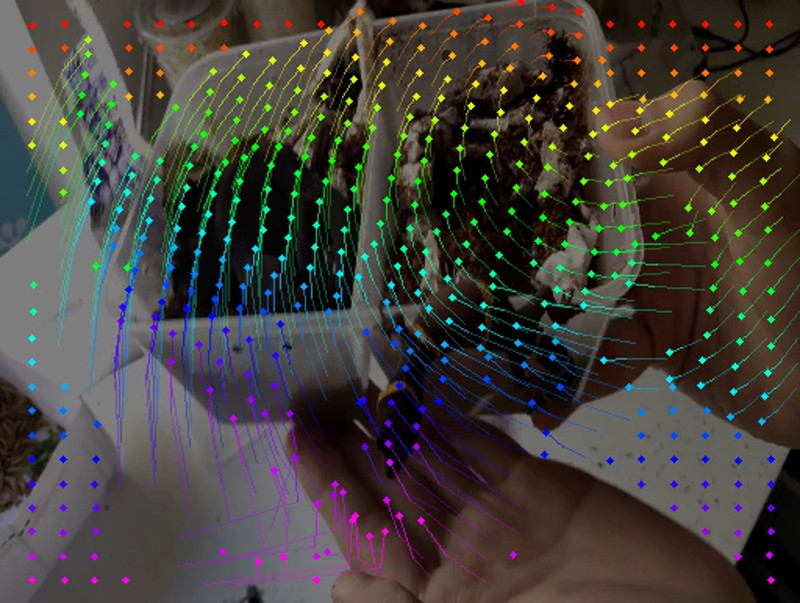}
      & \includegraphics[width=\abw]{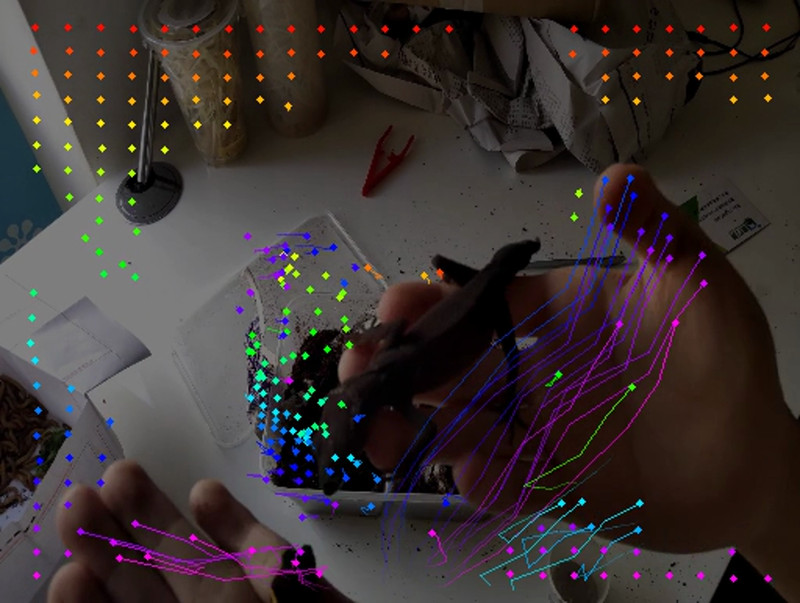}
      & \includegraphics[width=\abw]{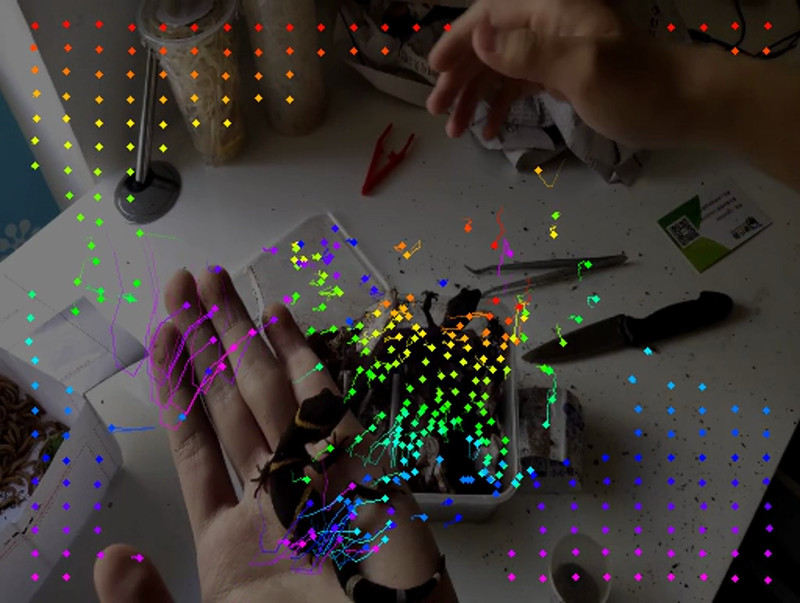} \\[\absep]
    \ablabel{Delta\,V2} %
      & \includegraphics[width=\abw]{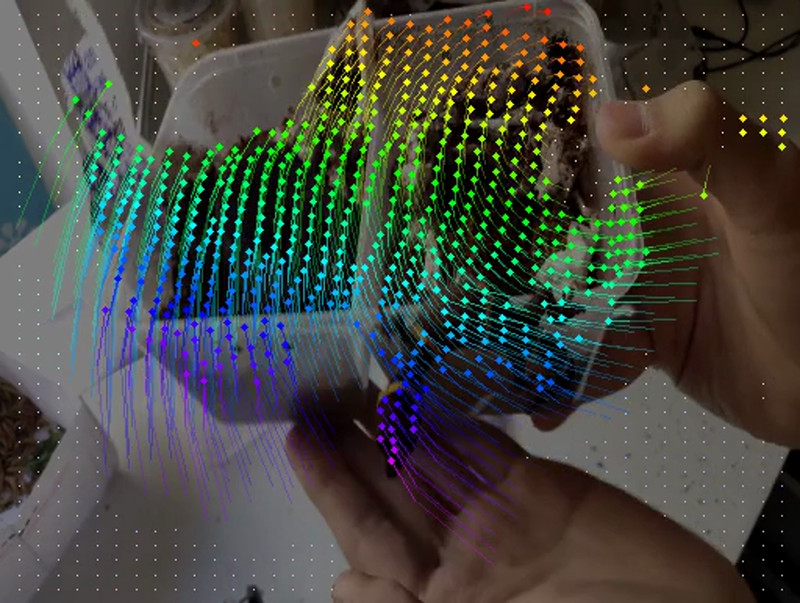}
      & \includegraphics[width=\abw]{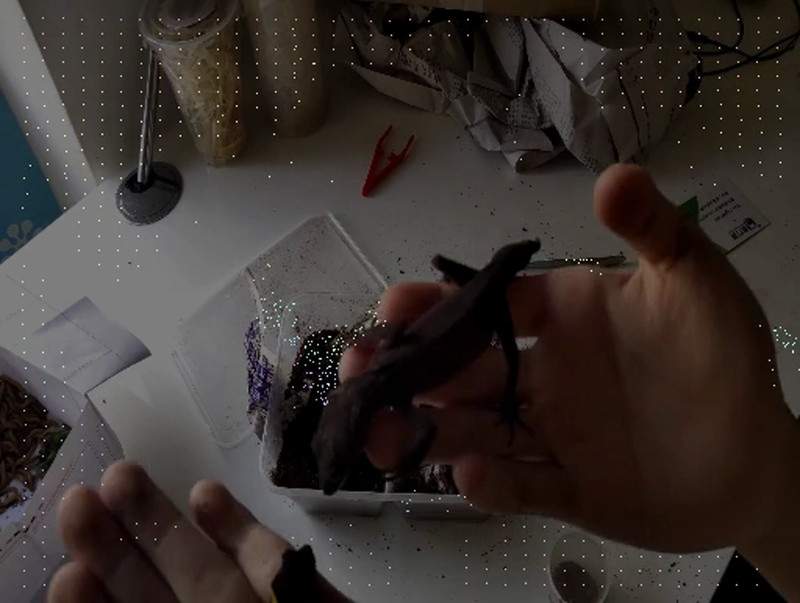}
      & \includegraphics[width=\abw]{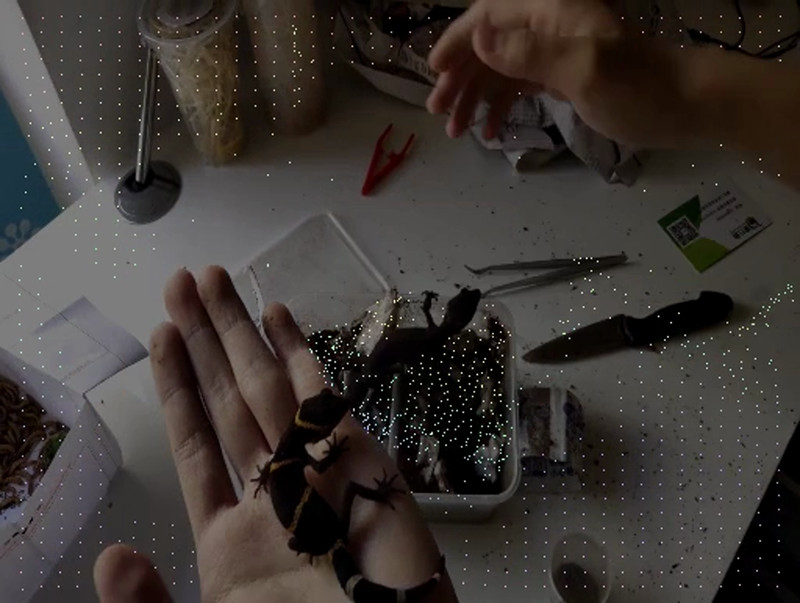} \\
  \end{tabular}\par
  \caption[]{\textbf{Qualitative comparison against baselines} (continued).
  Hand-manipulation sequence.}
\end{figure}

\newcommand{\sdcols}{5}

\newlength{\sdsep}\setlength{\sdsep}{1.5pt}
\newlength{\sdlabw}\setlength{\sdlabw}{10pt}
\newlength{\sdgap}\setlength{\sdgap}{3pt}
\newlength{\sdw}
\setlength{\sdw}{\dimexpr(\textwidth-\sdlabw-\sdgap-4\sdsep)/\sdcols\relax}

\newcommand{\sdlabel}[1]{%
  \makebox[\sdlabw][c]{%
    \raisebox{\dimexpr0.375\sdw-0.5\height\relax}{\rotatebox{90}{\scriptsize #1}}}}

\begin{figure}[H]
  \centering
  \setlength{\tabcolsep}{0pt}
  \begin{tabular}{@{}c@{\hspace{\sdgap}}c@{\hspace{\sdsep}}c@{\hspace{\sdsep}}c@{\hspace{\sdsep}}c@{\hspace{\sdsep}}c@{}}
    \sdlabel{Breakdance-flare} %
      & \includegraphics[width=\sdw]{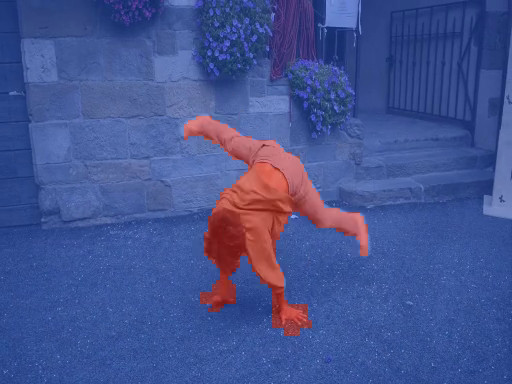}
      & \includegraphics[width=\sdw]{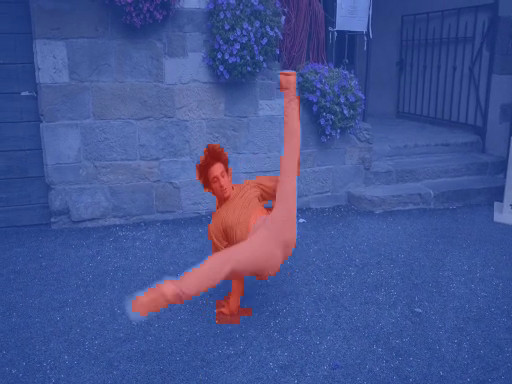}
      & \includegraphics[width=\sdw]{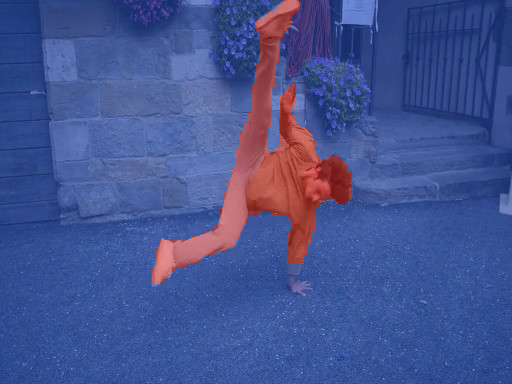}
      & \includegraphics[width=\sdw]{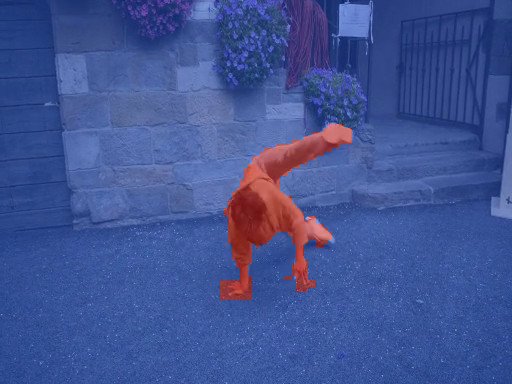}
      & \includegraphics[width=\sdw]{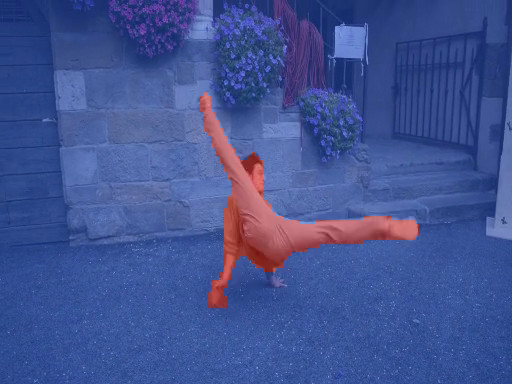} \\[\sdsep]
    \sdlabel{Dance-jump} %
      & \includegraphics[width=\sdw]{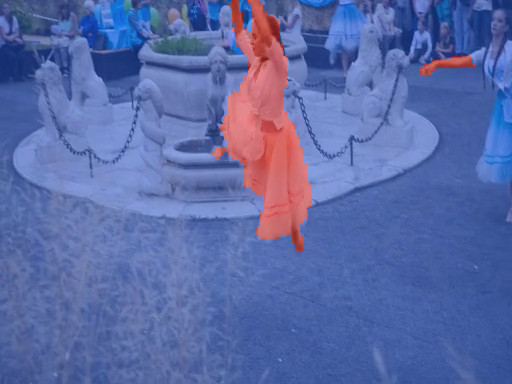}
      & \includegraphics[width=\sdw]{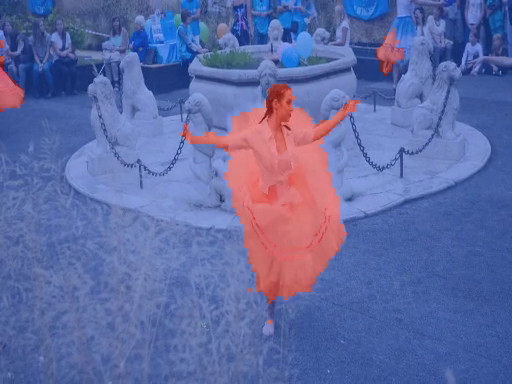}
      & \includegraphics[width=\sdw]{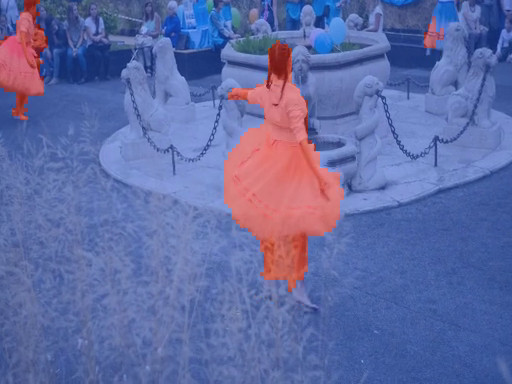}
      & \includegraphics[width=\sdw]{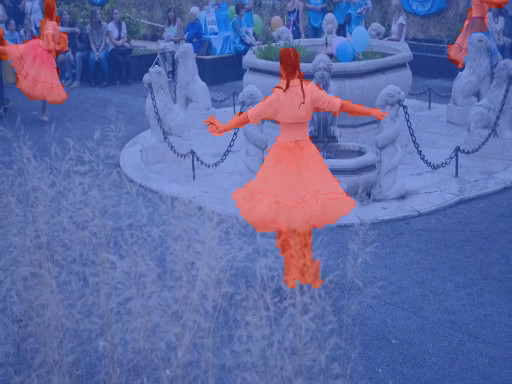}
      & \includegraphics[width=\sdw]{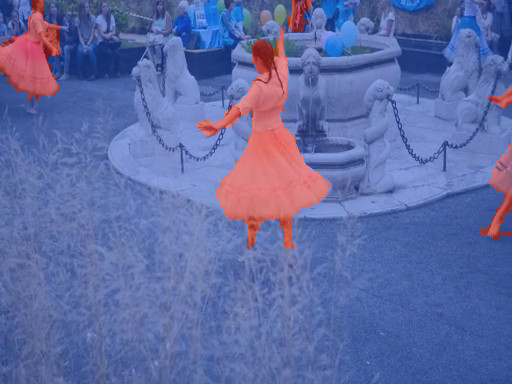} \\[\sdsep]
    \sdlabel{Tennis} %
      & \includegraphics[width=\sdw]{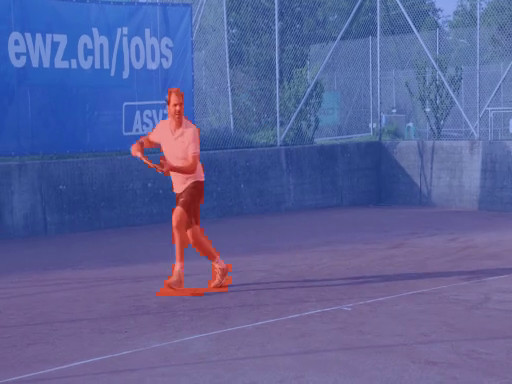}
      & \includegraphics[width=\sdw]{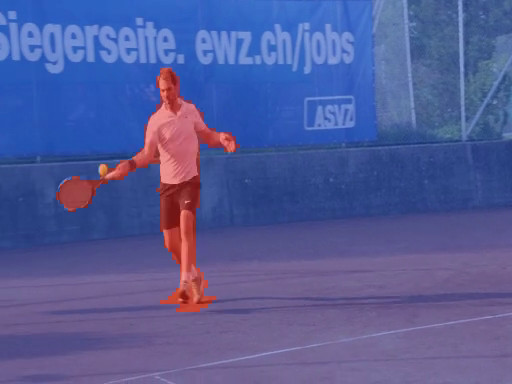}
      & \includegraphics[width=\sdw]{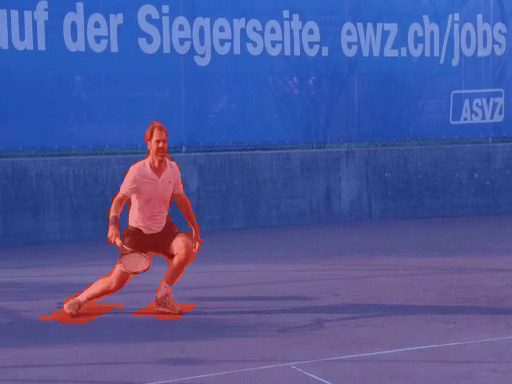}
      & \includegraphics[width=\sdw]{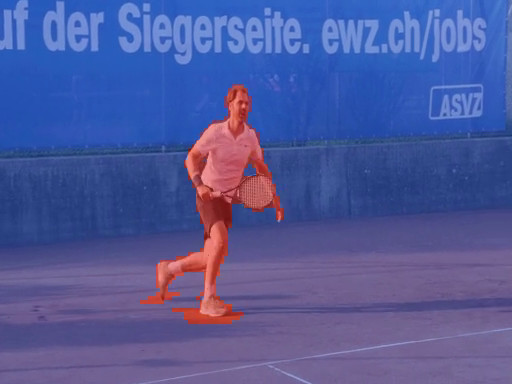}
      & \includegraphics[width=\sdw]{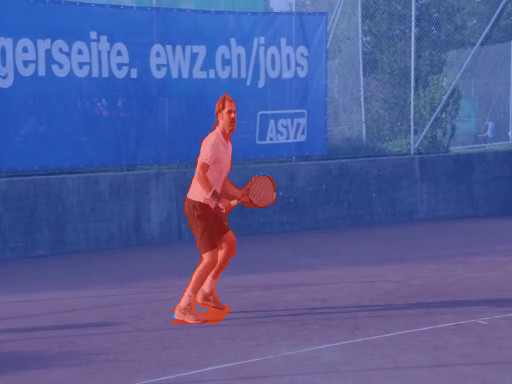} \\[\sdsep]
    \sdlabel{Road} %
      & \includegraphics[width=\sdw]{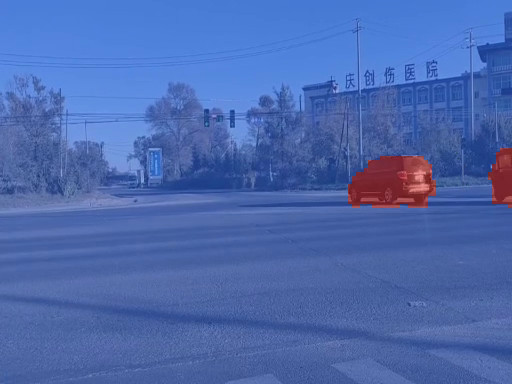}
      & \includegraphics[width=\sdw]{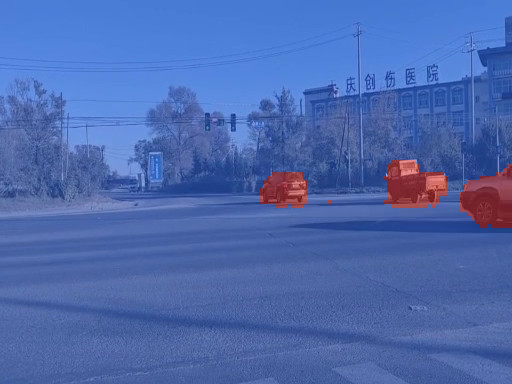}
      & \includegraphics[width=\sdw]{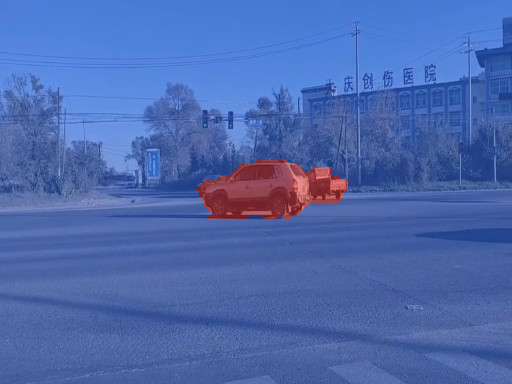}
      & \includegraphics[width=\sdw]{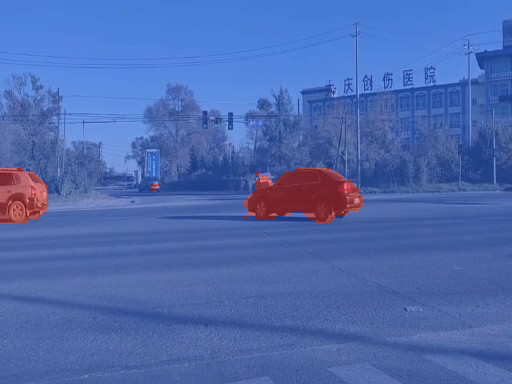}
      & \includegraphics[width=\sdw]{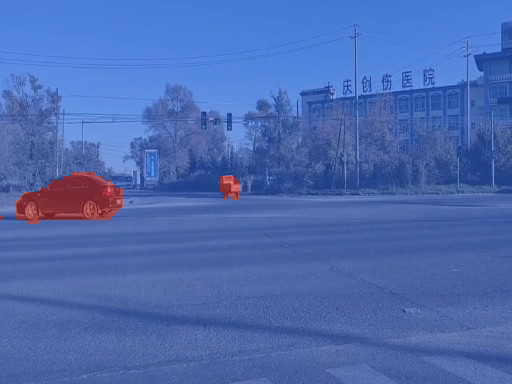} \\[\sdsep]
    \sdlabel{Wildlife} %
      & \includegraphics[width=\sdw]{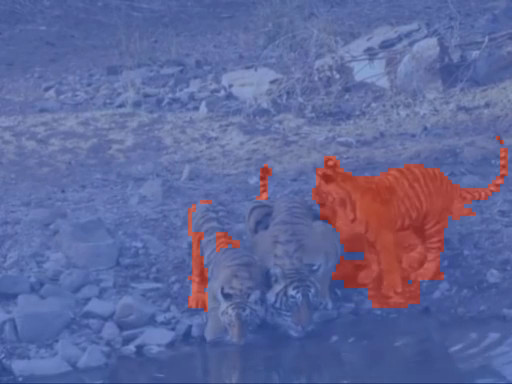}
      & \includegraphics[width=\sdw]{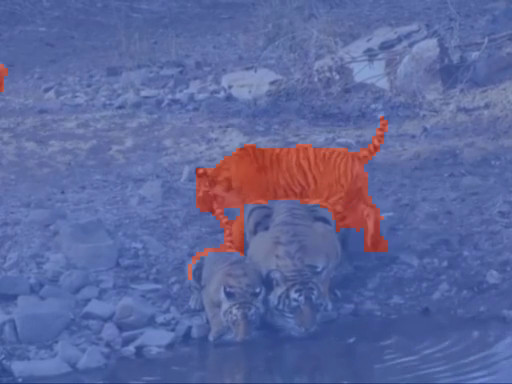}
      & \includegraphics[width=\sdw]{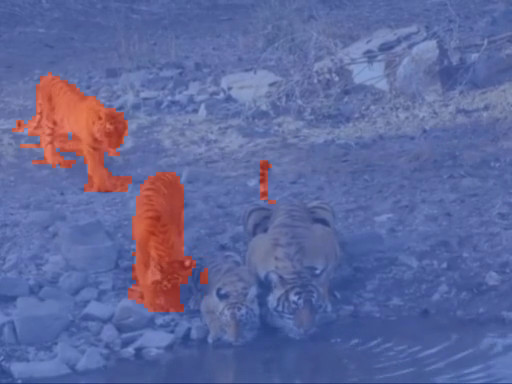}
      & \includegraphics[width=\sdw]{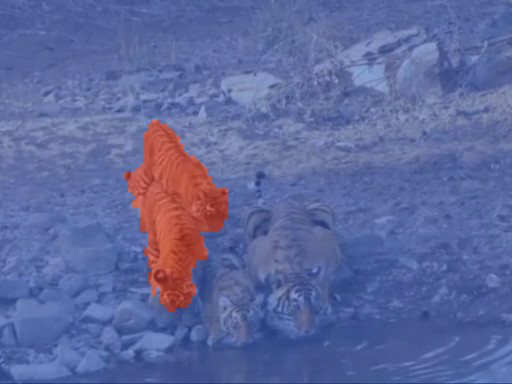}
      & \includegraphics[width=\sdw]{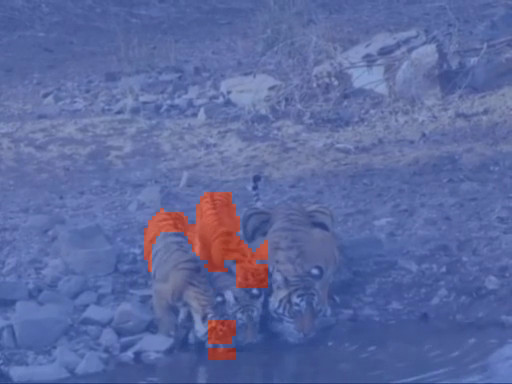} \\
  \end{tabular}
  \caption{%
    \textbf{Static/Dynamic Classification.}
    Points classified as dynamic are shown in red, static in blue.
    Time advances left to right. Notice how in the last example, the model correctly classifies some tigers as static while others as dynamic, and the static/dynamic assignment shifts as objects start or stop moving within a window.
  }
  \label{fig:static_dynamic}
\end{figure}

\subsection{TAPVid-3D Detailed Metrics}
\Cref{tab:tapvid3d_supp_48,tab:tapvid3d_supp_full} report AJ, OA, and EPE alongside APD-P / APD-M on TAPVid-3D under the same 48-frame and full-length regimes as \Cref{tab:tapvid3d_full}. All conclusions remain similar to those drawn from \Cref{tab:tapvid3d_full} in the main paper.
\providecommand{\dgray}[1]{\textcolor{gray!55}{#1}}
\providecommand{\na}{--}
\providecommand{\second}[1]{\underline{#1}}
\providecommand{\nrun}{N/A}
\providecommand{\nrunrow}{\nrun & \nrun & \nrun & \nrun & \nrun & \nrun & \nrun & \nrun}
\providecommand{\nrunrowg}{\dgray{\nrun} & \dgray{\nrun} & \dgray{\nrun} & \dgray{\nrun} & \dgray{\nrun} & \dgray{\nrun} & \dgray{\nrun} & \dgray{\nrun}}
\providecommand{\oom}{\textsc{oom}}
\providecommand{\TEname}{\shortstack[l]{\textbf{TrackEverything}\\[-0.12em]\textbf{(Ours)}}}
\providecommand{\TEnamesingle}{\textbf{TrackEverything (Ours)}}
\providecommand{\geoO}{$\Omega$}
\providecommand{\geoP}{Pi3}
\providecommand{\geoN}{--}
\definecolor{oursblue}{RGB}{232,240,254}
\providecommand{\oursrow}{\rowcolor{oursblue}}

\providecommand{\nrunxv}{\nrun & \nrun & \nrun & \nrun & \nrun & \nrun & \nrun & \nrun & \nrun & \nrun & \nrun & \nrun & \nrun & \nrun & \nrun & \nrun & \nrun}
\providecommand{\nrunxvg}{\dgray{\nrun} & \dgray{\nrun} & \dgray{\nrun} & \dgray{\nrun} & \dgray{\nrun} & \dgray{\nrun} & \dgray{\nrun} & \dgray{\nrun} & \dgray{\nrun} & \dgray{\nrun} & \dgray{\nrun} & \dgray{\nrun} & \dgray{\nrun} & \dgray{\nrun} & \dgray{\nrun} & \dgray{\nrun} & \dgray{\nrun}}

\begingroup
\scriptsize
\setlength{\tabcolsep}{0.85pt}
\renewcommand{\arraystretch}{1.02}
\setlength{\LTpre}{4pt}
\setlength{\LTpost}{8pt}
\begin{longtable}{@{}lc*{15}{c}@{\hspace{0.35em}}*{2}{c}@{}}
    \caption{\textbf{TAPVid-3D detailed metrics, 48-frame.}
    Bold / underline: best / second-best within the block (D4RT excluded).
    P/M: APD-P/APD-M; AJ: Average Jaccard; OA: occlusion accuracy; EPE: end-point error.
    $\Omega$: VGGT-$\Omega$; Pi3: Pi3; --: native geometry.}
    \label{tab:tapvid3d_supp_48}\\
    \toprule
    & & \multicolumn{5}{c}{ADT} & \multicolumn{5}{c}{DriveTrack} & \multicolumn{5}{c}{PStudio} & \multicolumn{2}{c}{Avg} \\
    \cmidrule(lr){3-7} \cmidrule(lr){8-12} \cmidrule(lr){13-17} \cmidrule{18-19}
    \textbf{Method} & Geo
    & P & M & AJ & OA & EPE$\downarrow$
    & P & M & AJ & OA & EPE$\downarrow$
    & P & M & AJ & OA & EPE$\downarrow$
    & P & M \\
    \midrule
    \endfirsthead
    \multicolumn{19}{@{}l}{\tablename~\thetable\ (continued)}\\[2pt]
    \toprule
    & & \multicolumn{5}{c}{ADT} & \multicolumn{5}{c}{DriveTrack} & \multicolumn{5}{c}{PStudio} & \multicolumn{2}{c}{Avg} \\
    \cmidrule(lr){3-7} \cmidrule(lr){8-12} \cmidrule(lr){13-17} \cmidrule{18-19}
    \textbf{Method} & Geo
    & P & M & AJ & OA & EPE$\downarrow$
    & P & M & AJ & OA & EPE$\downarrow$
    & P & M & AJ & OA & EPE$\downarrow$
    & P & M \\
    \midrule
    \endhead
    \bottomrule
    \endfoot
    \bottomrule
    \endlastfoot
        \multicolumn{19}{@{}l}{\textit{Sparse tracking}} \\
        CoTracker3 & \geoP
            & 38.7 & 88.0 & 27.2 & 88.8 & \second{0.16} & 29.2 & 41.9 & 20.3 & 88.5 & 2.13 & 23.7 & 83.1 & 14.5 & 85.0 & 0.18 & 30.5 & 71.0 \\
        CoTracker3 & \geoO
            & 43.1 & 87.8 & 31.3 & 88.8 & \second{0.16} & \second{29.7} & \textbf{42.8} & \textbf{21.4} & 88.5 & 2.33 & 29.6 & 87.9 & 19.0 & 85.0 & 0.13 & \second{34.1} & \textbf{72.8} \\
        SpatialTracker-v2 & \geoN
            & 37.7 & 87.0 & 26.9 & \second{90.9} & \textbf{0.15} & 28.6 & 41.3 & 20.0 & \textbf{89.1} & \textbf{1.62} & 25.1 & 84.1 & 17.2 & 82.1 & 0.15 & 30.5 & 70.8 \\
        SpatialTracker-v2 & \geoP
            & 39.9 & 88.1 & 28.9 & \textbf{91.0} & \second{0.16} & \textbf{29.8} & \second{42.5} & 20.6 & \second{89.0} & \second{1.63} & 20.6 & 80.7 & 12.5 & 83.7 & 0.20 & 30.1 & 70.4 \\
        SpatialTracker-v2 & \geoO
            & 43.8 & 87.4 & 32.9 & 90.7 & \second{0.16} & 29.4 & 42.4 & \second{21.1} & 88.8 & 1.97 & 24.9 & 84.2 & 15.8 & 82.2 & 0.17 & 32.7 & 71.3 \\
        TAPIP-3D & \geoP
            & 39.6 & 88.5 & 28.0 & 89.2 & \second{0.16} & 28.5 & 41.0 & 19.8 & 87.3 & 1.84 & 23.9 & 83.4 & 14.8 & 85.1 & 0.16 & 30.7 & 71.0 \\
        TAPIP-3D & \geoO
            & \second{45.6} & \textbf{89.0} & \second{34.0} & 88.6 & \textbf{0.15} & 26.7 & 39.4 & 19.2 & 87.3 & 2.31 & \second{30.0} & \second{88.2} & \second{19.5} & 85.5 & \second{0.12} & \second{34.1} & 72.2 \\
        \oursrow
        \TEnamesingle & \geoP
            & 39.3 & \second{88.9} & 28.5 & 88.8 & \second{0.16} & 27.7 & 39.8 & 19.4 & 84.6 & 2.04 & 23.5 & 83.6 & 15.1 & \second{88.4} & 0.16 & 30.2 & 70.8 \\
        \oursrow
        \TEnamesingle & \geoO
            & \textbf{45.7} & 88.7 & \textbf{34.4} & 90.2 & \textbf{0.15} & 28.2 & 40.3 & 20.1 & 83.4 & 2.26 & \textbf{30.1} & \textbf{88.7} & \textbf{20.2} & \textbf{88.9} & \textbf{0.11} & \textbf{34.7} & \second{72.6} \\
        \addlinespace[0.25em]
        \multicolumn{19}{@{}l}{\textit{First-frame dense tracking}} \\
        ST4RTrack & \geoN
            & \na & \na & \na & \na & \na & \na & 1.0 & \na & \na & 14.63 & \na & 53.1 & \na & \na & 0.21 & \na & \na \\
        Any4D & \geoN
            & 6.0 & 64.4 & 3.5 & \second{90.0} & 0.35 & 7.1 & 11.8 & 4.9 & 82.2 & 4.89 & 3.0 & 63.8 & 1.7 & 86.2 & 0.30 & 5.4 & 46.7 \\
        Any4D & \geoP
            & 6.9 & 64.0 & 4.3 & 86.4 & 0.35 & 6.9 & 9.7 & 3.8 & 81.2 & 4.52 & 3.7 & 64.1 & 2.1 & 83.5 & 0.31 & 5.8 & 45.9 \\
        Any4D & \geoO
            & 6.8 & 63.2 & 4.0 & 86.4 & 0.35 & 7.7 & 10.9 & 4.2 & 81.7 & 4.51 & 4.5 & 65.8 & 2.5 & 84.0 & 0.29 & 6.3 & 46.6 \\
        DeltaV2 & \geoP
            & 39.5 & 88.4 & 28.2 & 88.2 & \second{0.16} & \textbf{28.4} & \second{41.1} & 19.6 & 84.1 & \textbf{1.71} & 24.9 & \second{83.7} & 15.5 & 84.3 & \second{0.16} & 30.9 & 71.1 \\
        DeltaV2 & \geoO
            & \second{44.3} & \second{88.7} & \second{32.4} & 87.6 & \textbf{0.15} & \textbf{28.4} & \textbf{41.5} & \textbf{20.2} & \second{84.3} & \second{2.02} & \textbf{30.4} & \textbf{88.7} & \second{20.0} & 85.1 & \textbf{0.11} & \second{34.4} & \textbf{73.0} \\
        \oursrow
        \TEnamesingle & \geoP
            & 39.3 & \textbf{88.9} & 28.5 & 88.8 & \second{0.16} & 27.7 & 39.8 & 19.4 & \textbf{84.6} & 2.04 & 23.5 & 83.6 & 15.1 & \second{88.4} & \second{0.16} & 30.2 & 70.8 \\
        \oursrow
        \TEnamesingle & \geoO
            & \textbf{45.7} & \second{88.7} & \textbf{34.4} & \textbf{90.2} & \textbf{0.15} & \second{28.2} & 40.3 & \second{20.1} & 83.4 & 2.26 & \second{30.1} & \textbf{88.7} & \textbf{20.2} & \textbf{88.9} & \textbf{0.11} & \textbf{34.7} & \second{72.6} \\
        \addlinespace[0.25em]
        \multicolumn{19}{@{}l}{\textit{All-frame dense tracking}} \\
        \dgray{D4RT} & \dgray{\geoN}
            & \dgray{40.8} & \dgray{\na} & \dgray{30.7} & \dgray{92.6} & \dgray{\na} & \dgray{41.0} & \dgray{\na} & \dgray{30.4} & \dgray{87.5} & \dgray{\na} & \dgray{49.6} & \dgray{\na} & \dgray{37.2} & \dgray{89.7} & \dgray{\na} & \dgray{43.8} & \dgray{\na} \\
        VDPM & \geoN
            & 4.8 & 65.0 & 2.7 & 86.3 & 0.34 & 14.7 & 23.3 & 10.4 & \textbf{85.3} & 2.86 & 13.1 & 82.1 & 8.6 & 82.2 & \second{0.15} & 10.9 & 56.8 \\
        \oursrow
        \TEnamesingle & \geoP
            & \second{39.3} & \textbf{88.9} & \second{28.5} & \second{88.8} & \second{0.16} & \second{27.7} & \second{39.8} & \second{19.4} & \second{84.6} & \textbf{2.04} & \second{23.5} & \second{83.6} & \second{15.1} & \second{88.4} & 0.16 & \second{30.2} & \second{70.8} \\
        \oursrow
        \TEnamesingle & \geoO
            & \textbf{45.7} & \second{88.7} & \textbf{34.4} & \textbf{90.2} & \textbf{0.15} & \textbf{28.2} & \textbf{40.3} & \textbf{20.1} & 83.4 & \second{2.26} & \textbf{30.1} & \textbf{88.7} & \textbf{20.2} & \textbf{88.9} & \textbf{0.11} & \textbf{34.7} & \textbf{72.6} \\
\end{longtable}
\endgroup

\begingroup
\scriptsize
\setlength{\tabcolsep}{0.85pt}
\renewcommand{\arraystretch}{1.02}
\setlength{\LTpre}{4pt}
\setlength{\LTpost}{8pt}
\begin{longtable}{@{}lc*{15}{c}@{\hspace{0.35em}}*{2}{c}@{}}
    \caption{\textbf{TAPVid-3D detailed metrics, full-length.}
    Bold / underline: best / second-best within the block (D4RT excluded).
    N/A: method cannot run on full-length video.
    Column definitions match \Cref{tab:tapvid3d_supp_48}.}
    \label{tab:tapvid3d_supp_full}\\
    \toprule
    & & \multicolumn{5}{c}{ADT} & \multicolumn{5}{c}{DriveTrack} & \multicolumn{5}{c}{PStudio} & \multicolumn{2}{c}{Avg} \\
    \cmidrule(lr){3-7} \cmidrule(lr){8-12} \cmidrule(lr){13-17} \cmidrule{18-19}
    \textbf{Method} & Geo
    & P & M & AJ & OA & EPE$\downarrow$
    & P & M & AJ & OA & EPE$\downarrow$
    & P & M & AJ & OA & EPE$\downarrow$
    & P & M \\
    \midrule
    \endfirsthead
    \multicolumn{19}{@{}l}{\tablename~\thetable\ (continued)}\\[2pt]
    \toprule
    & & \multicolumn{5}{c}{ADT} & \multicolumn{5}{c}{DriveTrack} & \multicolumn{5}{c}{PStudio} & \multicolumn{2}{c}{Avg} \\
    \cmidrule(lr){3-7} \cmidrule(lr){8-12} \cmidrule(lr){13-17} \cmidrule{18-19}
    \textbf{Method} & Geo
    & P & M & AJ & OA & EPE$\downarrow$
    & P & M & AJ & OA & EPE$\downarrow$
    & P & M & AJ & OA & EPE$\downarrow$
    & P & M \\
    \midrule
    \endhead
    \bottomrule
    \endfoot
    \bottomrule
    \endlastfoot
        \multicolumn{19}{@{}l}{\textit{Sparse tracking}} \\
        CoTracker3 & \geoP
            & 31.4 & 82.3 & 18.9 & 84.8 & 0.20 & 24.3 & 40.5 & 17.0 & 89.7 & 2.30 & 18.7 & 80.2 & 10.6 & 81.7 & 0.20 & 24.8 & 67.7 \\
        CoTracker3 & \geoO
            & 34.5 & 84.4 & 21.5 & 84.8 & \textbf{0.18} & \second{27.4} & \second{43.9} & \second{19.8} & 89.7 & 2.21 & 27.1 & 87.9 & 16.5 & 81.7 & \second{0.13} & 29.7 & \textbf{72.1} \\
        SpatialTracker-v2 & \geoN
            & \oom & \oom & \oom & \oom & \oom & \oom & \oom & \oom & \oom & \oom & 23.9 & 84.0 & 15.5 & 80.5 & 0.15 & \na & \na \\
        SpatialTracker-v2 & \geoP
            & 32.1 & 82.4 & \na & \na & \na & 26.2 & 43.5 & 18.7 & \textbf{91.0} & \textbf{1.46} & 17.2 & 78.5 & 10.3 & 82.3 & 0.22 & 25.2 & 68.1 \\
        SpatialTracker-v2 & \geoO
            & 36.6 & 84.6 & 23.5 & 85.1 & \second{0.19} & \textbf{28.1} & \textbf{45.1} & \textbf{20.2} & \second{90.7} & \second{1.58} & 23.7 & 84.6 & 15.0 & 80.9 & 0.17 & 29.5 & 71.4 \\
        TAPIP-3D & \geoP
            & 32.7 & 82.0 & 21.9 & \second{89.3} & 0.21 & 25.0 & 40.9 & 17.7 & 88.0 & 2.01 & 19.1 & 80.7 & 11.0 & 82.9 & 0.19 & 25.6 & 67.9 \\
        TAPIP-3D & \geoO
            & \second{37.2} & \second{85.0} & \second{25.6} & \textbf{89.4} & \textbf{0.18} & 24.5 & 40.3 & 17.6 & 88.3 & 2.17 & \textbf{27.6} & \second{88.2} & \second{17.2} & 83.1 & \textbf{0.12} & \second{29.8} & 71.2 \\
        \oursrow
        \TEnamesingle & \geoP
            & 31.0 & 81.9 & 21.1 & 87.9 & 0.21 & 24.7 & 40.7 & 17.3 & 84.2 & 2.04 & 18.9 & 81.0 & 11.4 & \second{86.0} & 0.18 & 24.9 & 67.9 \\
        \oursrow
        \TEnamesingle & \geoO
            & \textbf{40.0} & \textbf{85.1} & \textbf{29.1} & 88.9 & \second{0.19} & 26.5 & 42.4 & 18.9 & 83.1 & 2.03 & \second{27.2} & \textbf{88.5} & \textbf{17.7} & \textbf{87.1} & \textbf{0.12} & \textbf{31.2} & \second{72.0} \\
        \addlinespace[0.25em]
        \multicolumn{19}{@{}l}{\textit{First-frame dense tracking}} \\
        ST4RTrack & \geoN
            & \nrunxv \\
        Any4D & \geoN
            & \nrunxv \\
        Any4D & \geoP
            & \nrunxv \\
        Any4D & \geoO
            & \nrunxv \\
        DeltaV2 & \geoP
            & 30.9 & 78.5 & 21.8 & \second{88.1} & 0.40 & \second{26.8} & \second{43.7} & 18.6 & 82.2 & \textbf{1.62} & 19.8 & 80.8 & 11.8 & 82.0 & 0.19 & 25.8 & 67.7 \\
        DeltaV2 & \geoO
            & \second{34.7} & 79.9 & \second{25.1} & 87.6 & 0.37 & \textbf{27.3} & \textbf{44.2} & \textbf{19.5} & 82.4 & \second{1.73} & \textbf{27.7} & \second{87.9} & \second{17.6} & 83.5 & \second{0.13} & \second{29.9} & \second{70.7} \\
        \oursrow
        \TEnamesingle & \geoP
            & 31.0 & \second{81.9} & 21.1 & 87.9 & \second{0.21} & 24.7 & 40.7 & 17.3 & \textbf{84.2} & 2.04 & 18.9 & 81.0 & 11.4 & \second{86.0} & 0.18 & 24.9 & 67.9 \\
        \oursrow
        \TEnamesingle & \geoO
            & \textbf{40.0} & \textbf{85.1} & \textbf{29.1} & \textbf{88.9} & \textbf{0.19} & 26.5 & 42.4 & \second{18.9} & \second{83.1} & 2.03 & \second{27.2} & \textbf{88.5} & \textbf{17.7} & \textbf{87.1} & \textbf{0.12} & \textbf{31.2} & \textbf{72.0} \\
        \addlinespace[0.25em]
        \multicolumn{19}{@{}l}{\textit{All-frame dense tracking}} \\*
        \dgray{D4RT} & \dgray{\geoN}
            & \nrunxvg \\*
        VDPM & \geoN
            & \nrunxv \\*
        \oursrow
        \TEnamesingle & \geoP
            & \second{31.0} & \second{81.9} & \second{21.1} & \second{87.9} & \second{0.21} & \second{24.7} & \second{40.7} & \second{17.3} & \textbf{84.2} & \second{2.04} & \second{18.9} & \second{81.0} & \second{11.4} & \second{86.0} & \second{0.18} & \second{24.9} & \second{67.9} \\*
        \oursrow
        \TEnamesingle & \geoO
            & \textbf{40.0} & \textbf{85.1} & \textbf{29.1} & \textbf{88.9} & \textbf{0.19} & \textbf{26.5} & \textbf{42.4} & \textbf{18.9} & \second{83.1} & \textbf{2.03} & \textbf{27.2} & \textbf{88.5} & \textbf{17.7} & \textbf{87.1} & \textbf{0.12} & \textbf{31.2} & \textbf{72.0} \\
\end{longtable}
\endgroup

\subsection{Complexity Analysis}
\begin{figure}[H]
  \centering
  \includegraphics[width=\textwidth]{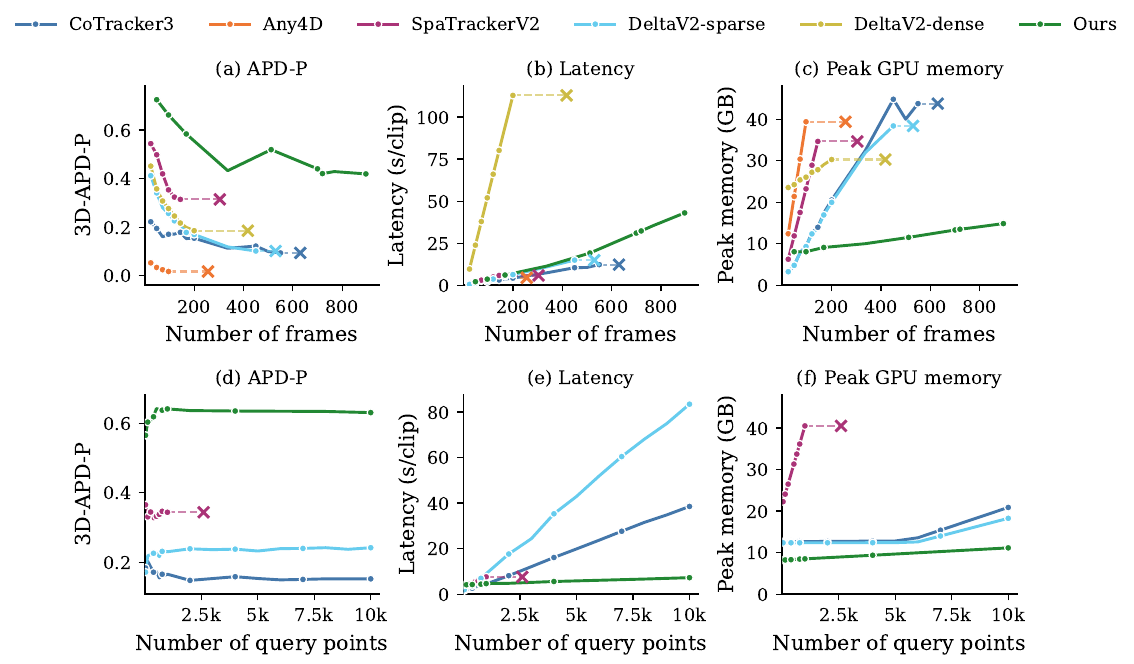}
  \caption{\textbf{Complexity analysis on PointOdyssey (APD-P).} Accuracy, latency, and peak GPU memory as functions of (top row, a–c) video length at 384 query points, and (bottom row, d–f) number of query points at 120 frames. ‘x’ marks out-of-memory.}
  \label{fig:combined_scaling_APD-P}
\end{figure}

We repeat the complexity analysis of \Cref{fig:all_methods_benchmarking} in the main paper using the APD-P metric instead of APD-M; results are shown in \Cref{fig:combined_scaling_APD-P}. All conclusions remain similar.

\end{document}